\documentclass{article}

 \usepackage[preprint]{neurips_2026}

\usepackage[utf8]{inputenc} % allow utf-8 input
\usepackage[T1]{fontenc}    % use 8-bit T1 fonts
\usepackage{hyperref}       % hyperlinks
\usepackage{url}            % simple URL typesetting
\usepackage{booktabs}       % professional-quality tables
\usepackage{amsfonts}       % blackboard math symbols
\usepackage{nicefrac}       % compact symbols for 1/2, etc.
\usepackage{microtype}      % microtypography
\usepackage{xcolor}         % colors
\usepackage{amsmath}
\usepackage{amsthm}
\usepackage{multirow}
\usepackage{graphicx}
\usepackage{subcaption}
\usepackage{bm}
\usepackage{float}
\usepackage{algorithm}
\usepackage{algorithmic}

\title{Aperiodic and Low-Frequency Spectral Bias in Reconstruction based EEG Foundation Models}

\author{
  Aditya Kommineni$^{1}$ \\
  \texttt{akommine@usc.edu} \\
  \And 
  Emily Zhou$^{2}$ \\
  \texttt{emilyzho@usc.edu} \\
  \And 
  Kleanthis Avramidis$^{2}$ \\
  \texttt{avramidi@usc.edu} \\
  \And 
  Simon Bock Segaard$^{3}$\\
  \texttt{ssegaa21@student.aau.dk} \\
  \And 
  Jeppe Roden Münster$^{3}$\\
  \texttt{jmunst21@student.aau.dk} \\
  \And 
  Andreas Peter Juhl Hansen$^{3}$ \\
  \texttt{apjh21@student.aau.dk} \\
  \And 
  Takfarinas Medani$^{1}$ \\ 
  \texttt{medani@usc.edu} \\
  \And 
  Tiantian Feng$^{1}$ \\
  \texttt{tiantiaf@usc.edu} \\
  \And 
  Richard Leahy$^{1}$ \\
  \texttt{leahy@usc.edu} \\
  \And 
  Shrikanth Narayanan$^{1,2}$ \\
  \texttt{shri@usc.edu} \\
    \And
  Ming Hsieh Department of Electrical and Computer Engineering$^{1}$\\
  University of Southern California\\
  \And 
  Thomas Lord Department of Computer Science$^{2}$\\
  University of Southern California\\
  \And 
  Department of Mathematical Sciences$^{3}$ \\
  Aalborg University \\
}

\begin{document}

\maketitle

\begin{abstract}
  EEG foundation models, pre-trained on large-scale unlabelled EEG data, have emerged as a promising direction towards learning generalizable EEG representations. 
  Despite showing positive results in data-rich regimes, they often fail to outperform significantly smaller supervised models in low-resource settings compared to fully supervised models. 
  We provide a mechanistic account of this shortcoming, attributing it to a fundamental mismatch between reconstruction-based pretext tasks and the idiosyncratic spectral structure of EEG signals, which decompose into distinct high-power aperiodic and low-power oscillatory components.
  Using controlled, synthetically-generated EEG inputs, we demonstrate that EEG foundation model embeddings are biased to capture the aperiodic components of the EEG signal while under-representing oscillatory components, particularly at higher frequencies.
  Additionally, linear probe evaluations on real-world BCI datasets further reveal that embeddings encode subject identity more strongly than task-relevant information, thereby reinforcing the low-frequency and aperiodic component bias in foundation model embeddings trained primarily on reconstruction based objectives. 
  Together, these findings elucidate a failure mode in reconstruction based EEG foundation models and motivate future work to incorporate auxiliary losses explicitly targeting high-frequency oscillatory structure as a path toward more capable and generalizable EEG representations.
\end{abstract}

\section{Introduction}
Foundation models in language~\citep{gpt3, gpt4, gemini, gemini25}, speech~\citep{whisper} and vision~\citep{clip} modalities have resulted in impressive advances in the respective domains over the past few years.
These foundation models have provided generalized representations that are able to provide noticeable performance improvements over their fully supervised counterparts in low-resource settings, where large scale data collections are not feasible.
These developments have inspired numerous efforts to build generalized representations/foundation models for brain signals, especially Electroencephalography (EEG)~\citep{kostas2021bendr, wang2025cbramod, jiang2024large, zhou2025csbrain, liu2026echo, avramidis2025neural, s4eeg_kommineni, wang2024eegpt, chien2022maeeg, cui2024neuro}.
\par 
Despite being pre-trained with large-scale unlabelled EEG data, the performance improvements provided by EEG foundation models over fully supervised domain specific models remain limited.
While some prior works~\citep{wang2025cbramod, zhou2025csbrain} have hypothesized that these limitations could be alleviated through increased data scale and model size, recent works have not found strong effects with scaling model sizes in EEG foundation models~\citep{liu2026eeg}. 
Additionally, we argue this view is fundamentally constrained by the data acquisition challenges inherent to EEG.
Unlike text, audio, and images, domains where foundation models benefit from web-scale data, EEG recordings require specialized hardware, and trained personnel during acquisition, making large-scale data collection extremely resource-intensive and scaling laws difficult to realize in practice. 
In this work, we attempt to provide an explanation from the pretext task perspective. 
This view is further motivated by recent findings showing that EEG foundation models, when evaluated under resource-constrained settings, whether through limitations on training samples or model parameters frequently underperform their fully supervised counterparts~\citep{yang2026arefoundationmodelsworthit, liu2026eeg, kuruppu2026eeg}, warranting deeper investigation into the representations learned by these models.
Identifying the factors that lead to this phenomenon could provide insights to inform improved EEG foundation model design and construction. 
Current EEG foundation models borrow pre-training objectives from multimedia modalities (images, speech and text), which have fundamental differences to biological brain signals like EEG. In particular, there has been a predominant adoption of reconstruction based objectives as the primary means of training EEG foundation models~\citep{jiang2024large, wang2025cbramod, zhou2025csbrain, ouahidi2025reve}. The present work will hence focus on this formulation. 
\par  
Furthermore, EEG signals have unique characteristics that can impact the observed behavior of current foundation models. Notably, prior neuroscience work has consistently shown that spontaneous brain activity dominates measured signals in both energy and variance~\citep{raichle2006brain, raichle2010two},  with task-evoked responses constituting relatively small perturbations~\citep{eeg_variability}. 
In EEG, this is reflected in the prominence of scale-free (1/f) aperiodic activity~\citep{donoghue2020parameterizing}, which has been linked to subject-specific physiological properties such as age and cognitive states~\citep{donoghue2020parameterizing}. 
In this work, we hypothesize that the observed sub-optimal task classification in linear probe experiments for EEG foundation models can be attributed to the combination of reconstruction based objectives and aforementioned spectral dominance of aperiodic components in EEG.
\par 
Through controlled experiments on synthetically generated EEG signals and empirical results on real world EEG-BCI tasks, we show that reconstruction based EEG foundation models exhibit a spectral bias towards encoding aperiodic and low-frequency information.
This bias, consequently results in models representations forming subject-centric clusters over task-centric clusters as shown through empirical experiments on BCI datasets.
These observations motivate designing domain specific pretext tasks for EEG foundation models, that could include auxiliary losses to capture high-frequency oscillatory components. The contributions of this work are as formulated below:
\begin{itemize}
    \item \textbf{Spectral bias in reconstruction-based EEG foundation models}. We diagnose a key limitation of existing reconstruction based EEG foundation models- internal representations of models preferentially encode information of aperiodic, low-frequency components over high-frequency oscillatory information. We attribute this to a mismatch between standard reconstruction-based self-supervised objectives and the spectral structure of EEG signals.
    \item \textbf{Synthetic validation via controlled EEG simulations}. Using a controlled single-channel EEG simulation that independently varies aperiodic and oscillatory components, we show that model embeddings preferentially encode aperiodic structure. Specifically, aperiodic components are linearly decodable, while oscillatory components are only weakly decodable primarily at low frequencies, demonstrating a bias in representations.
    \item \textbf{Empirical evidence from BCI tasks}. Foundation model embeddings achieve significantly higher performance on subject identification than on task decoding. Given that the evaluated BCI tasks rely predominantly on oscillatory activity, this gap supports our hypothesis that models exhibit a spectral bias toward low-frequency, aperiodic features.
\end{itemize}
\section{Related Work}
\subsection{EEG Foundation Models}
Early attempts to build EEG foundation models were inspired from architectures and pretext tasks from image and speech domains such as contrastive predictive coding in BENDR~\citep{kostas2021bendr}, contrastive learning through augmentations of signals in BIOT~\citep{biot} and masked reconstruction based objectives in EEGPT~\citep{wang2024eegpt}.
Findings by ~\citep{chien2022maeeg} show masked reconstruction objectives perform better than contrastive objectives on EEG signals and better stability in masked reconstruction training, reconstruction based models have become the predominant method for EEG foundation model training.
Within reconstruction based models, two distinct paths were explored, one with discrete neural tokenizer augmented with masked token modelling~\citep{jiang2024large, avramidis2025neural} while the second included masked reconstruction on raw EEG signals~\citep{wang2025cbramod, zhou2025csbrain, ouahidi2025reve, wang2026deeperbrain, cui2024neuro, doner2025luna, chen2026unintfm}.

\subsection{Aperiodic and Oscillatory Components in EEG}
\begin{figure}
    \centering
    \includegraphics[width=\linewidth]{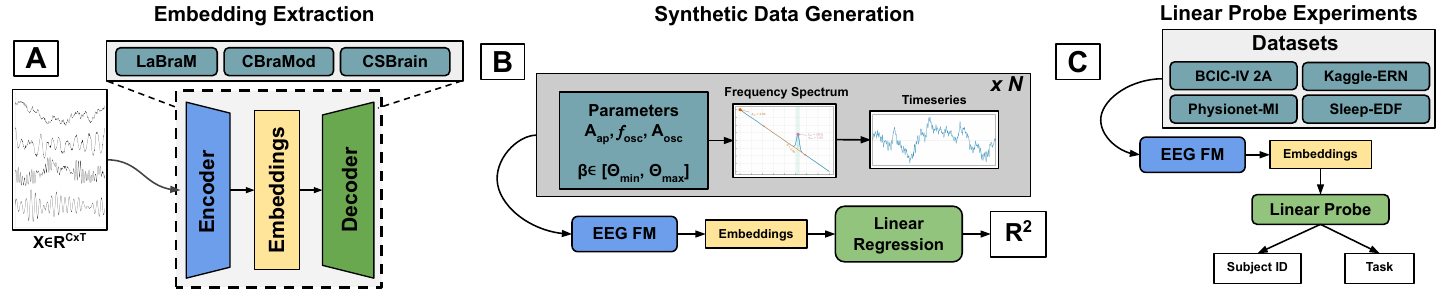}
    \caption{\textbf{Simulated and real-world EEG experiments} (A) Pipeline for obtaining embeddings from multi-channel EEG signals. Embeddings are extracted from the last layer of the encoder for three foundation models (LaBraM, CBraMod and CSBrain) (B) Pipeline for computing linear decodability through controlled synthetic single channel EEG generation (In the figure, aperiodic exponent $\beta$ is sampled between [$\theta_{min}$, $\theta_{max}$] to create N samples. Embeddings are extracted for samples followed by linear regression to compute linear decodability. (C) Linear probing experiments on four real-world EEG datasets for classification on subject identity and task. Better subject identity results would indicate that models capture more subject specific information in the internal representations.}
    \label{fig:schematic}
\end{figure}
The spectral information in EEG signals is composed of aperiodic broadband signal and oscillatory information~\citep{donoghue2020parameterizing, brake2024neurophysiological}.
The aperiodic signals exhibit a 1/f power spectra and are scale free. Additionally, the aperiodic components constitute a significant portion of the spontaneous electrical field potentials of EEG recordings~\citep{raichle2006brain, raichle2010two}.
The aperiodic signals have been hypothesized to depend on global changes in the excitation/inhibition (E/I) balance~\citep{gao2017inferring} while oscillatory components are related to population asynchrony and facilitate the dynamic temporal and spatial organization of neuronal activity~\citep{donoghue2022methodological, voytek2015dynamic}.
The aperiodic components of EEG were found to vary according to age, cognitive states and task demands~\citep{donoghue2020parameterizing}, have been shown to act as potential biomarkers for neurological disorders such as Depression, ADHD and Parkinsons~\citep{pani2022clinical}. 
While prior works have established correlations between aperiodic components capturing subject specific signatures and oscillatory components primarily capturing task specific signatures, the relation between these observations and impact on EEG foundation models has been under-explored.

\section{Spectral Bias in Reconstruction Based EEG SSL Models}
\label{sec:eeg_reconstruction_based}
Masked reconstruction has been the primary pretext task for training SSL models on EEG signals, on patches of raw EEG data~\citep{wang2025cbramod, zhou2025csbrain} or through discrete tokenization~\citep{wang2024eegpt, jiang2024large}.
While masked reconstruction based pretext tasks have proven to be effective at learning generalized representations in images and speech modalities, it is not inherently clear whether reconstruction only objectives would lead to generalized representations in EEG that consistently outperform supervised baselines. 
\paragraph{Problem Setup}
Let $\mathbf{X} \in \mathbb{R}^{C \times T}$ be an EEG epoch over $C$ channels and
$T$ timesteps, vectorized as $\mathbf{x} = \mathrm{vec}(\mathbf{X}) \in \mathbb{R}^{CT}$.
We model the signal as:
\begin{equation}
    \mathbf{x} = (\mathbf{A} \otimes \mathbf{I}_T )(\mathbf{z}_{\text{ap}} +
    \mathbf{z}_{\text{osc}}) + \boldsymbol{\varepsilon}
    \label{eq:forward}
\end{equation}
where $\mathbf{A} \in \mathbb{R}^{C \times N_s}$ is the leadfield (forward model) matrix that linearly maps 
$N_s$ neural sources into the $C$  EEG sensor space, $\mathbf{z}_{\text{ap}}$ and $\mathbf{z}_{\text{osc}}$
are the vectorized aperiodic and oscillatory source signals respectively, and
$\boldsymbol{\varepsilon} \sim \mathcal{N}(\mathbf{0}, \Sigma_\varepsilon)$.

Let $\mathbf{m} \in \{0,1\}^{CT}$ be a binary masking vector where each entry is drawn
independently, with $m_i = 0$ indicating that index $i$ is masked. The masked
observation is defined as $\tilde{\mathbf{x}} = \mathbf{m} \odot \mathbf{x}$, where
$\odot$ denotes the Hadamard product. Let $f_\theta : \mathbb{R}^{CT} \to \mathbb{R}^{CT}$
and $g_\phi : \mathbb{R}^{CT} \to \mathbb{R}^{CT}$ be encoder and decoder modules respectively, both
being overparameterized. The reconstruction objective is:
\begin{equation}
    \mathcal{L}(\theta, \phi) = \mathbb{E}_{\mathbf{x}, \mathbf{m}}\bigl[\|(1 - \mathbf{m}) \odot (\mathbf{x} - g_\phi(f_\theta(\tilde{\mathbf{x}})))\|^2\bigr]
\end{equation}
where the loss is computed only over the masked indices. Assuming $\mathbf{z}_{\text{ap}}$
and $\mathbf{z}_{\text{osc}}$ are mutually independent, the covariance matrix of
$\mathbf{x}$ factorizes as:
\begin{equation}
    \Sigma_\mathbf{x} = (\mathbf{A} \otimes \mathbf{I}_T)\bigl(\Sigma_{\text{ap}} + \Sigma_{\text{osc}}\bigr)(\mathbf{A} \otimes \mathbf{I}_T)^\top + \Sigma_\varepsilon
\end{equation}
where $\Sigma_{\text{ap}} = \mathrm{Cov}(\mathbf{z}_{\text{ap}})$ and
$\Sigma_{\text{osc}} = \mathrm{Cov}(\mathbf{z}_{\text{osc}})$. A well-established
empirical observation in EEG is that the aperiodic component dominates the signal
power~\citep{donoghue2020parameterizing}, i.e.:
\begin{equation}
    \mathrm{tr}(\Sigma_{\text{ap}}) \gg \mathrm{tr}(\Sigma_{\text{osc}})
\end{equation}
so that $\Sigma_\mathbf{x}$ is dominated by the aperiodic term
$(\mathbf{A} \otimes \mathbf{I}_T)\Sigma_{\text{ap}}(\mathbf{A} \otimes \mathbf{I}_T)^\top$.
Furthermore, the aperiodic component is characterized by a $1/f^\beta$ power spectral
density~\citep{donoghue2020parameterizing}, concentrating its energy predominantly in
lower frequencies. In contrast, the oscillatory component $\mathbf{z}_{\text{osc}}$ carries task-relevant neural information that is spatially localized and contributes lesser to signal variance.
Alongside the idiosyncratic characteristics of EEG, spectral bias of
reconstruction-based neural networks in learning lower-frequency components at a faster rate than
higher-frequency ones~\citep{pmlr-v97-rahaman19a, xu2020frequency} has been identified in prior works. This frequency
prioritization has also been observed in masked reconstruction frameworks,
where the model preferentially captures low-frequency structure to minimize the
dominant terms of the reconstruction loss~\citep{zhang2022mask}. Since $\Sigma_\mathbf{x}$ is dominated by the low-frequency aperiodic component, the
gradients of model parameters will be overwhelmingly determined by the
aperiodic variance. The oscillatory signals, being spatially localized and of
comparatively low power, could contribute only weakly to the reconstruction loss.
\paragraph{Hypothesis}
Masked reconstruction models trained on EEG will predominantly learn representations of low-frequency aperiodic structure, and will fail to reliably capture the spatially localized, higher-frequency oscillatory components that are most informative for BCI downstream tasks.
\par 
In order to validate the stated hypothesis, we adopt a two-fold approach. 
In Section~\ref{sec:synthetic_data}, through controlled synthetic EEG data generation (as shown in Fig~\ref{fig:schematic}B), the linear decodability of foundation models to both aperiodic and oscillatory variables is examined. 
Following this, empirical results (linear probe experiments as shown in Fig~\ref{fig:schematic}C) on real world EEG BCI tasks are provided in Section~\ref{sec:empirical_bci_results} on representative BCI tasks for the model embeddings ability to decode both trait-like (subject-specific information) and state-like (task-specific information).
\section{Models}
In order to analyze reconstruction based EEG foundation models on the synthetic and real-world EEG datasets, we pick three representative models that have demonstrated robust downstream performance across different tasks, LaBraM~\citep{jiang2024large}, CBraMod~\citep{wang2025cbramod} and CSBrain~\citep{zhou2025csbrain}.
\textbf{CBraMod} is a foundation model trained on 25,000 hours pre-training data that aims to model temporal and spatial characteristics through distinct (criss-cross) attention mechanisms. This model employs a patch-based masked reconstruction scheme for pre-training.
\textbf{CSBrain} an attention-based foundation model for EEG decoding with novel cross-scale spatiotemporal tokenization and structured sparse attention. Pre-trained using masked reconstruction objective.
\textbf{LaBraM} is an EEG foundation model with a pre-training objective based on masked token prediction. An underlying neural tokenizer is trained with large-scale EEG data through patching the EEG time series into tokens.
\section{Synthetic Single Channel EEG Analysis}
\label{sec:synthetic_data}
\subsection{Synthetic Data Generation}
As stated earlier, EEG signals are composed of aperiodic (1/f) and oscillatory components. 
In order to study whether EEG foundation model representations capture each of these components, we generate single-channel EEG signals with controlled variances in both components.
The spectral components of the aperiodic and oscillatory components were modelled as:
\begin{equation}
    S_{ap}(f) = \frac{10^{A_{ap}}}{f^{\beta}} \text{;   } S_{osc}(f) = 10^{A_{osc}}e^\frac{-(f-f_{osc})^2}{2w^2}\text{;    } S(f) = S_{ap}(f) + S_{osc}(f)
\end{equation}
Where $A_{ap}$ is the aperiodic offset, $\beta$ is the aperiodic exponent, $A_{osc}$ is the oscillatory component power above the aperiodic component, $f_{osc}$ is the frequency of the oscillatory component and $w$ corresponds to the bandwidth of the oscillatory component. 
In order to obtain the time series from the frequency component, a uniform random phase is added to the amplitude component followed by an Inverse Fourier Transform. 
Fig.~\ref{fig:schematic}B illustrates the signal generation pipeline.
To study the impact of the aperiodic and oscillatory components individually, other parameters was fixed and multiple samples were generated by varying the parameter of interest. 
For each parameter, 1000 samples were generated through linear sweep of the value corresponding to parameter of interest. Ranges along which each of the parameters were varied for sample generation is reported in Table.~\ref{tab:parameter_sweep}.
% The ranges for each parameter was chosen based on physiologically plausible values.
All samples were generated at 200Hz sampling rate and the length of the signal was 5s.
Refer to Appendix~\ref{sec:synthetic_eeg_appendix} for details regarding sample generation pseudocode and example timeseries samples.
\captionsetup[subfigure]{labelformat=empty}
\begin{figure}[t]
    \centering
    \setlength{\tabcolsep}{2pt}
    
    % --- Column Titles ---
    \makebox[0.3\linewidth]{$\bm{\beta}$} \hfill
    \makebox[0.3\linewidth]{$\bm{A_{\text{ap}}}$} \hfill
    \makebox[0.3\linewidth]{$\bm{f_{\text{osc}}}$} \\ \vspace{0.1em}

    % --- Row 1 (CBraMod) ---
    \makebox[0pt][r]{\raisebox{1.0cm}[0pt][0pt]{\rotatebox[origin=c]{90}{\textbf{CBraMod}}}\hspace{0.6em}}%
    \begin{subfigure}[t]{0.29\linewidth}
        \centering
        \includegraphics[width=\linewidth]{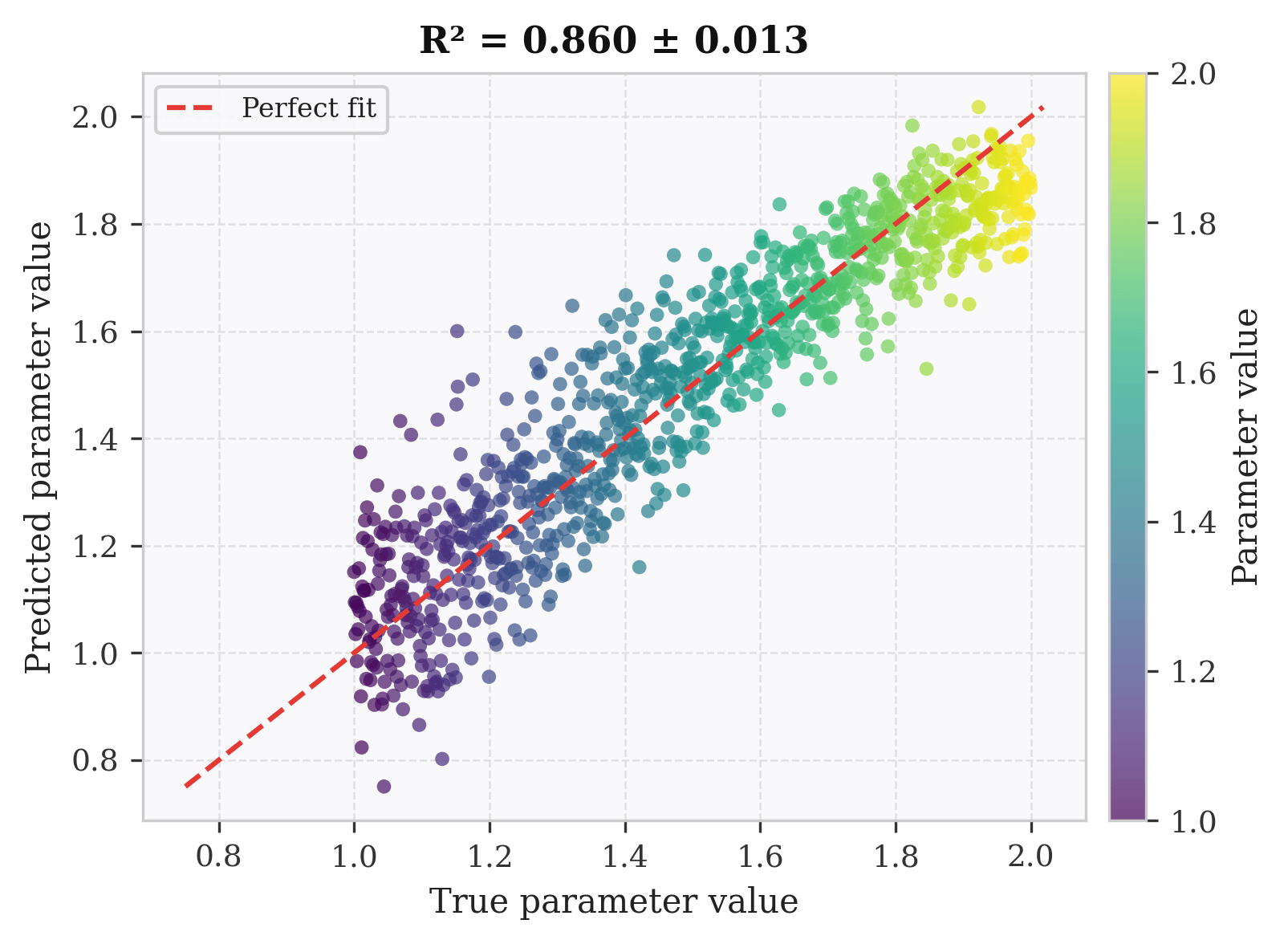}
        \caption{}
    \end{subfigure}
    \hfill
    \begin{subfigure}[t]{0.29\linewidth}
        \centering
        \includegraphics[width=\linewidth]{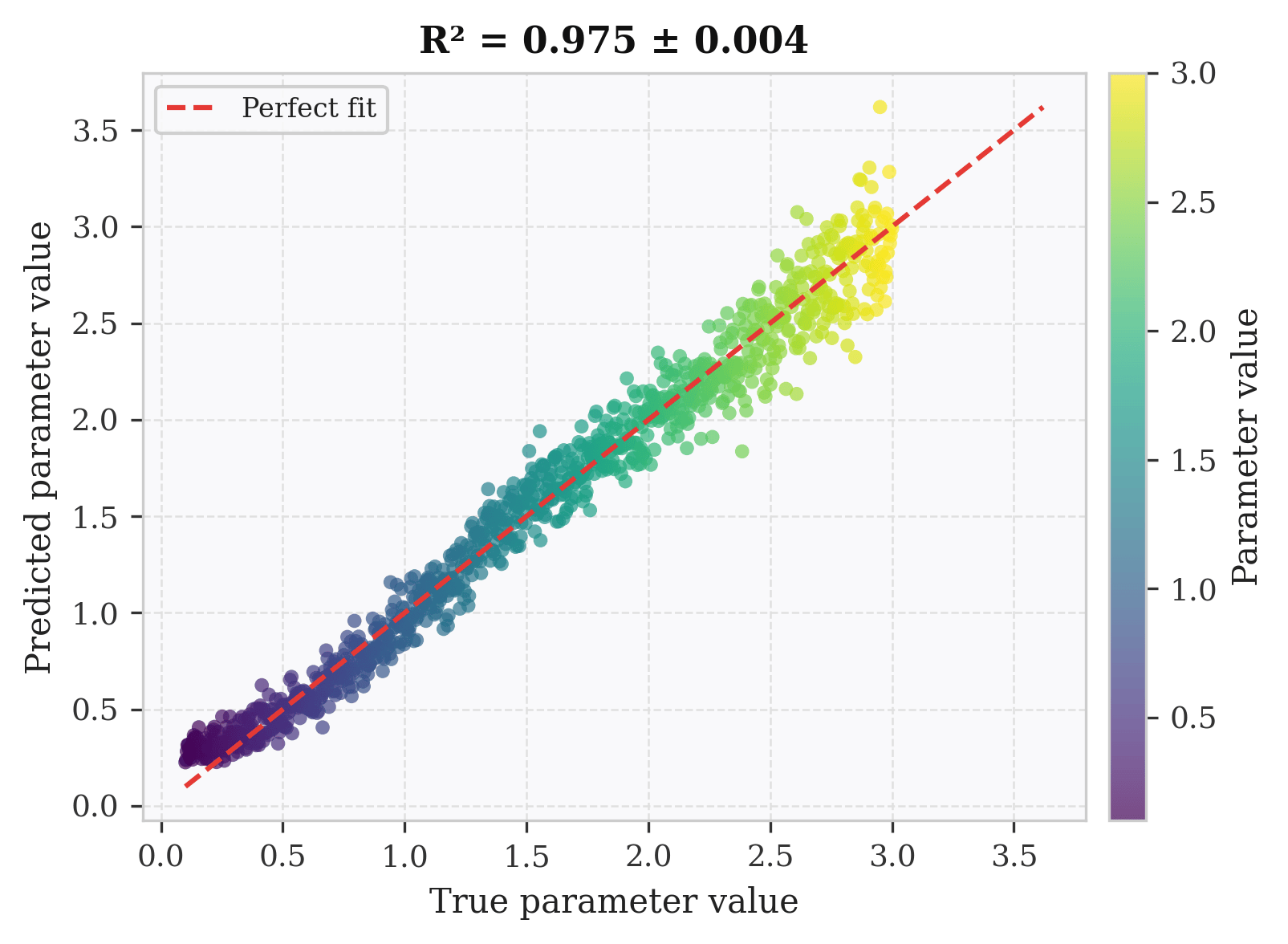}
        \caption{}
    \end{subfigure}
    \hfill
    \begin{subfigure}[t]{0.29\linewidth}
        \centering
        \includegraphics[width=\linewidth]{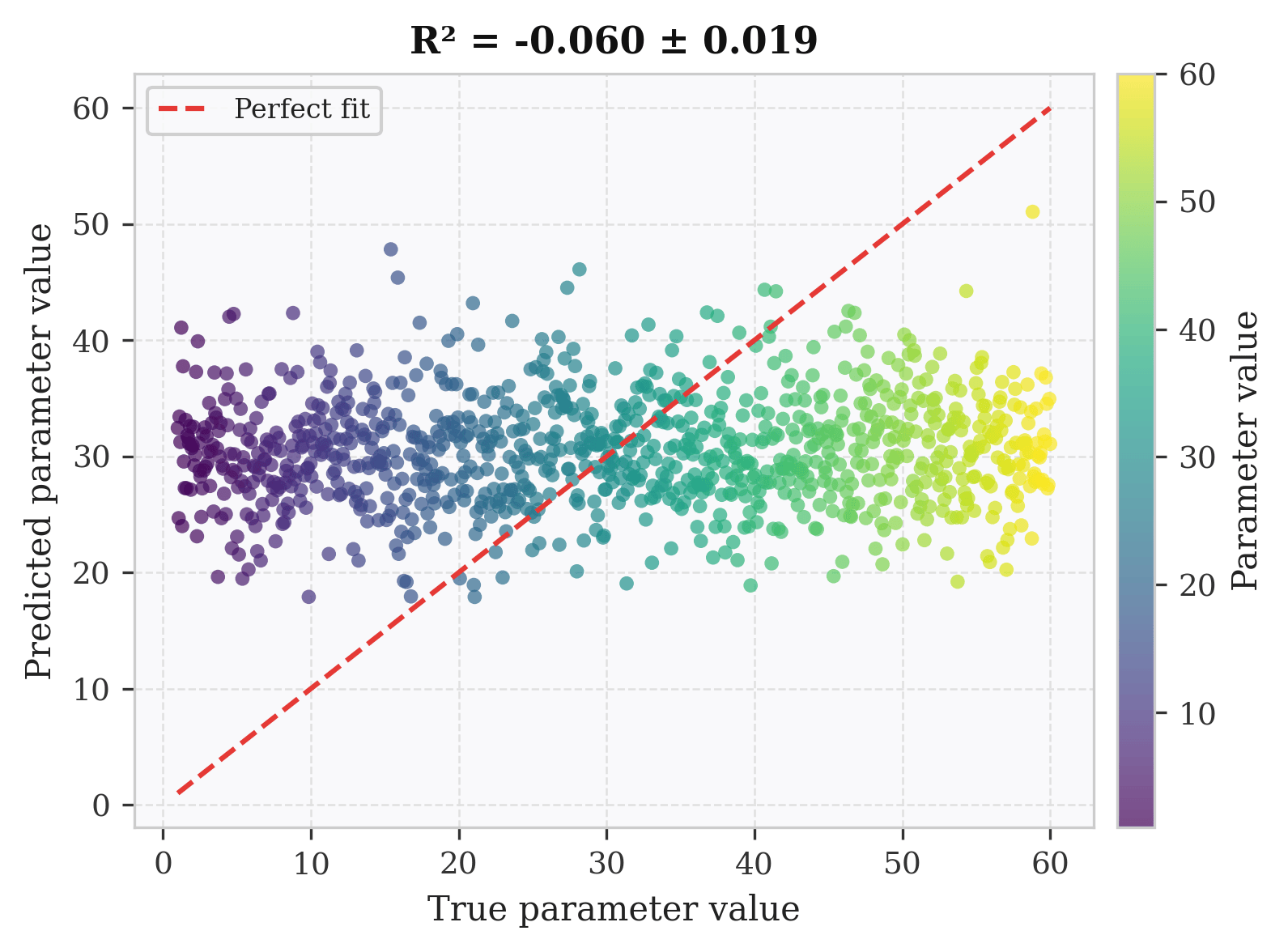}
        \caption{}
    \end{subfigure}
    \vspace{-0.7cm}

    % --- Row 2 (CSBrain) ---
    \makebox[0pt][r]{\raisebox{1.0cm}[0pt][0pt]{\rotatebox[origin=c]{90}{\textbf{CSBrain}}}\hspace{0.6em}}%
    \begin{subfigure}[t]{0.29\linewidth}
        \centering
        \includegraphics[width=\linewidth]{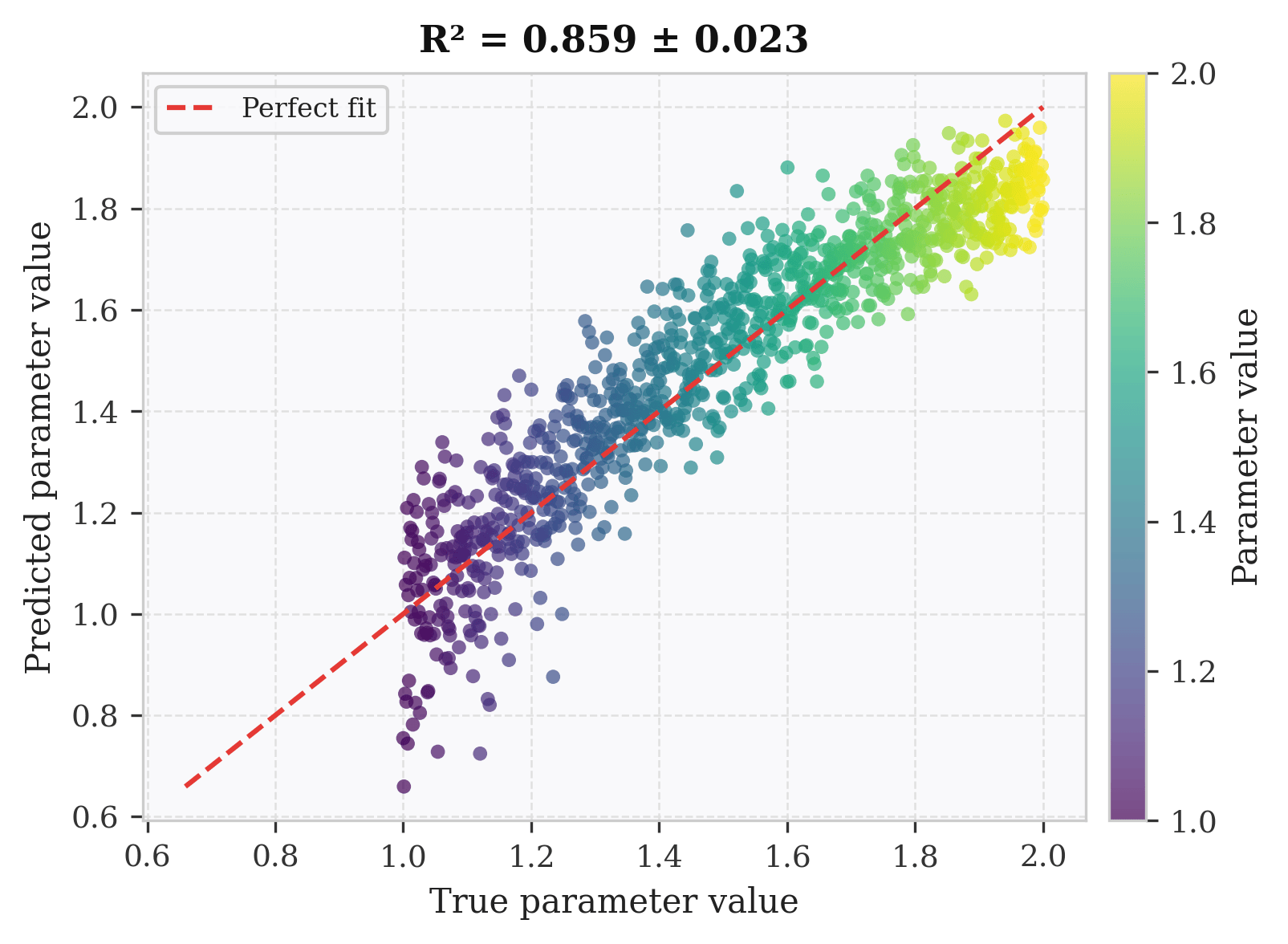}
        \caption{}
    \end{subfigure}
    \hfill
    \begin{subfigure}[t]{0.29\linewidth}
        \centering
        \includegraphics[width=\linewidth]{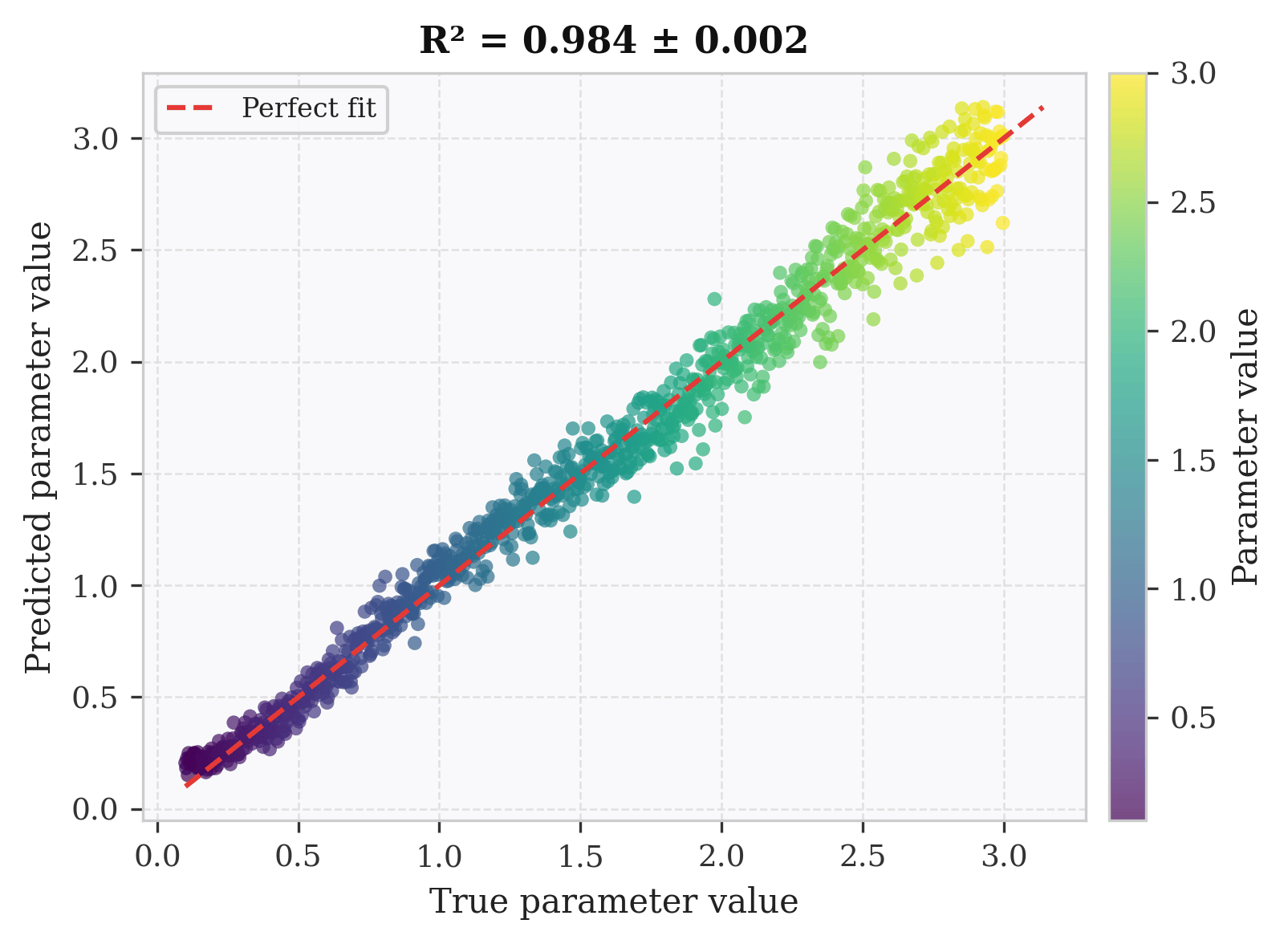}
        \caption{}
    \end{subfigure}
    \hfill
    \begin{subfigure}[t]{0.29\linewidth}
        \centering
        \includegraphics[width=\linewidth]{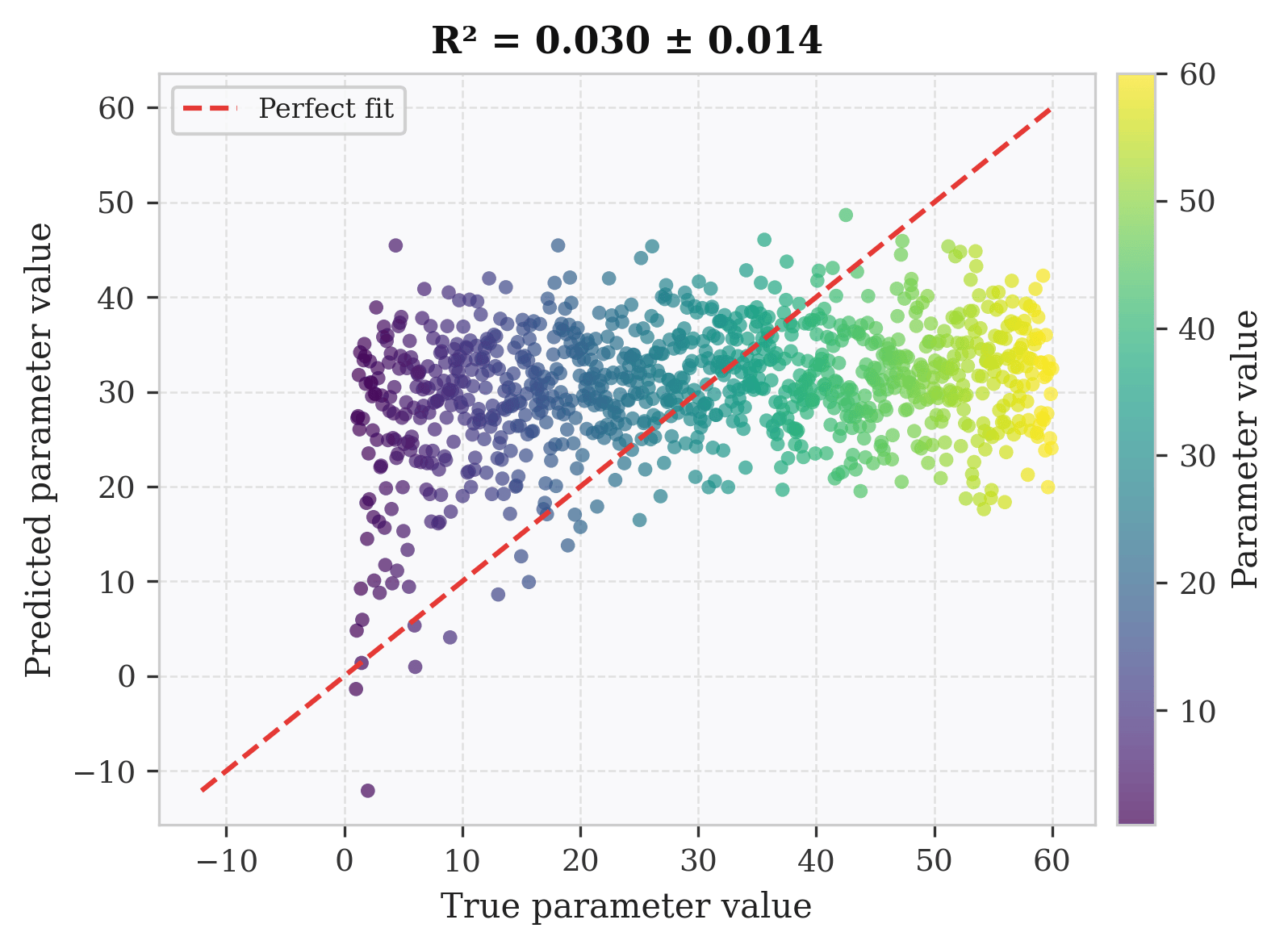}
        \caption{}
    \end{subfigure}
    \vspace{-0.7cm}

    % --- Row 3 (LaBraM) ---
    \makebox[0pt][r]{\raisebox{1.0cm}[0pt][0pt]{\rotatebox[origin=c]{90}{\textbf{LaBraM}}}\hspace{0.6em}}%
    \begin{subfigure}[t]{0.29\linewidth}
        \centering
        \includegraphics[width=\linewidth]{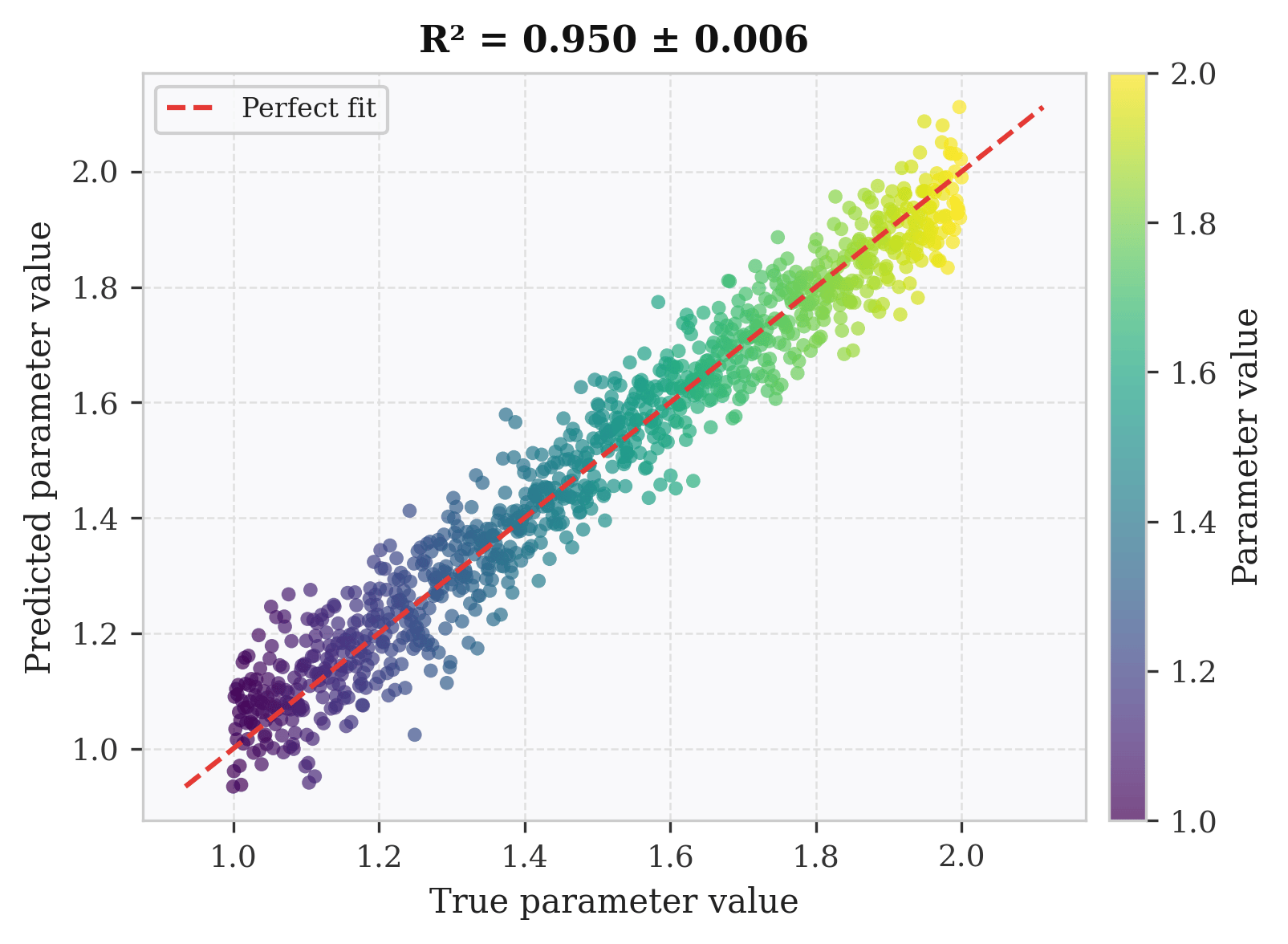}
        \caption{}
    \end{subfigure}
    \hfill
    \begin{subfigure}[t]{0.29\linewidth}
        \centering
        \includegraphics[width=\linewidth]{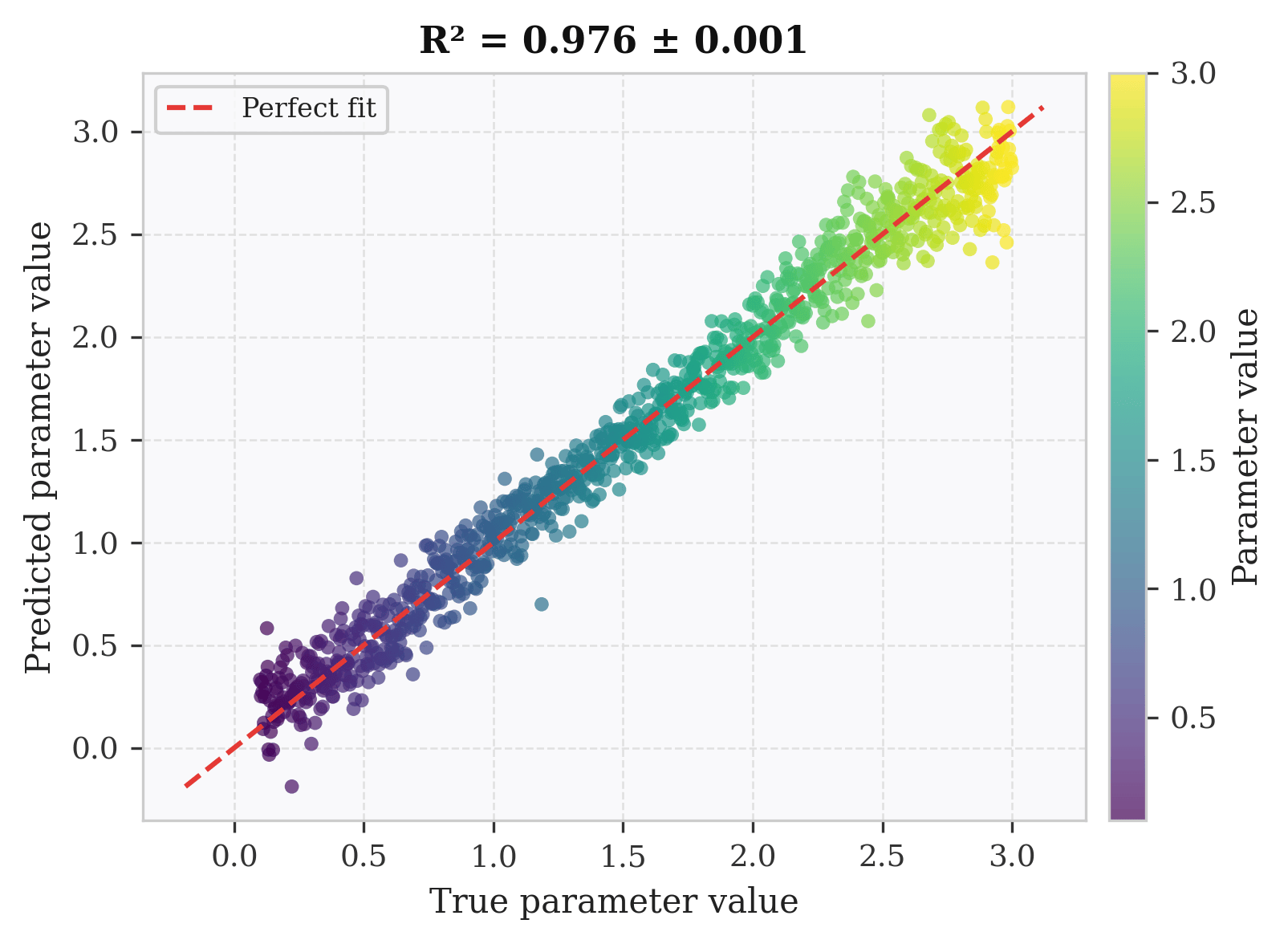}
        \caption{}
    \end{subfigure}
    \hfill
    \begin{subfigure}[t]{0.29\linewidth}
        \centering
        \includegraphics[width=\linewidth]{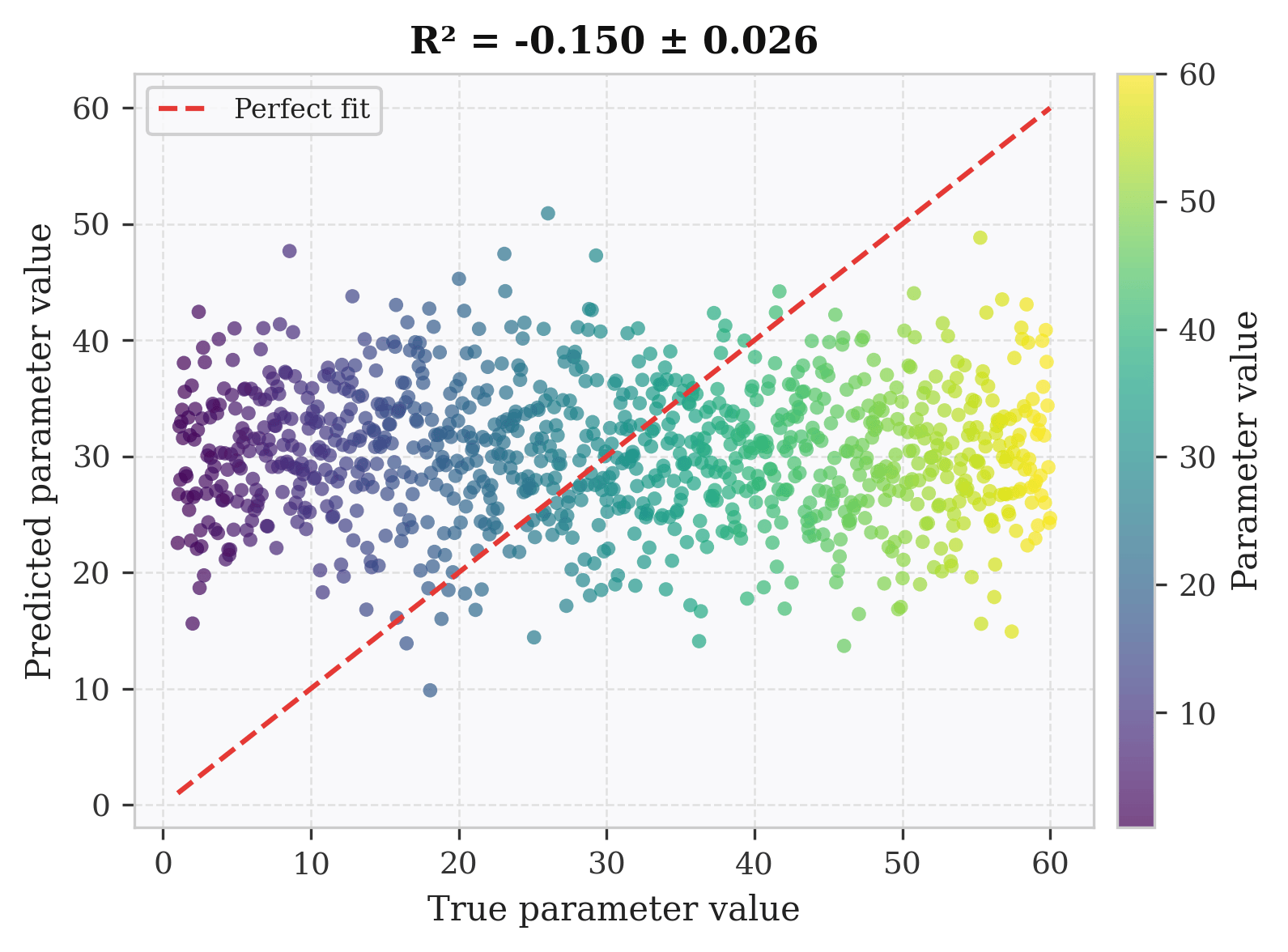}
        \caption{}
    \end{subfigure}
    \vspace{-0.3cm}

    \caption{
    Linear decodability $R^2$ values for Cz channel across three foundation models (CBraMod, CSBrain, LaBraM) for Aperiodic Exponent ($\beta$), Aperiodic Offset ($A_{\text{ap}}$) and Oscillation frequency ($f_{\text{osc}}$). The linear decodability is computed through synthetic single channel EEG with the respective variables varied across the ranges mentioned in Table~\ref{tab:parameter_sweep}. Model representations encode information of aperiodic components whereas the oscillatory frequency is not well captured.
    }
    \label{fig:ap_exp_offset_osc_freq_plot}
\end{figure}
\begin{table}[]
    \centering
    \caption{Range of parameter sweep for synthetic data generation}
    \begin{tabular}{lcccc}
    \toprule
         \textbf{Parameter} & $\beta$ & $A_{ap}$ & $f_{osc}$ & $A_{osc}$  \\
    \midrule
         Range & [1.0, 2.0] & [0.1, 3.0] & [1.0, 60.0] & [0.1, 3.0] \\
    \bottomrule
    \end{tabular}
    \label{tab:parameter_sweep}
\end{table}
\subsection{Linear Decodability of Aperiodic and Oscillatory Components}
Using the generated synthetic samples, embeddings from the last layer of the encoder for three EEG foundation models (LaBraM, CBraMod and CSBrain as shown in Fig.~\ref{fig:schematic}(A)) were extracted (the generated EEG samples were passed as an input corresponding to a particular EEG channel, e.g. Cz, Fz, Pz, Oz).
These embeddings, along with the true variable values used to generate the corresponding signals, are used to train linear regression models using a 5-fold nested cross-validation approach.
The objective of the linear regression models is predict the value of the underlying variable (aperiodic exponent $\beta$, oscillation frequency $f_{osc}$) using the embeddings as inputs. 
The $R^2$ value corresponding to each linear regression model provides a measure of how well the underlying variable can be linearly decoded from the model embeddings. Linear decodability plots for EEG foundation models and three components (aperiodic exponent, aperiodic offset and oscillation frequency) are plotted in Fig~\ref{fig:ap_exp_offset_osc_freq_plot}.
\begin{figure}[t]
    \centering
    
        % --- Column Titles ---
    \makebox[0.19\linewidth]{\textbf{10Hz}} \hfill
    \makebox[0.19\linewidth]{\textbf{20Hz}} \hfill
    \makebox[0.19\linewidth]{\textbf{30Hz}} \hfill
    \makebox[0.19\linewidth]{\textbf{40Hz}} \hfill
    \makebox[0.19\linewidth]{\textbf{50Hz}}\\ \vspace{0.2em}

    \makebox[0pt][r]{\raisebox{0.8cm}[0pt][0pt]{\rotatebox[origin=c]{90}{\textbf{CBraMod}}}\hspace{1em}}%
    \begin{subfigure}[t]{0.19\linewidth}
        \centering
        \includegraphics[width=\linewidth]{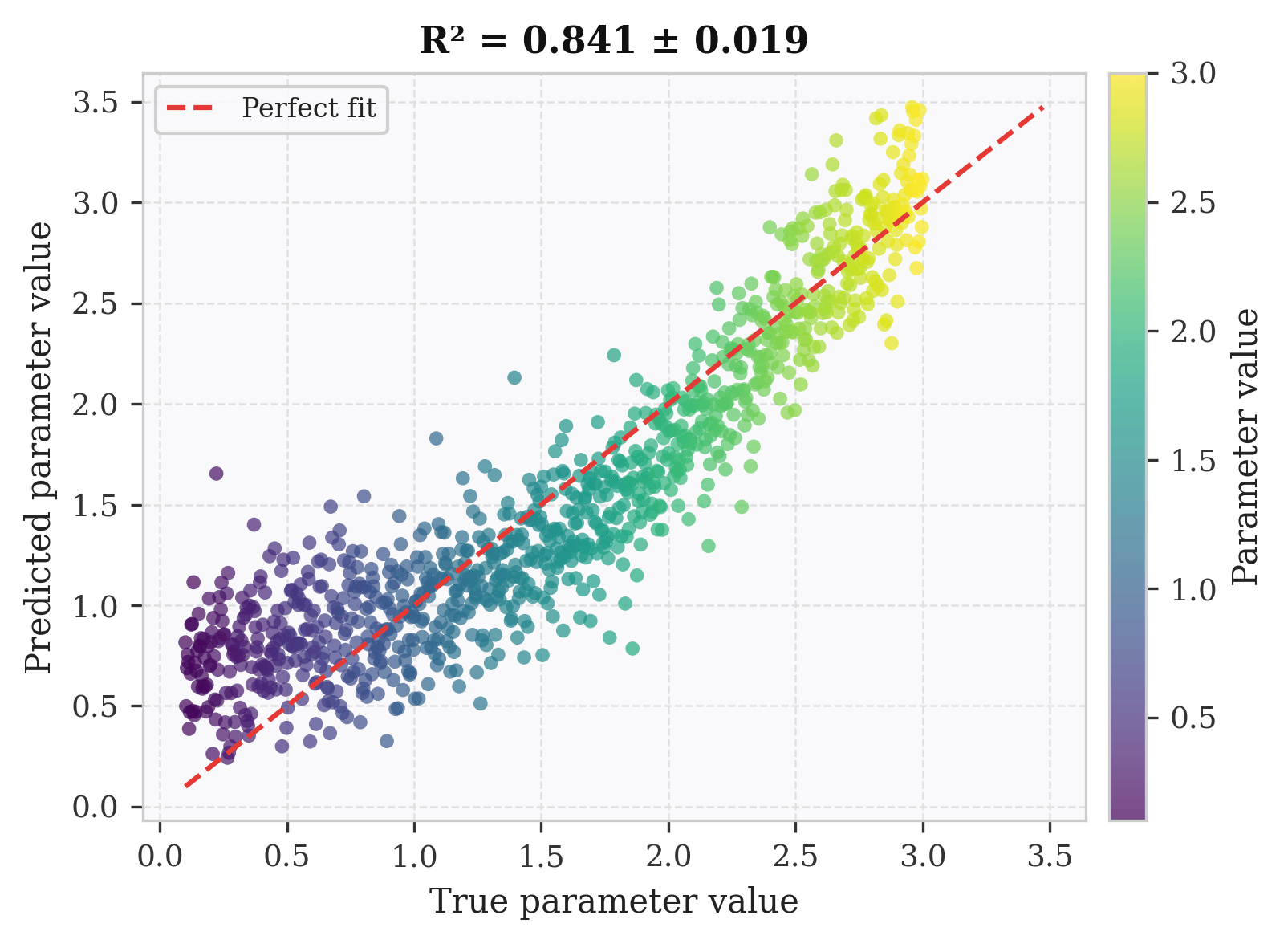}
        \caption{}
    \end{subfigure}%
    \hfill
    \begin{subfigure}[t]{0.19\linewidth}
        \centering
        \includegraphics[width=\linewidth]{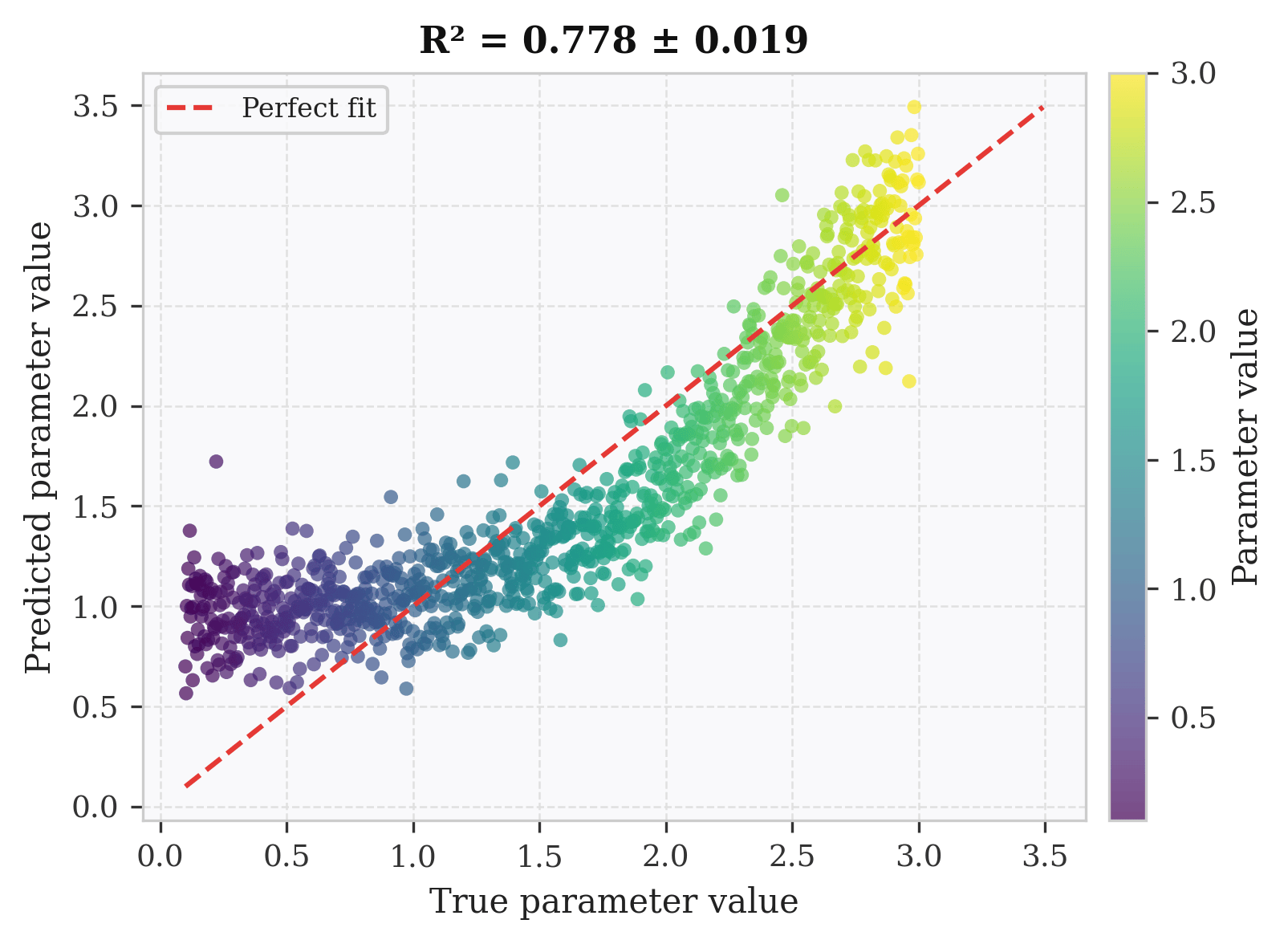}
        \caption{}
    \end{subfigure}%
    \hfill
    \begin{subfigure}[t]{0.19\linewidth}
        \centering
        \includegraphics[width=\linewidth]{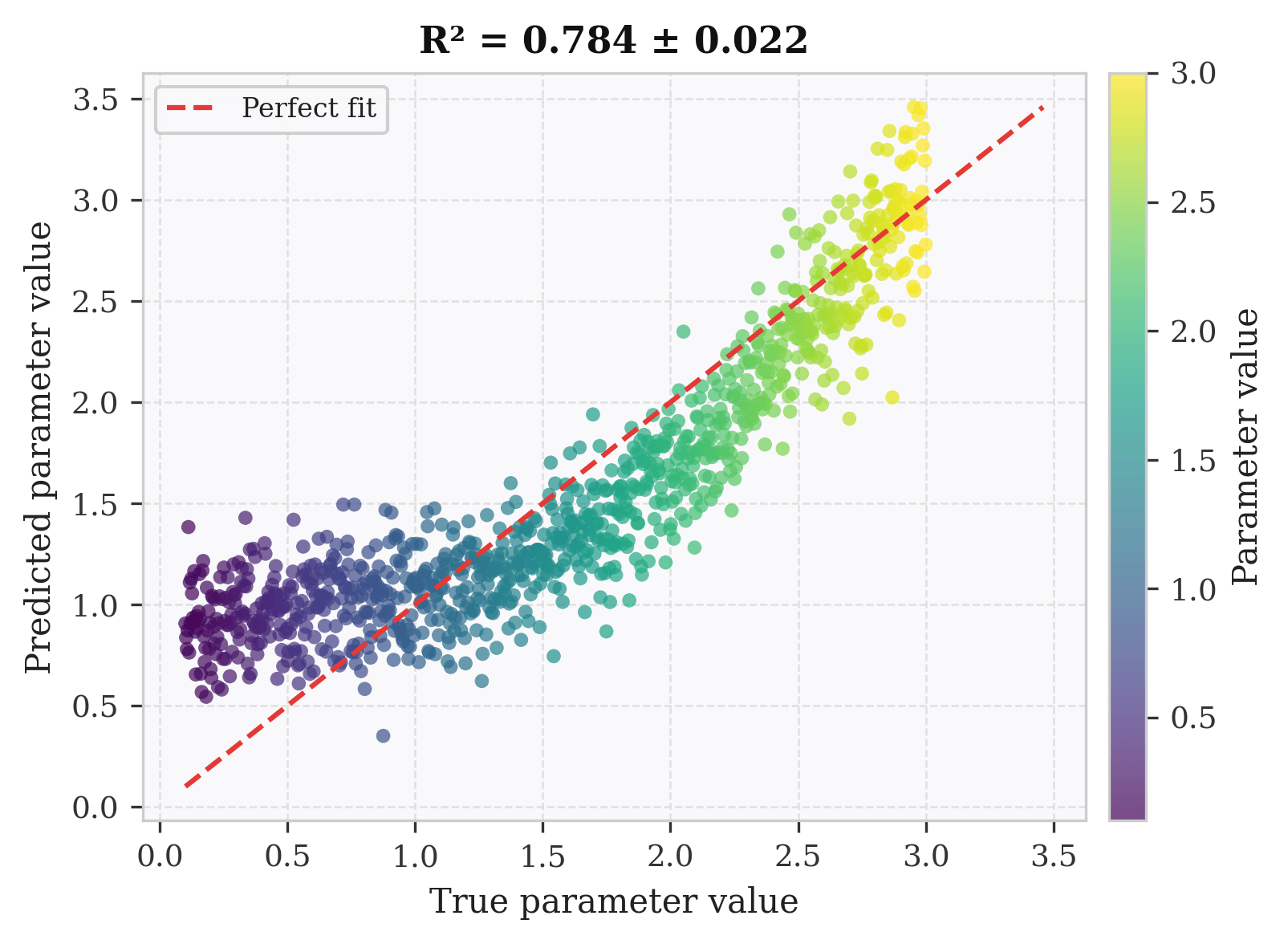}
        \caption{}
    \end{subfigure}%
    \hfill
    \begin{subfigure}[t]{0.19\linewidth}
        \centering
        \includegraphics[width=\linewidth]{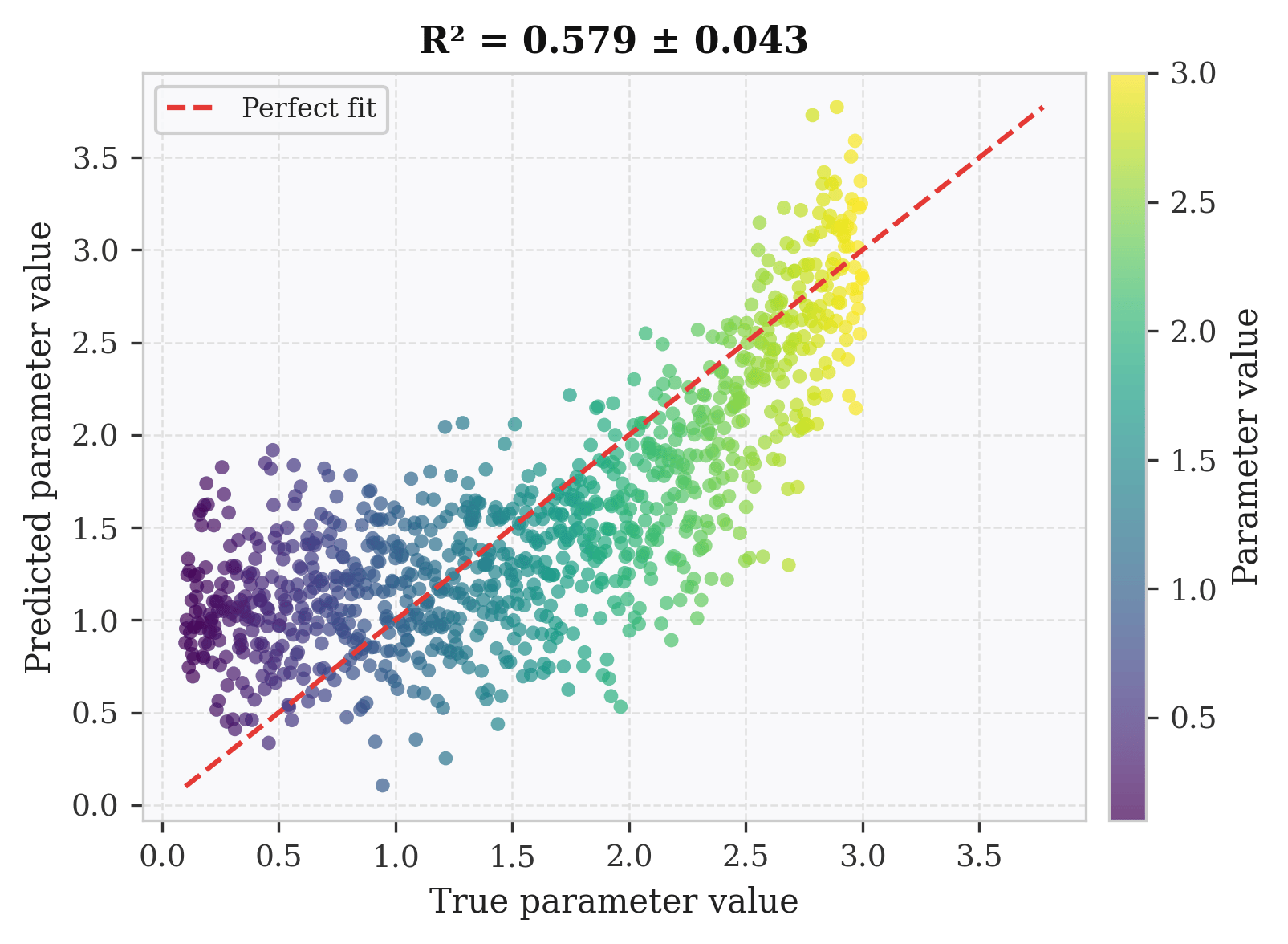}
        \caption{}
    \end{subfigure}
    \hfill
    \begin{subfigure}[t]{0.19\linewidth}
        \centering
        \includegraphics[width=\linewidth]{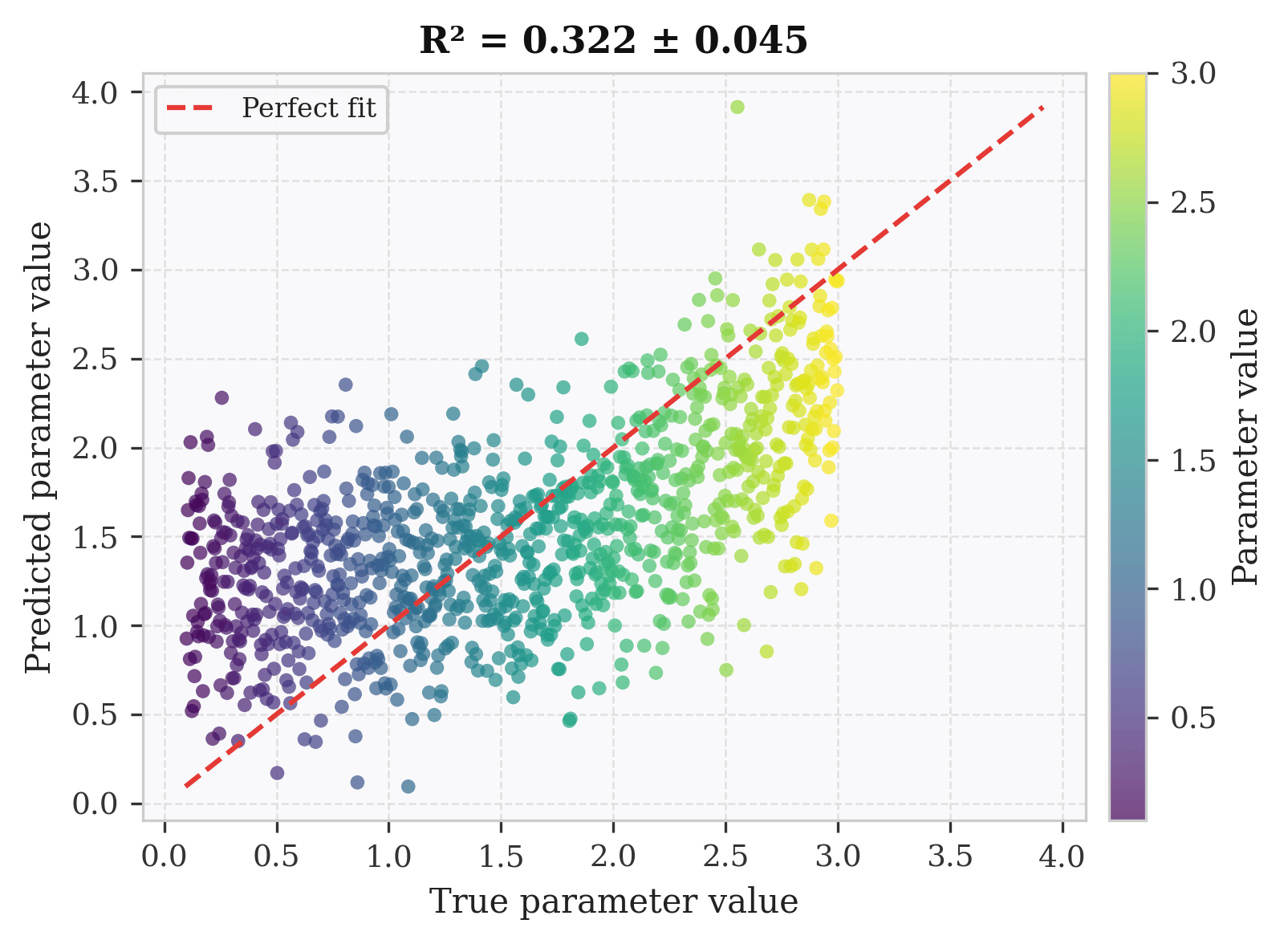}
        \caption{}
    \end{subfigure}

    \vspace{-0.5em}
    
    \makebox[0pt][r]{\raisebox{0.8cm}[0pt][0pt]{\rotatebox[origin=c]{90}{\textbf{CSBrain}}}\hspace{1em}}%
    \begin{subfigure}[t]{0.19\linewidth}
        \centering
        \includegraphics[width=\linewidth]{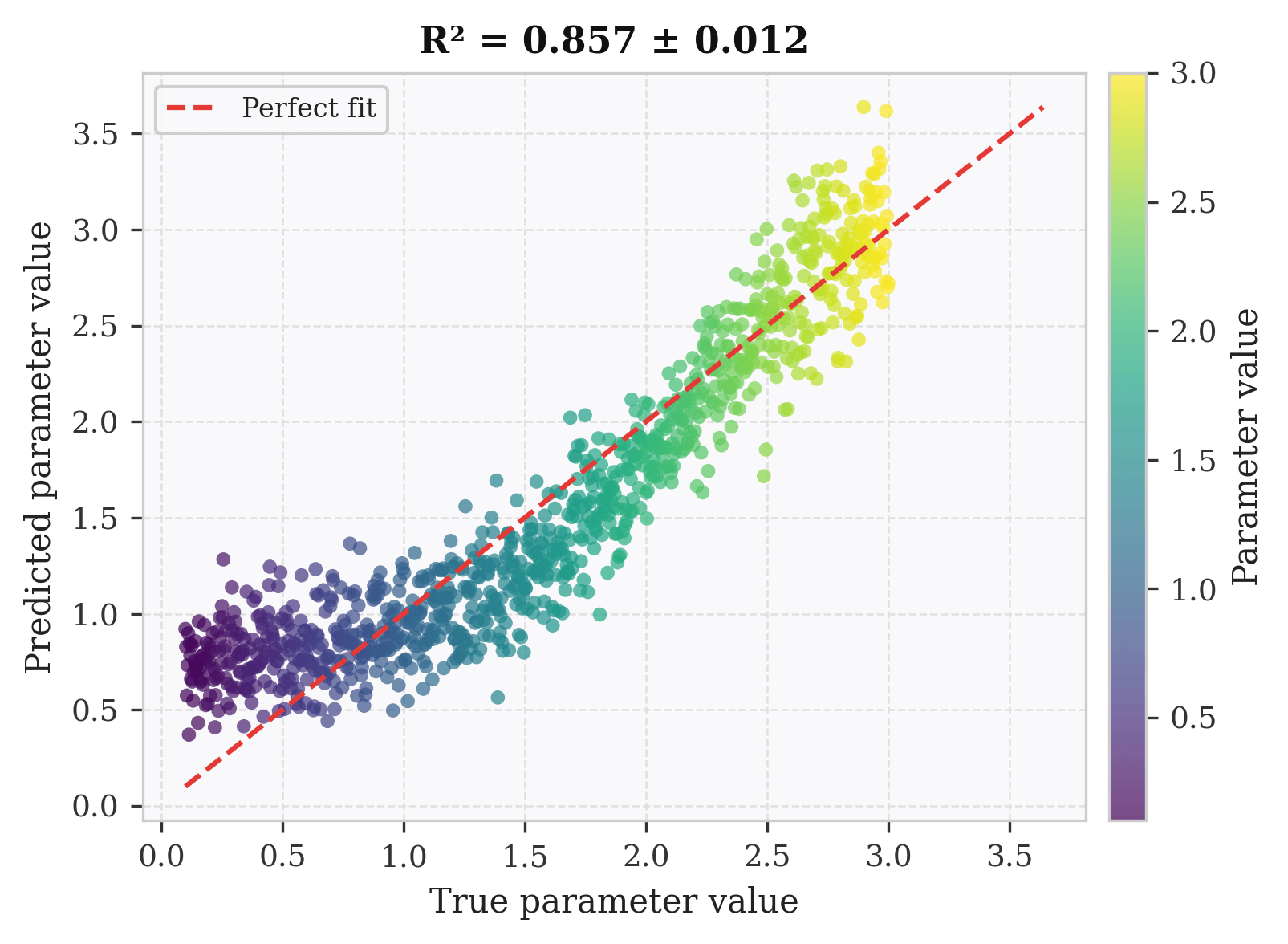}
        \caption{}
    \end{subfigure}%
    \hfill
    \begin{subfigure}[t]{0.19\linewidth}
        \centering
        \includegraphics[width=\linewidth]{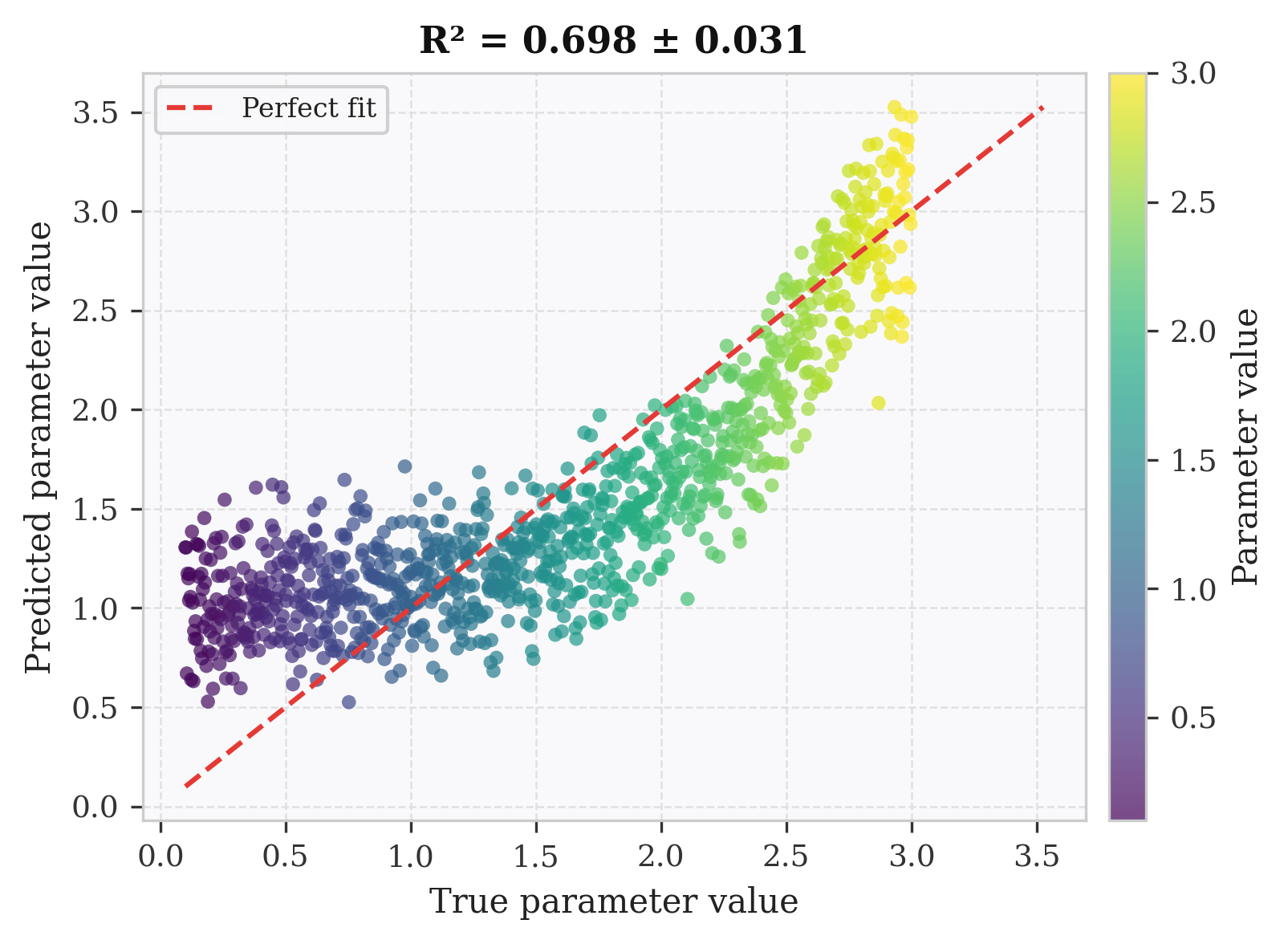}
        \caption{}
    \end{subfigure}%
    \hfill
    \begin{subfigure}[t]{0.19\linewidth}
        \centering
        \includegraphics[width=\linewidth]{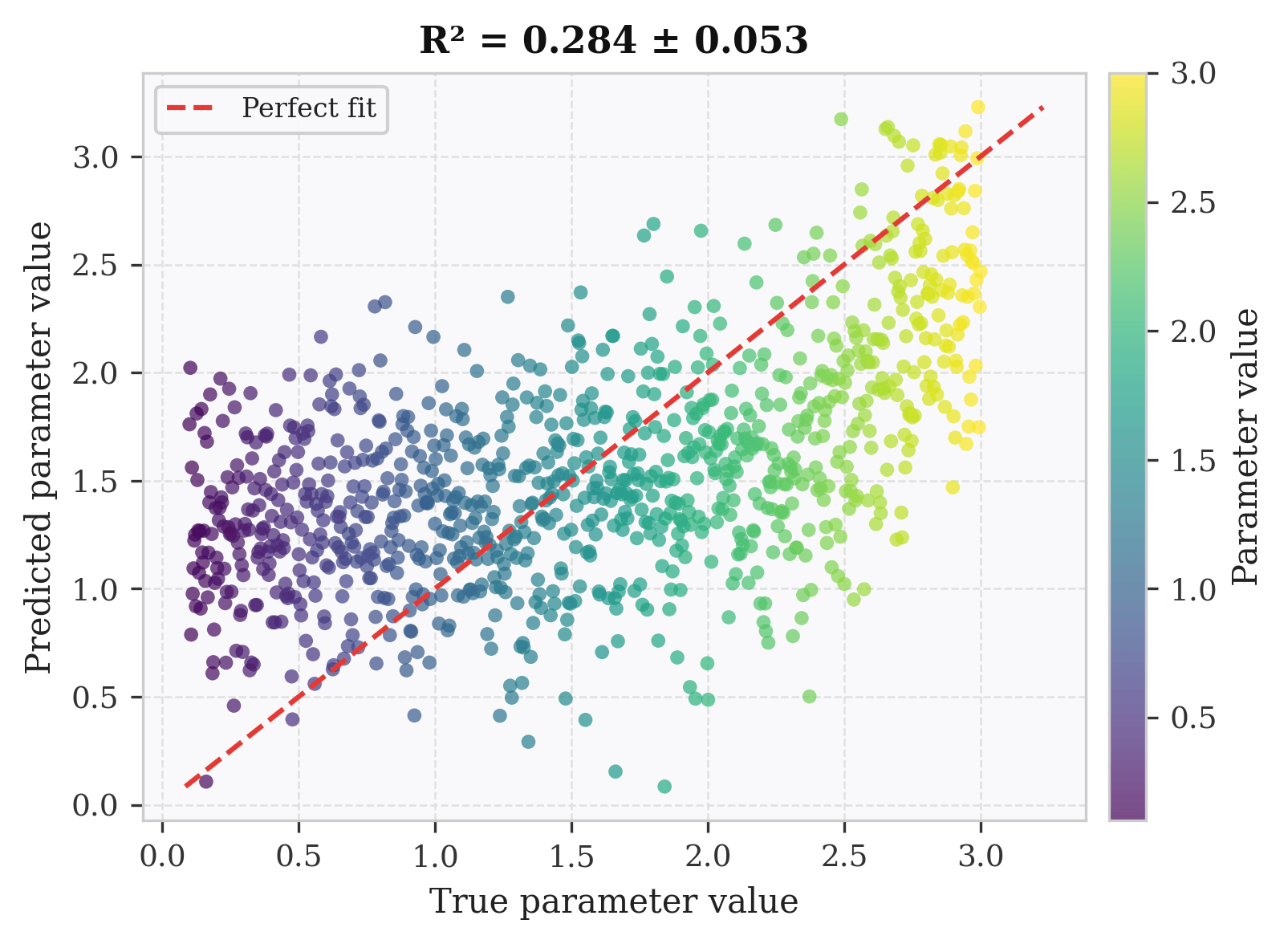}
        \caption{}
    \end{subfigure}%
    \hfill
    \begin{subfigure}[t]{0.19\linewidth}
        \centering
        \includegraphics[width=\linewidth]{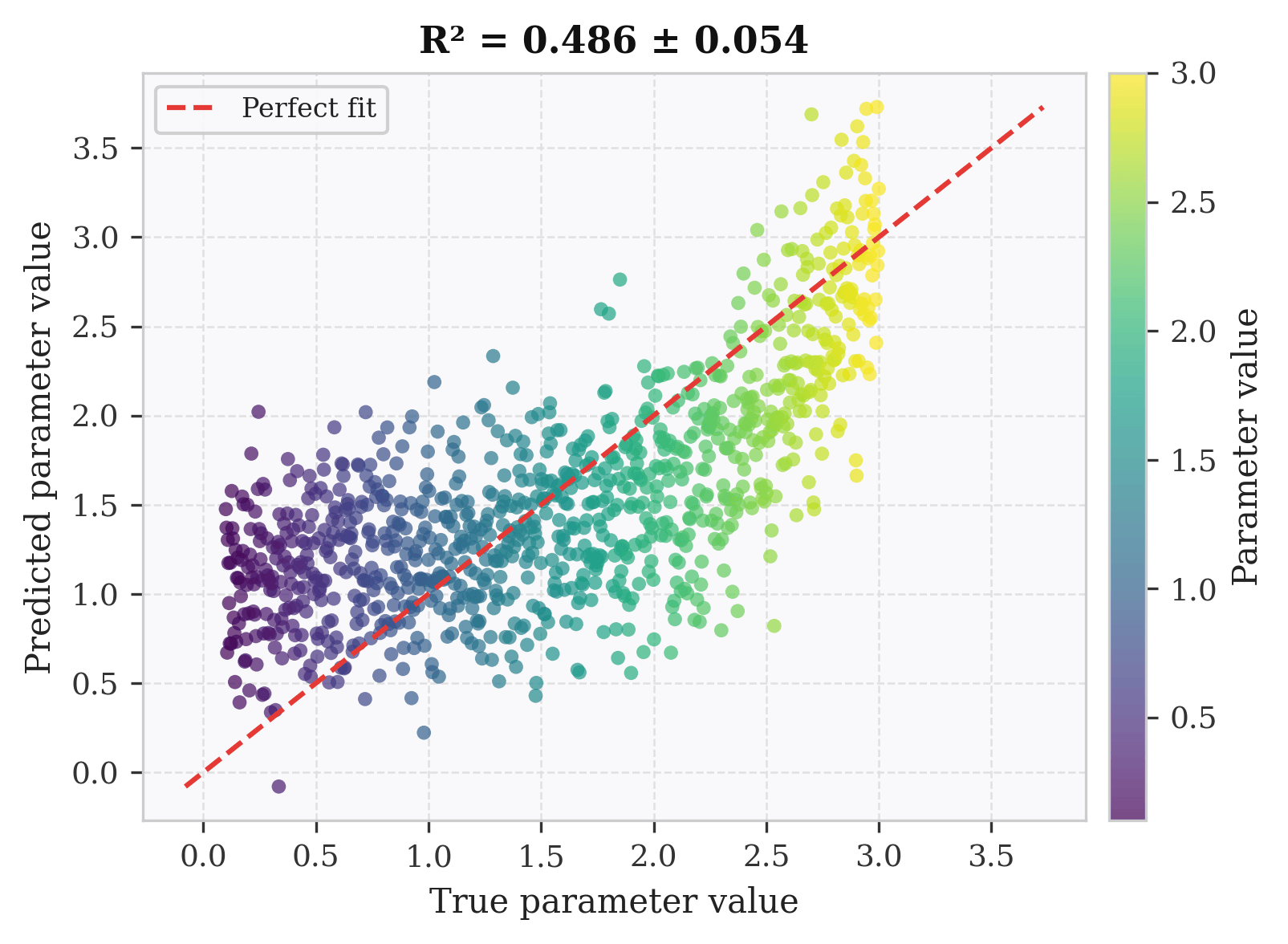}
        \caption{}
    \end{subfigure}
    \hfill
    \begin{subfigure}[t]{0.19\linewidth}
        \centering
        \includegraphics[width=\linewidth]{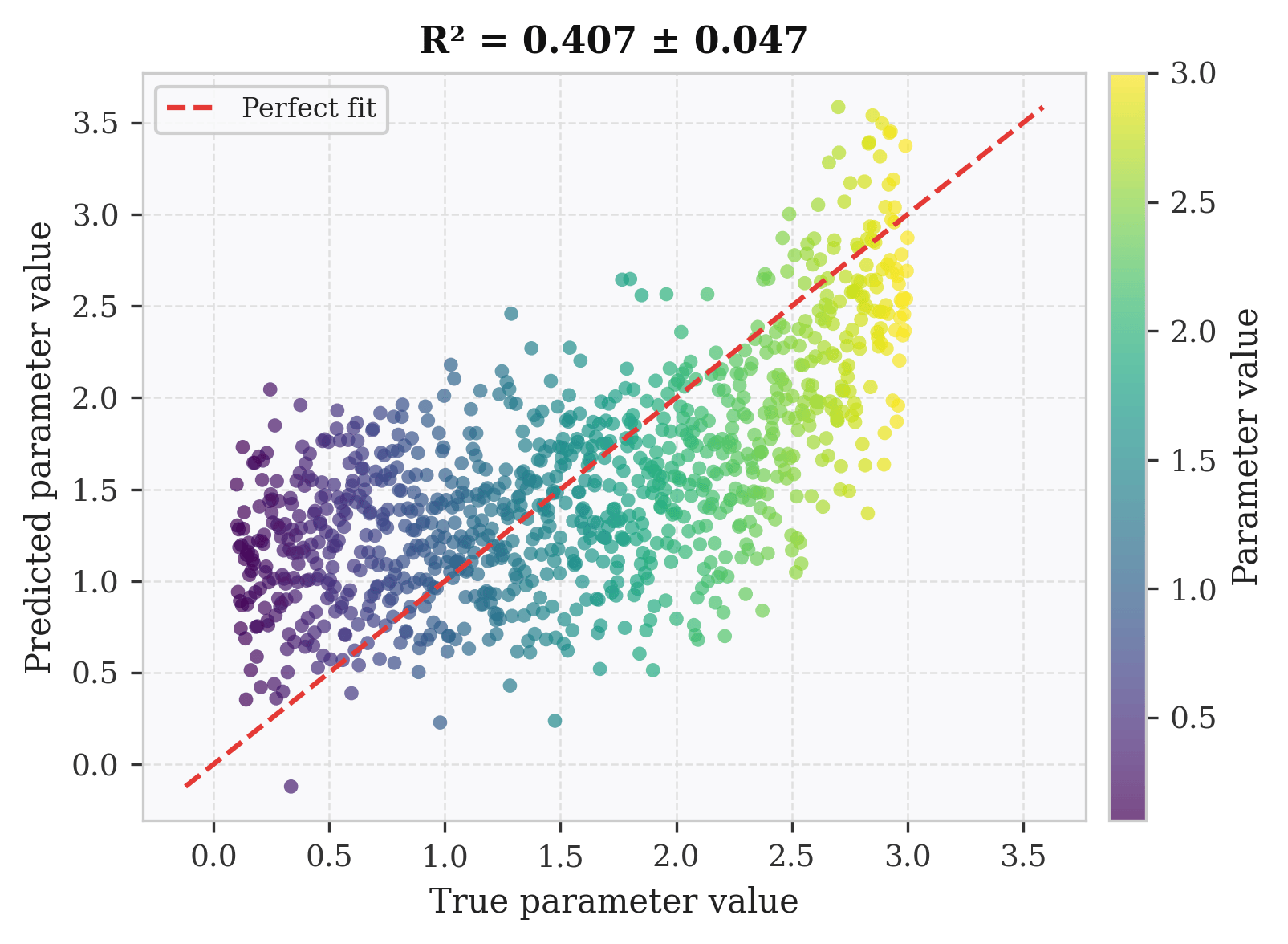}
        \caption{}
    \end{subfigure}

    \vspace{-0.5em}

    \makebox[0pt][r]{\raisebox{0.8cm}[0pt][0pt]{\rotatebox[origin=c]{90}{\textbf{LaBraM}}}\hspace{1em}}%
    \begin{subfigure}[t]{0.19\linewidth}
        \centering
        \includegraphics[width=\linewidth]{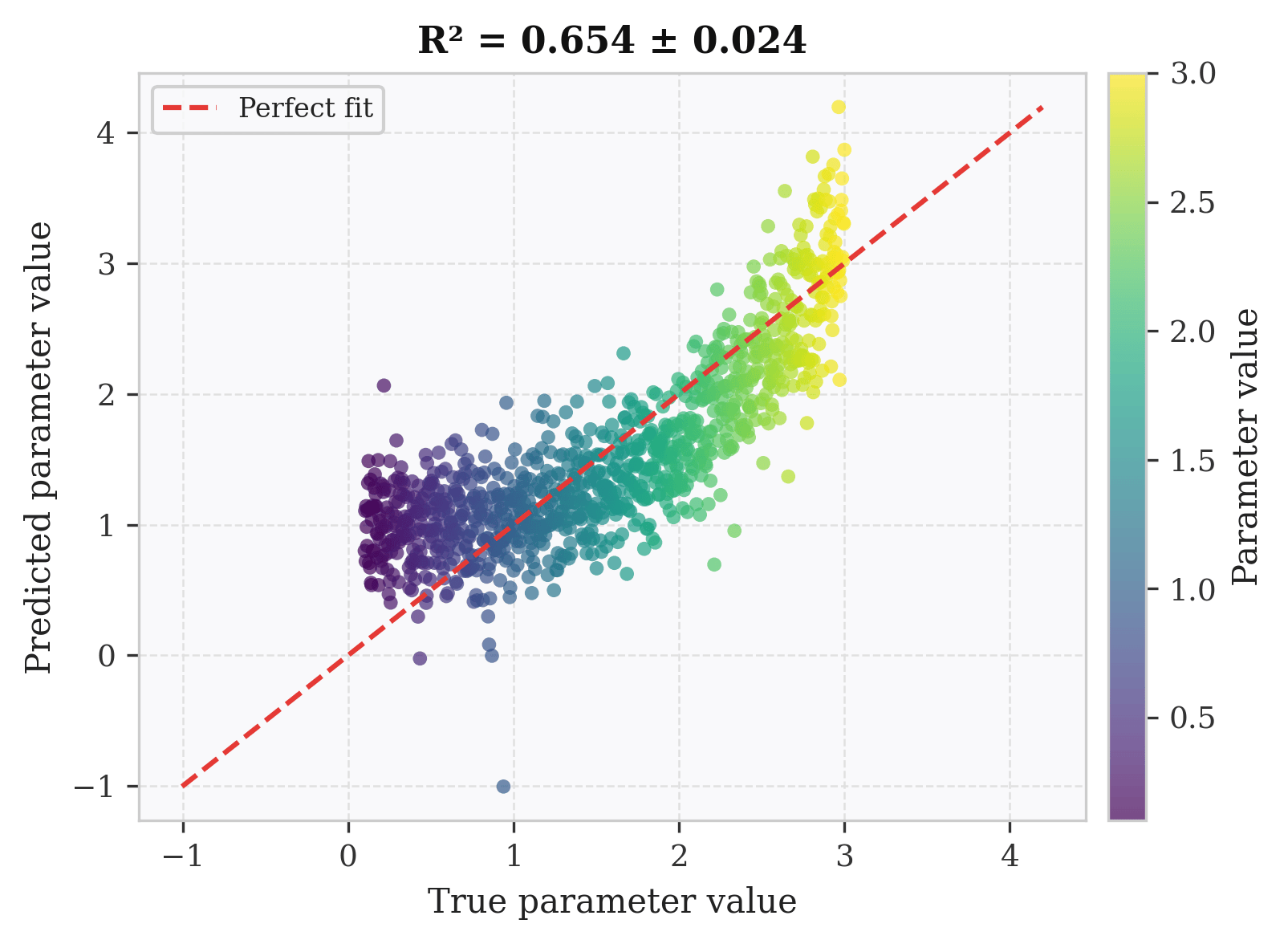}
        \caption{}
    \end{subfigure}%
    \hfill
    \begin{subfigure}[t]{0.19\linewidth}
        \centering
        \includegraphics[width=\linewidth]{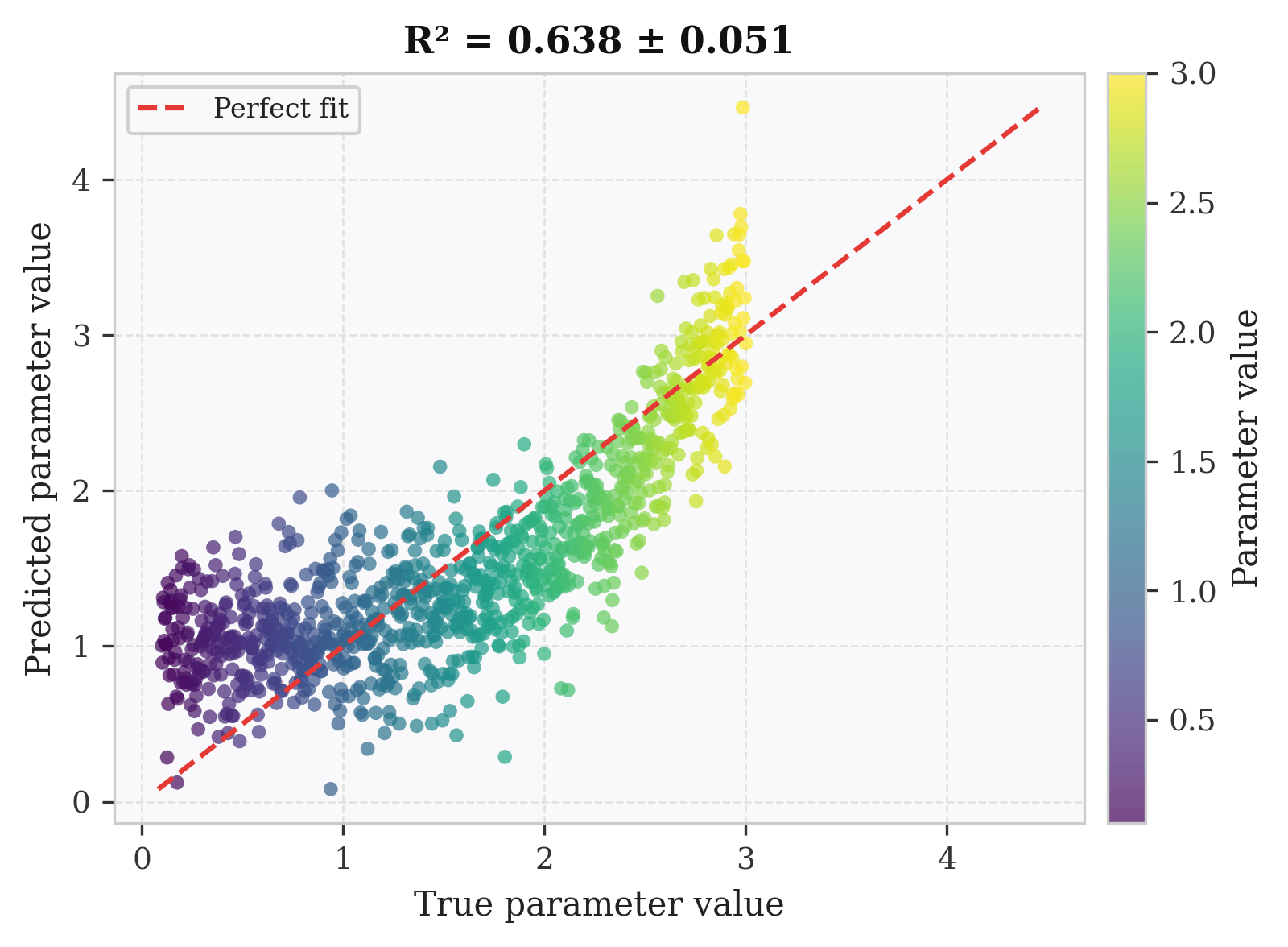}
        \caption{}
    \end{subfigure}%
    \hfill
    \begin{subfigure}[t]{0.19\linewidth}
        \centering
        \includegraphics[width=\linewidth]{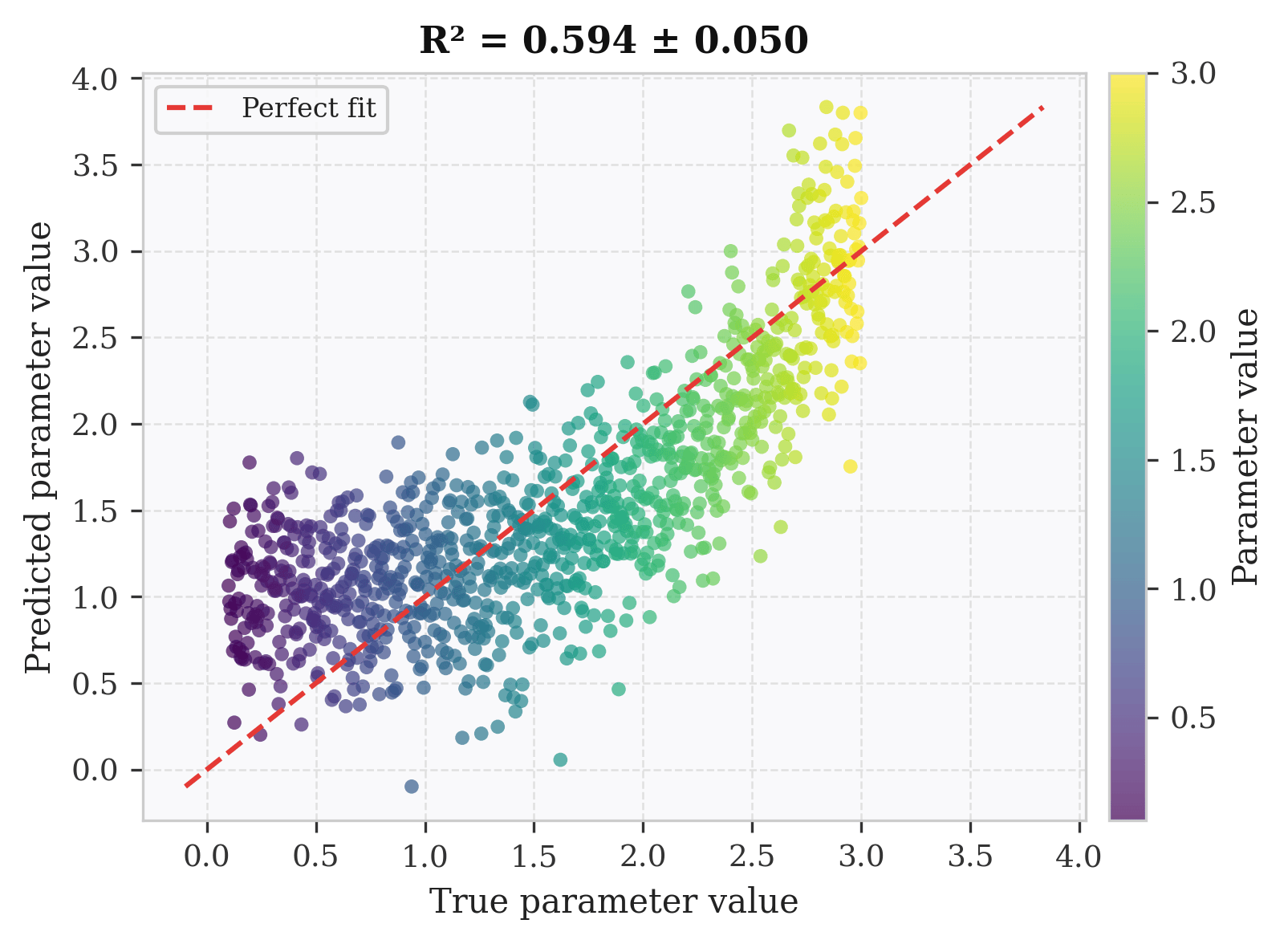}
        \caption{}
    \end{subfigure}%
    \hfill
    \begin{subfigure}[t]{0.19\linewidth}
        \centering
        \includegraphics[width=\linewidth]{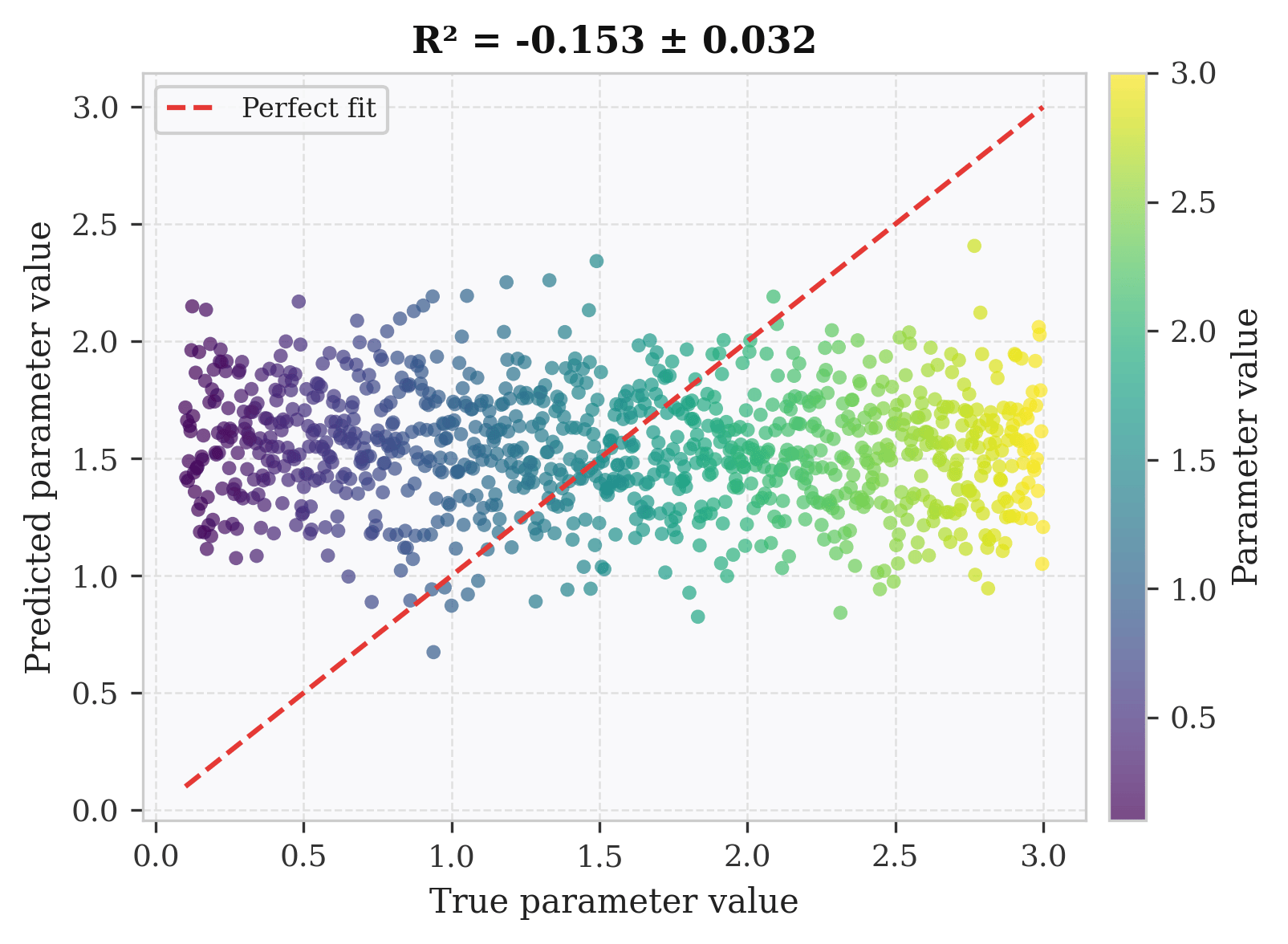}
        \caption{}
    \end{subfigure}
    \hfill
    \begin{subfigure}[t]{0.19\linewidth}
        \centering
        \includegraphics[width=\linewidth]{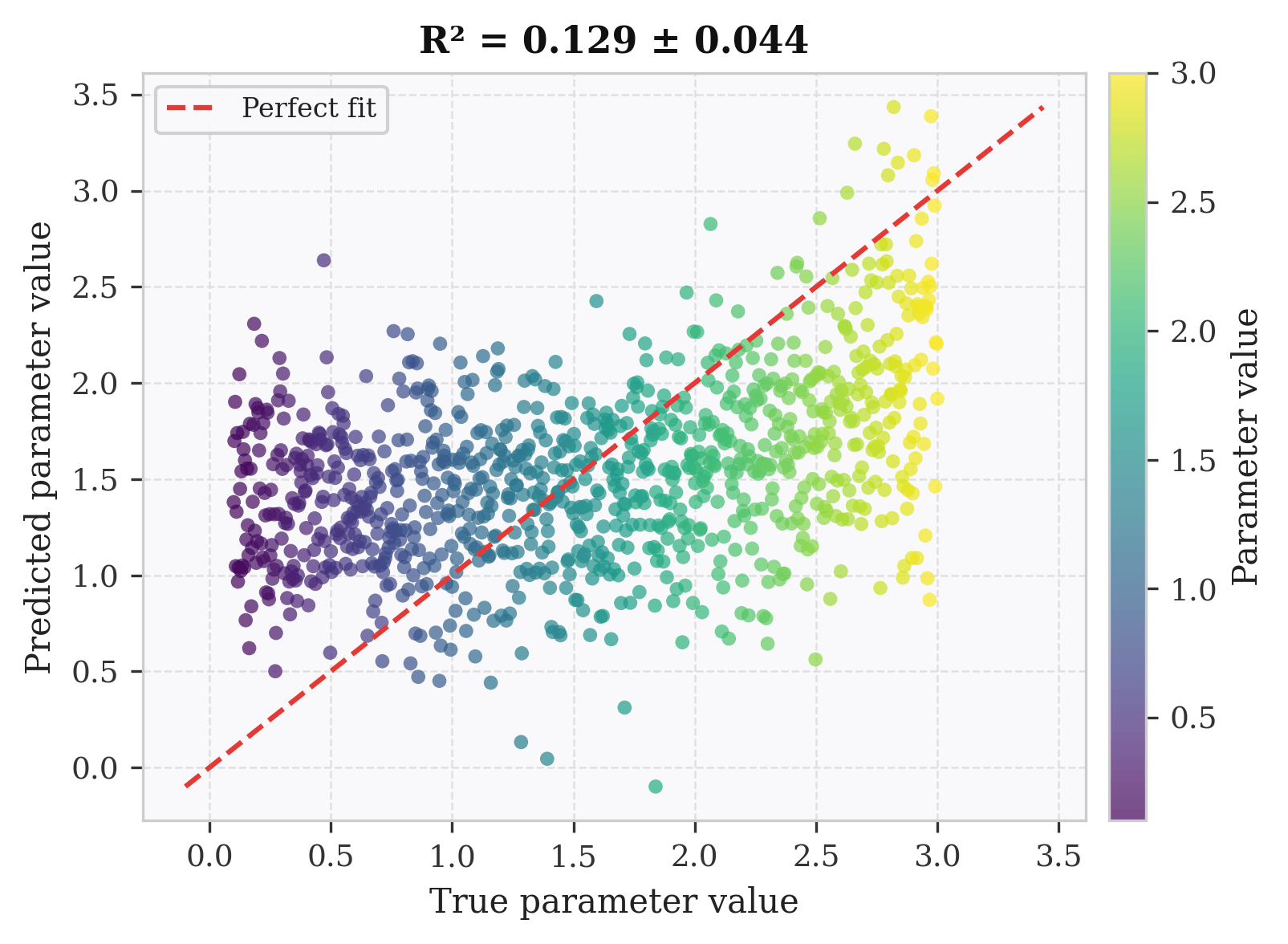}
        \caption{}
    \end{subfigure}

    \caption{
        Linear decodability comparison of Cz channel across oscillatory frequencies ($f_{\text{osc}}$) with varying power of the oscillation ($A_{\text{osc}}$) for CSBrain, LaBraM and CBraMod models. Model sensitivity to predict the oscillatory power value decreases as the oscillatory frequency increases. This indicates that reconstruction based foundation models are less sensitive to modulation in higher frequencies.
    }
    \label{fig:csbrain_oscfreq_power}
\end{figure}
\paragraph{EEG Foundation model embeddings decode aperiodic components} As seen from Fig.~\ref{fig:ap_exp_offset_osc_freq_plot}, across three foundation models, there is a consistent trend wherein the $R^2$ value is high for aperiodic exponent ($\beta$) and offset ($A_{ap}$). 
This indicates that across different foundation models, reconstruction based objectives are able to provide embeddings that allow for linear decodability of aperiodic components of the signals.
However, with respect to the oscillation frequency, we see that the linear regression models are not able to provide robust predictions, indicating that the oscillation frequency is not linearly encoded in the embeddings.
\paragraph{EEG Foundation models exhibit a low-frequency oscillatory bias} While the foundation model embeddings did not provide linear decodability for oscillation frequency values,  they provide robust decoding of oscillatory frequency power ($A_{osc}$) in low frequencies i.e. 10Hz as shown in Fig.~\ref{fig:csbrain_oscfreq_power}, as the frequency of the oscillatory component increases, the model is not able to decode the corresponding power of the oscillatory component.
This could indicate that the embeddings are primarily encoding lower frequency higher-power components, which are the primary components that would lead to lower reconstruction losses owing to the spectral characteristics of EEG signals.
This inability of foundation model embeddings to encode modulation of higher oscillatory component magnitude could help partially explain the sub-optimal performance of EEG foundation models in linear probe setting on BCI tasks, which require encoding beta (13-30 Hz) and gamma ($\geq$30 Hz) frequencies.
While the results have been shown for channel Cz, results for other electrodes are consistent with the stated observations. Results for other channels (Fz, Pz, Oz) have been included in Appendix~\ref{sec:additional_linear_decoding_results}.

\section{BCI Datasets Linear Probe Analysis}
\label{sec:empirical_bci_results}
\subsection{Datasets}
\label{sec:dataset_description}
\paragraph{BCIC-IV 2A~\citep{brunner2008bci}} Motor Imagery Classification dataset with 9 participants across two sessions recorded on different days. Data were collected using 22 EEG channels at 250Hz for four imagined motor movements (left arm, right arm, both feet and tongue). Trials are segmented into 4 second chunks and a single fold with training and evaluation fold according to recording days.
\paragraph{Physionet-MI~\citep{physionet_mi, schalk2004bci2000}} 1500 1-2 minute EEG recordings from 109 participants, for four real and imagined motor tasks (open and close left fist, right fist, both fists and both feet). EEG recordings consist 64 channels and sampled at 160Hz. Each trial was segmented into 4-second windows and 5-fold experiments were performed with 20 randomly sampled subjects per fold. For each subject, the first 8 blocks are considered in training and the last four blocks as evaluation set.
\paragraph{Kaggle-ERN~\citep{inria-bci-challenge}} dataset includes EEG recordings from 26 participants who perform tasks using an online P300 speller interface, and is primarily used to study event-related potentials related to erroneous responses. The EEG data were collected using 56 EEG electrodes and were downsampled to 200 Hz. The classification task is to detect when the selected item is not the intended. The train set consists of first four blocks of all subjects and test comprises of the fifth block.
\paragraph{Sleep EDF~\citep{kemp2000analysis}} contains 197 whole-night PolySomnoGraphic sleep recordings with EEG, EOG and chin EMG recorded at 100Hz and annotated for sleep stages every 30s. In this work, EEG electrodes Fpz-Cz and Pz-Oz are considered for analysis. The sleep stages are annotated for Wake, REM, Movement, NREM-1, NREM-2, NREM-3 and NREM-4. For comparability to prior studies, NREM-3 and NREM-4 have been combined into a single class yielding 5 classes. Randomly samples 20 subjects in a 5-fold cross validation manner is used for evaluation with the training set comprised of the first days data and the testing set containing the second days sleep data.

\subsection{Results}
\par{\textbf{EEG foundation models classify subjects better than tasks in BCI datasets}}
The impact of the identified aperiodic low-frequency spectral bias on downstream tasks is demonstrated through linear probe experiments.
Prior works~\citep{donoghue2020parameterizing} have observed aperiodic components to be correlated to subject identities whereas modulation in oscillatory components are primarily associated with tasks.
To study whether the spectral bias in foundation models leads to embeddings capturing subject identities rather than task specific constructs, two distinct linear probe models are trained, one to predict the task for a given trial and the second to predict the subject identity from trial. 
Appendix~\ref{sec:appendix_hyperparameters} includes hyperparameters and linear probe training details.
We hypothesize that under linear probe setting, EEG foundation model embeddings preserve subject specific information resulting in better subject identification performance compared to task classification.
\par 
Three foundation models (LaBraM, CBraMod and CSBrain) are tested on three BCI datasets (BCIC IV 2A, Kaggle-ERN and Physionet-MI) and one sleep classification dataset (Sleep-EDF).
Sleep-EDF dataset was selected to demonstrate an exemplary task wherein the task-related power is much larger in magnitude compared to BCI tasks and the frequencies of interest are in lower frequencies compared to BCI tasks.
During the dataset split, identical training and testing sets as stated in Sec~\ref{sec:dataset_description} were used for both task classification and subject classification.
In order to provide a normalized metric for subject identification and task classification, Cohen's kappa metric is reported.
\par 
As seen from the results in Table~\ref{tab:eegfm_task_subject_lp_classification}, we can clearly observe that for all the three BCI tasks, the performance on subject identification is noticeably better compared to task performance. 
This observation holds despite the number of classes in task classification being noticeably lower compared to subject identification.
This indicates that the EEG foundation models capture session and subject information to a larger extent compared to task specific information.
This in turn leads to sup-optimal performance observed in linear-probe experiments for between-subjects setting as the models are unable to generalize to unseen subjects~\citep{yang2026arefoundationmodelsworthit, kuruppu2026eeg}.
Additionally, the observed discrepancy between subject identification and task classification is not solely owing to session variables as BCIC IV 2A dataset consists of training data and testing data from distinct days, indicating that the models are able to capture subject specific information over session variables.
\paragraph{EEG foundation models form task-centric representations in Sleep-EDF} While foundation models show a subject centric bias in BCI tasks, results in Sleep-EDF exhibit an opposite trend, wherein the models provide better task performance over subject identification. 
Firstly, unlike BCI tasks, sleep EEG recordings have large low-frequency task relevant components, which helps the foundation models better capture task specific information. 
The observed differences between BCI datasets and Sleep-EDF are further influenced by the EEG montage used. 
BCI datasets employ high-density montages (22–64 channels), enabling models to capture rich spatial patterns and cross-channel covariance structures that are strongly subject-specific~\citep{eeg_subjectid_iitm}. 
In contrast, Sleep-EDF uses only two bipolar channels (Fpz-Cz, Pz-Oz), which substantially limits spatial resolution and suppresses global signal components through differential referencing. 
As a result, subject-specific spatial signatures are less pronounced, reducing the model’s ability to encode subject identity.
Additionally, recent works suggest that sleep stages could include differences in aperiodic exponent, that could further explain the improved task performance~\citep{ameen2025temporally}.
\begin{table}[]
    \centering
    \caption{Linear probe performance of EEG foundation models on task and subject classification for BCIC-IV 2A, Sleep-EDF, Kaggle-ERN and Physionet-MI datasets. Cohens Kappa metric is reported as the evaluation metric. For the BCI tasks, foundation models encode subject identity over task specific information. For BCIC-IV 2A and Kaggle-ERN, since linear probe is run on single fold, standard deviations with five different random seeds are shown, whereas for Physionet-MI and Sleep-EDF five folds are used owing to larger number of subjects in datasets.}
    \resizebox{\columnwidth}{!}{
    \begin{tabular}{lcccccccccc}
        \toprule
        \multirow{3}{*}{\textbf{Model}} & \multicolumn{2}{c}{\textbf{BCIC-IV 2A}} & \multicolumn{2}{c}{\textbf{Sleep-EDF}} & \multicolumn{2}{c}{\textbf{Kaggle-ERN}} & \multicolumn{2}{c}{\textbf{Physionet-MI}}  \\
        & \multicolumn{2}{c}{(Channels=22)} & \multicolumn{2}{c}{(Channels=2)} & \multicolumn{2}{c}{(Channels=56)} & \multicolumn{2}{c}{(Channels=64)} \\
        \cmidrule(lr){2-3} \cmidrule(lr){4-5} \cmidrule(lr){6-7} \cmidrule(lr){8-9}
         & Task & Subject & Task & Subject & Task & Subject & Task & Subject \\
        \midrule
        LaBraM  & $0.34\pm0.020$ & $0.91\pm0.010$ & $0.69\pm0.02$ & $0.26\pm0.04$ & $0.26\pm0.014$ & $0.92\pm0.005$  & $0.21\pm0.03$ & $0.94\pm0.03$\\
        CbraMod & $0.27\pm0.003$ & $0.73\pm0.001$ & $0.66\pm0.03$ & $0.19\pm0.03$ & $0.21\pm0.006$ & $0.80\pm0.001$ & $0.30\pm0.03$ & $0.68\pm0.03$\\
        CSBrain & $0.31\pm0.002$ & $0.83\pm0.005$ & $0.66\pm0.02$ & $0.20\pm0.04$ & $0.28\pm0.008$ & $0.80\pm0.001$ & $0.27\pm0.04$ & $0.91\pm0.02$ \\
        \bottomrule
    \end{tabular}}
    \label{tab:eegfm_task_subject_lp_classification}
\end{table}
\begin{figure}
    \centering
    \includegraphics[width=\linewidth]{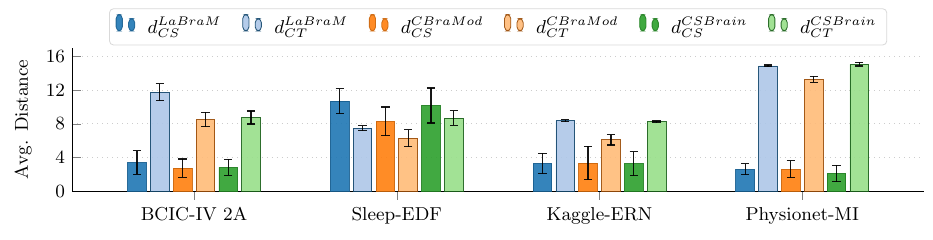}
    \caption{Euclidean distance for task and subject based clusters. As can be seen in the plots, for all the BCI datasets common subject ($d_{\text{CS}}$) clusters have much smaller average distance compared to the common task ($d_{\text{CT}}$) indicating the embeddings forming tight clusters based on subject identity.}
    \label{fig:embedding_distances}
\end{figure}
\begin{figure}
    \centering
    \includegraphics[width=0.8\linewidth]{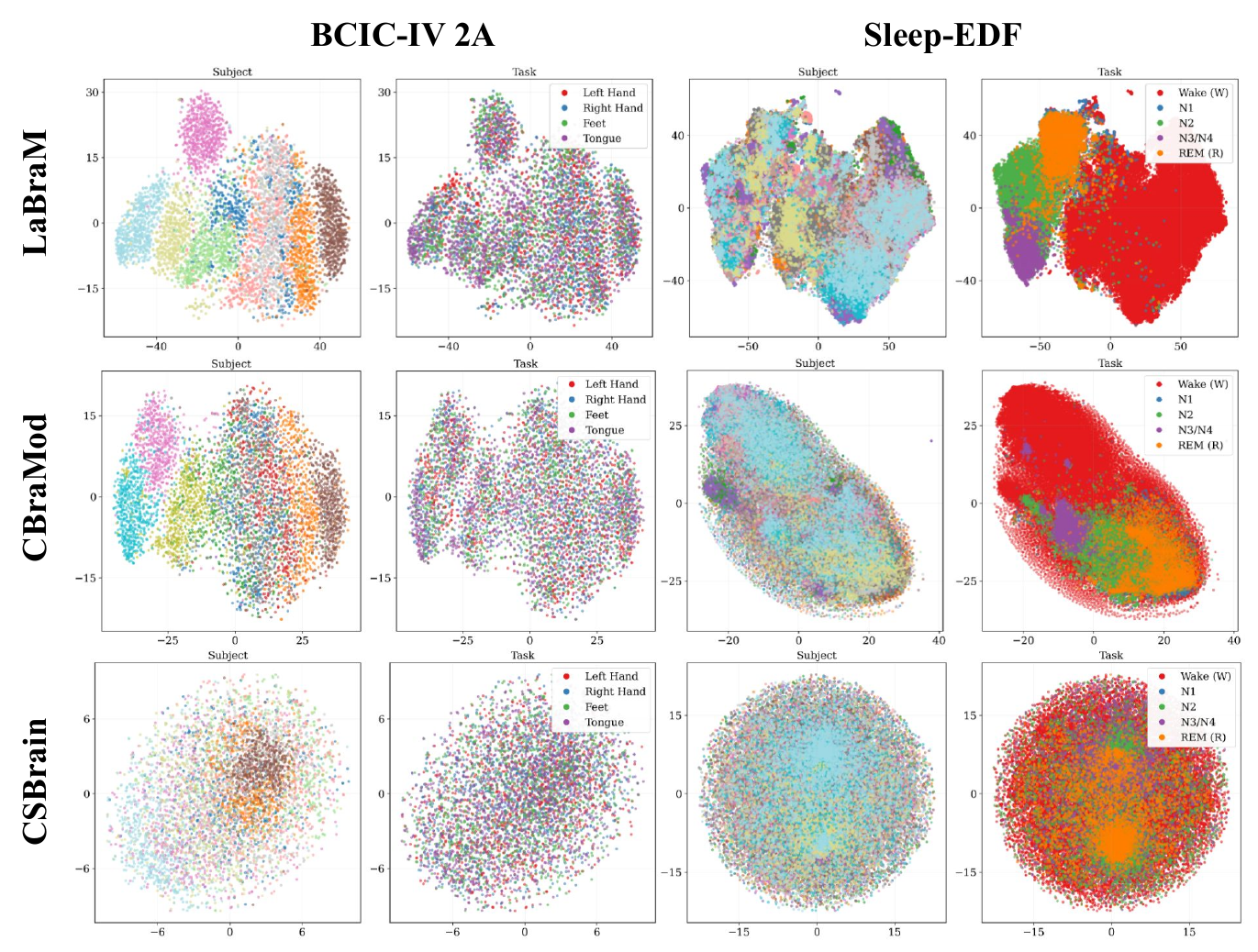}
    \caption{\textbf{t-SNE embeddings of pre-trained LaBraM, CBraMod, and CSBrain models across the BCIC IV 2a and Sleep-EDF datasets.} t-SNE plots indicate that models generally learn representations that tend to cluster by subject identity rather than by task label. Embeddings from a maximum of 15 subjects are shown for clarity.}
    \label{fig:tsne}
\end{figure}
\paragraph{\textbf{Representation cluster distances align with linear probe results}}
To analyze the geometric structure of learned embeddings, we define two distance-based metrics over class centroids in embedding space. Let $\mathbf{c}_{s,t} = \frac{1}{|\mathcal{D}_{s,t}|} \sum_{i \in \mathcal{D}_{s,t}} \mathbf{z}_i$ denote the centroid of embeddings $\mathbf{z}_i \in \mathbb{R}^d$ for subject $s$ and task $t$. \textbf{Common Subject} ($d_{\text{CS}}$) and \textbf{Common Task} ($d_{\text{CT}}$) distances are then defined as:
\begin{equation}
d_{\mathrm{CS}} = \frac{1}{|\mathcal{S}|} \sum_{s \in \mathcal{S}} \frac{\sum_{t_1 \neq t_2} \|\mathbf{c}_{s,t_1} - \mathbf{c}_{s,t_2}\|_2}{|\mathcal{T}_s|(|\mathcal{T}_s|-1)}, \quad
d_{\mathrm{CT}} = \frac{1}{|\mathcal{T}|} \sum_{t \in \mathcal{T}} \frac{\sum_{s_1 \neq s_2} \|\mathbf{c}_{s_1,t} - \mathbf{c}_{s_2,t}\|_2}{|\mathcal{S}_t|(|\mathcal{S}_t|-1)},
\end{equation}
where $\mathcal{S}$, $\mathcal{T}$ are the sets of subjects and tasks, and $\mathcal{T}_s$, $\mathcal{S}_t$ denote tasks for subject $s$ and subjects for task $t$, respectively. As shown in Fig.~\ref{fig:embedding_distances}, $d_{\mathrm{CS}} \ll d_{\mathrm{CT}}$ for BCI tasks, indicating that embeddings cluster more tightly by subject than by task, suggesting subject-specific characteristics dominate the representation space. For Sleep-EDF, the trend is reversed aligning with linear probe results, reflecting a weaker subject-centric effect and stronger task-related structure.
Further analysis on model representations through t-SNE plots with for both subject and task labels indicate clusters primarily based on the subject identity rather the task in BCI datasets as seen in Fig.~\ref{fig:tsne}.
This further reinforces the initial hypothesis of the model embeddings capturing low-frequency and aperiodic components rather than oscillatory information. 
Additional t-SNE plots are provided in Appendix~\ref{sec:additional_tsne_plots}.

\section{Limitations and Future Work}
\label{sec:limitations}
While this work identifies aperiodic and low-frequency spectral bias to be an impediment in EEG foundation models ability to provide generalized representations for robust downstream performance, mitigation strategies through auxiliary losses that enable encoding high-frequency oscillatory information explicitly must be explored in future work.
Some directions that have been explored in the context of reconstruction models include augmenting knowledge-guided objectives~\citep{s4eeg_kommineni, wang2026deeperbrain} and distinct subspaces for subject- and task-specific representations~\citep{mishra2025subject}. 
While the controlled synthetic EEG experiments help identify the sensitivity of representations to aperiodic and oscillatory component effects, the analysis is limited to stationary single-channel signals. 
Further analysis including multi-channel non-stationary signals could provide deeper insights into the representation topology of EEG foundation models.
The primary focus of this work was to characterize the spectral bias in reconstruction-based EEG foundation models and examine its downstream impact on BCI performance. 
While this bias negatively affects tasks that rely on high-frequency neural signatures, it may conversely benefit tasks driven by low-frequency, high-amplitude components, as suggested by the improved performance observed on the sleep staging task in this work. 
Additionally, our analysis is scoped to reconstruction-based pre-training objectives, which have seen increasing adoption in the EEG foundation model literature. 
Alternative objectives such as contrastive or predictive approaches may exhibit different inductive biases, and understanding how these compare to the spectral bias identified here remains an important direction for future work.
\section{Conclusion}
Our work provides a mechanistic explanation for sub-optimal linear probe results in EEG foundation models by identifying the spectral bias in foundation models towards aperiodic and low-frequency components.
This spectral bias is attributed to reconstruction based objectives, wherein the large spectral power in the aperiodic components with 1/f spectral nature incentivizes the model to learn lower frequency components.
This bias towards aperiodic components is empirically demonstrated in foundation models through a controlled synthetic EEG samples wherein the representations of models are able to linearly decode aperiodic exponent and offset but not oscillatory information. 
Further, we demonstrate that this spectral bias results in representations encoding subject specific information in real world EEG datasets, leading to much higher subject identification performance over task classification.
These observations motivate employing modified pretext tasks that explicitly model high-frequency components in EEG.

\bibliographystyle{plainnat}
\bibliography{main}

\appendix

\section{Generative AI Use Disclosure}
Large Language Models were employed for refining the quality of writing in this manuscript. 
However, all content generated through Large Language Models was verified by the authors and modified accordingly.
Large Language Models were used to generate code to run experiments and format tables after the generated code was verified by the authors for its veracity.

\section{Training Details \& Hyperparameters}
\label{sec:appendix_hyperparameters}
All linear probe models were trained on a single NVIDIA RTXA6000 GPU. For Sleep-EDF learning rate 1e-2 was used, whereas for the BCI datasets, CSBrain and CBramod use 1e-3 whereas LaBraM uses 5e-4.
For the linear probe experiments, models were initialized with pre-trained checkpoints.
Embeddings were extracted from the last layer of the encoder. 
Code for synthetic EEG generation and linear probe training has been uploaded to ~\href{https://anonymous.4open.science/r/spectralbiasaperiodic/}{https://anonymous.4open.science/r/spectralbiasaperiodic/}.
Table~\ref{tab:hyperparameters} reports the hyperparameters for the experiments.
\begin{table}[H]
    \centering
    \caption{Hyperparameters for linear probe experiments}
    \begin{tabular}{lc}
    \toprule
       \textbf{Hyperparameter}  & \textbf{Value}  \\
       \midrule
     Training Epochs & $31$ \\
     Batch Size & 64 \\
     Learning Rate & $\{1e-2, 1e-3, 5e-4\}$ \\
     Adam $\beta$ & (0.9, 0.999) \\ 
     Adam $\epsilon$ & $1e-8$ \\ 
     Optimizer    &  AdamW \\
     Weight decay & 1e-2 \\ 
     Learning Rate Scheduler & CosineAnnealingLR \\ 
     Minimal Learning Rate & 1e-5 \\
    \bottomrule
    \end{tabular}
    \label{tab:hyperparameters}
\end{table}
\section{Synthetic EEG Single Channel Examples}
\label{sec:synthetic_eeg_appendix}
\begin{figure}
    \centering
    \includegraphics[width=\linewidth]{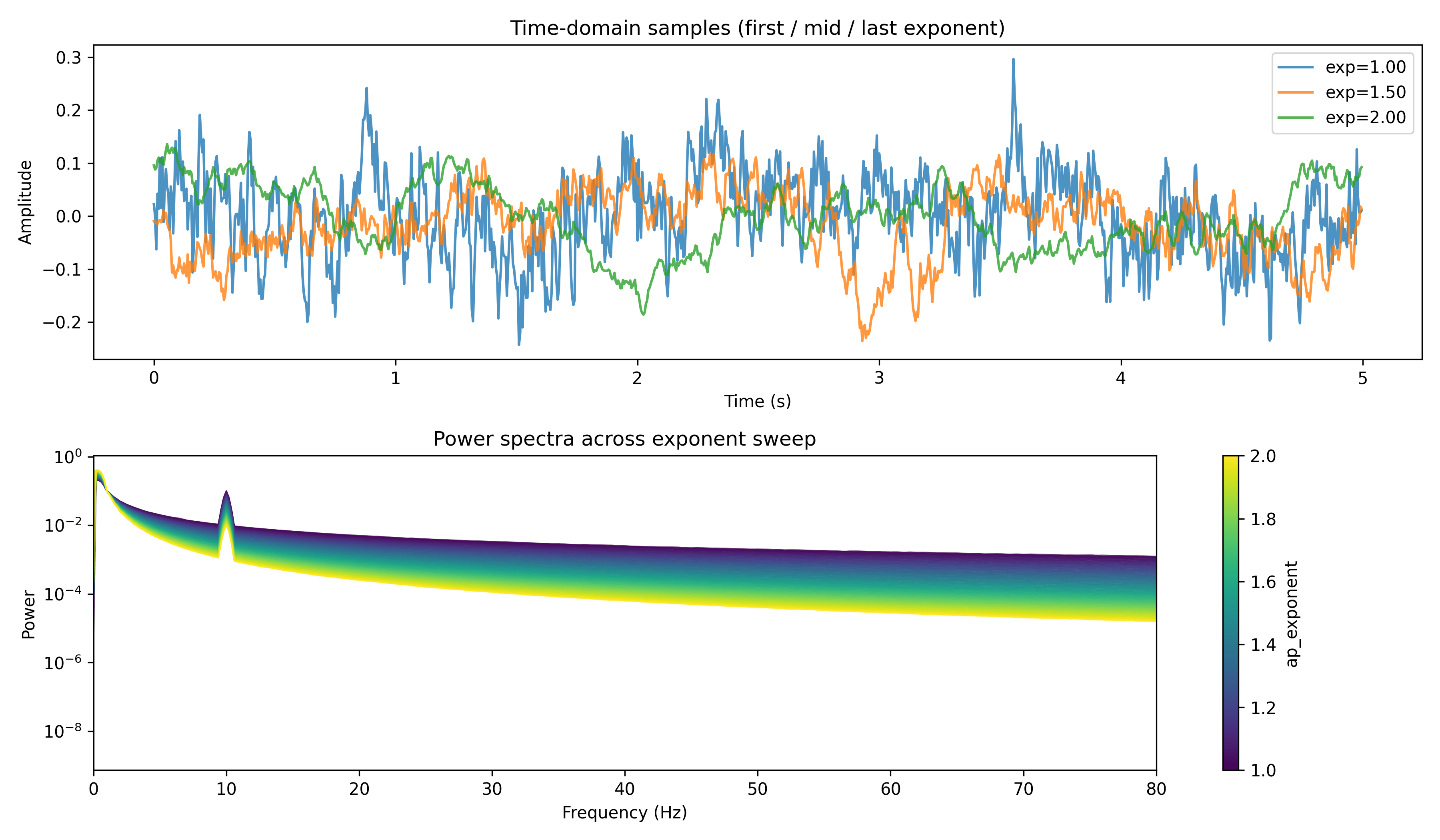}
    \caption{(Top) Time series of generated EEG signal for $\beta \in \{1.0, 1.5, 2.0\}$ (Bottom) Frequency spectra plot for EEG signals when a sweep of exponent value from 1.0 to 2.0 is performed.}
    \label{fig:sweep_exponent}
\end{figure}

Pseudocode for single synthetic EEG sample generation is provided in Pseudocode~\ref{alg:timeseries} and the sweep for a variable of interest is provided in Pesudocode~\ref{alg:sweep}.
FOOOF python package was used for generating the frequency spectrum of single channel EEG~\citep{donoghue2020parameterizing}. 
Fig.~\ref{fig:sweep_exponent} illustrates the examples generated by sweep of exponent values from 1.0 to 2.0. 

\captionsetup[algorithm]{labelformat=empty}
\begin{algorithm}[t]
\caption{\textbf{Pseudocode 1} Synthetic Time Series Generation from Parameters}
\label{alg:timeseries}
\begin{algorithmic}[1]
\REQUIRE Frequency range $[f_{\min}, f_{\max}]$, sampling rate $f_s$, duration $T$, aperiodic parameters $(\beta, A_{ap})$, optional oscillation information $f_{osc}$, $A_{osc}$
\ENSURE Time series $x \in \mathbb{R}^{L}$

\STATE $L \leftarrow f_s \cdot T$
\STATE $(\mathbf{f}, \mathbf{P}) \leftarrow \texttt{gen\_power\_spectrum}([f_{\min}, f_{\max}], [\beta, A_{ap}], [f_{osc}, A_{osc}])$
\STATE $\mathbf{f}_{FFT} \leftarrow \texttt{rfftfreq}(L, 1/f_s)$
\STATE $\mathbf{P}_{interp} \leftarrow \texttt{interp}(\mathbf{f}_{FFT}, \mathbf{f}, \mathbf{P})$
\STATE $\boldsymbol{\phi} \sim \mathcal{U}(0, 2\pi)$
\STATE $\mathbf{A} \leftarrow \sqrt{\mathbf{P}_{interp}}$
\STATE $\mathbf{S} \leftarrow \mathbf{A} \odot e^{i\boldsymbol{\phi}}$
\STATE $x \leftarrow \texttt{irfft}(\mathbf{S}, L)$
\RETURN $x$
\end{algorithmic}
\end{algorithm}

\begin{algorithm}[t]
\caption{\textbf{Pseudocode 2} Parameter Sweep for Dataset Generation}
\label{alg:sweep}
\begin{algorithmic}[1]
\REQUIRE Parameter name $p$, range $[\theta_{\min}, \theta_{\max}]$, number of samples $N$, base parameters $\Psi$
\ENSURE Dataset $\mathcal{D} = \{x^{(i)}\}_{i=1}^N$, parameter values $\boldsymbol{\theta}$

\STATE $\boldsymbol{\theta} \leftarrow \texttt{linspace}(\theta_{\min}, \theta_{\max}, N)$
\STATE $\mathcal{D} \leftarrow \emptyset$

\FOR{$i = 1$ to $N$}
    \STATE $\Psi[p] \leftarrow \boldsymbol{\theta}_i$
    \STATE $x^{(i)} \leftarrow \texttt{generate\_time\_series}(\Psi)$
    \STATE $\mathcal{D} \leftarrow \mathcal{D} \cup \{x^{(i)}\}$
\ENDFOR

\RETURN $\mathcal{D}, \boldsymbol{\theta}$
\end{algorithmic}
\end{algorithm}
\section{Linear Decodability Results}
\label{sec:additional_linear_decoding_results}
\begin{figure}[H]
    \centering
    
    % --- Column Titles ---
    \makebox[0.3\linewidth]{$\bm{\beta}$} \hfill
    \makebox[0.3\linewidth]{$\bm{A_{\text{ap}}}$} \hfill
    \makebox[0.3\linewidth]{$\bm{f_{\text{osc}}}$} \\ \vspace{0.2em}

    % --- Row 1 (CBraMod) ---
    \makebox[0pt][r]{\raisebox{1.4cm}[0pt][0pt]{\rotatebox[origin=c]{90}{\textbf{CBraMod}}}\hspace{1em}}%
    \begin{subfigure}[t]{0.3\linewidth}
        \centering
        \includegraphics[width=\linewidth]{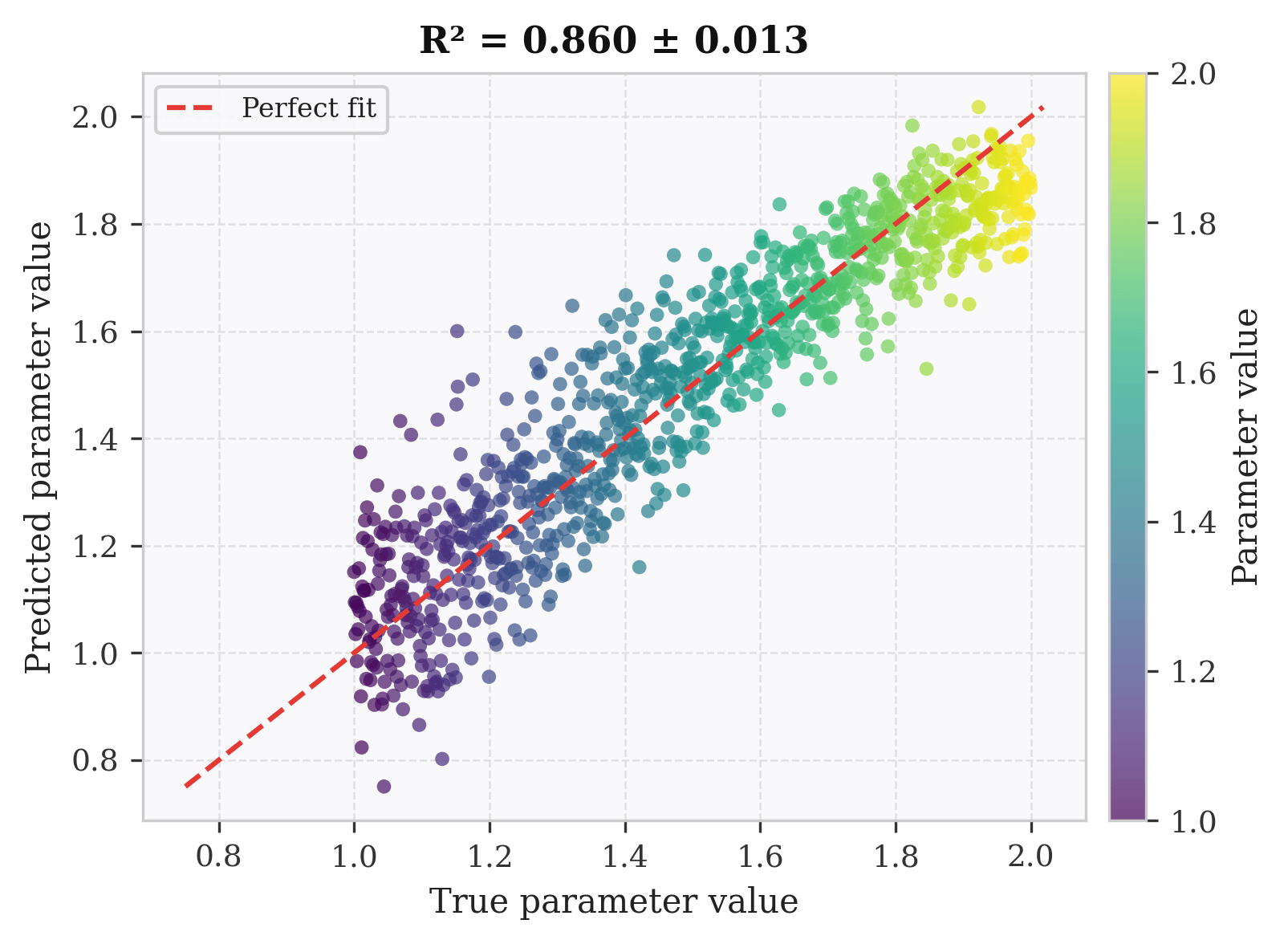}
        \caption{}
    \end{subfigure}
    \hfill
    \begin{subfigure}[t]{0.3\linewidth}
        \centering
        \includegraphics[width=\linewidth]{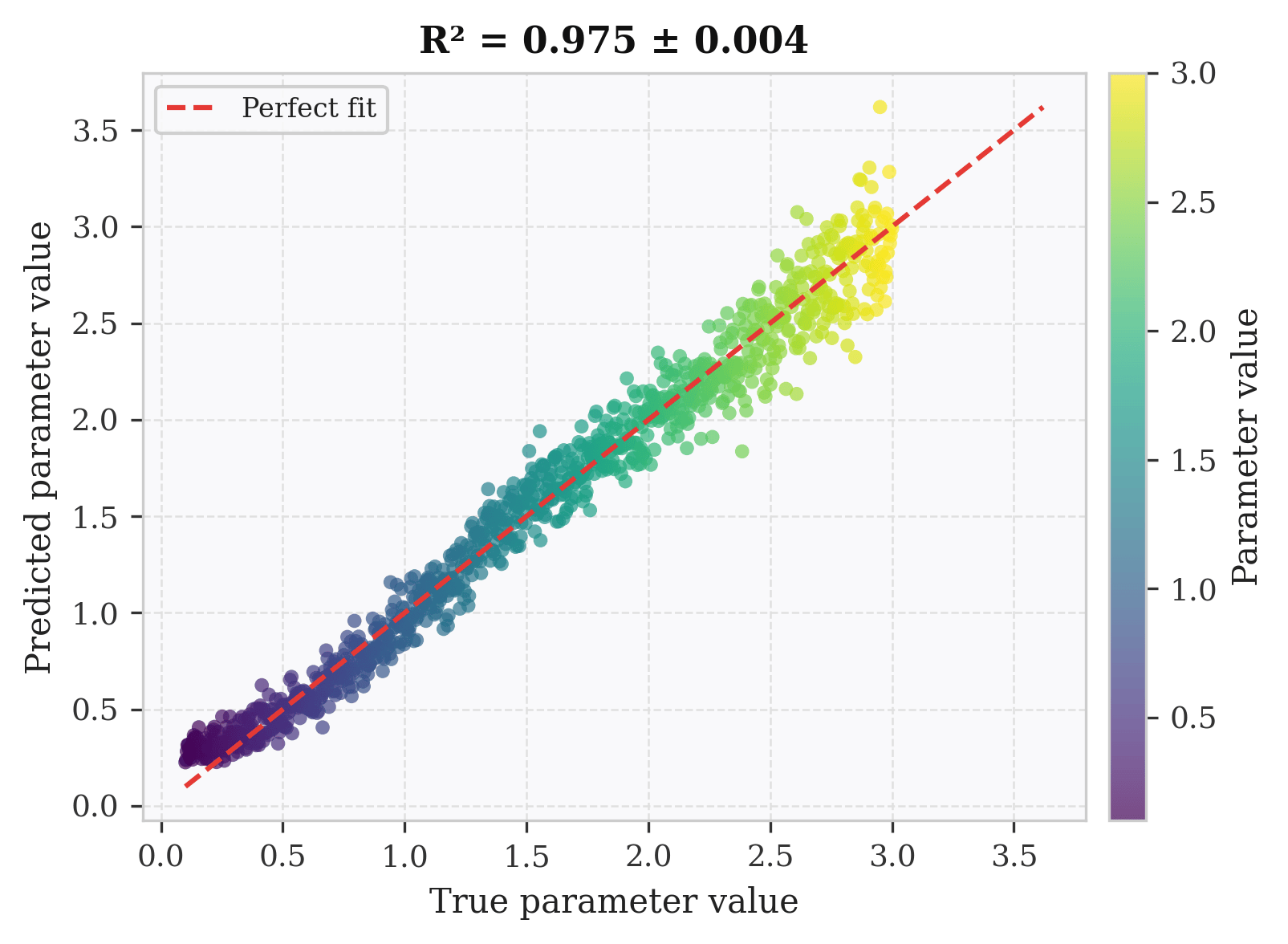}
        \caption{}
    \end{subfigure}
    \hfill
    \begin{subfigure}[t]{0.3\linewidth}
        \centering
        \includegraphics[width=\linewidth]{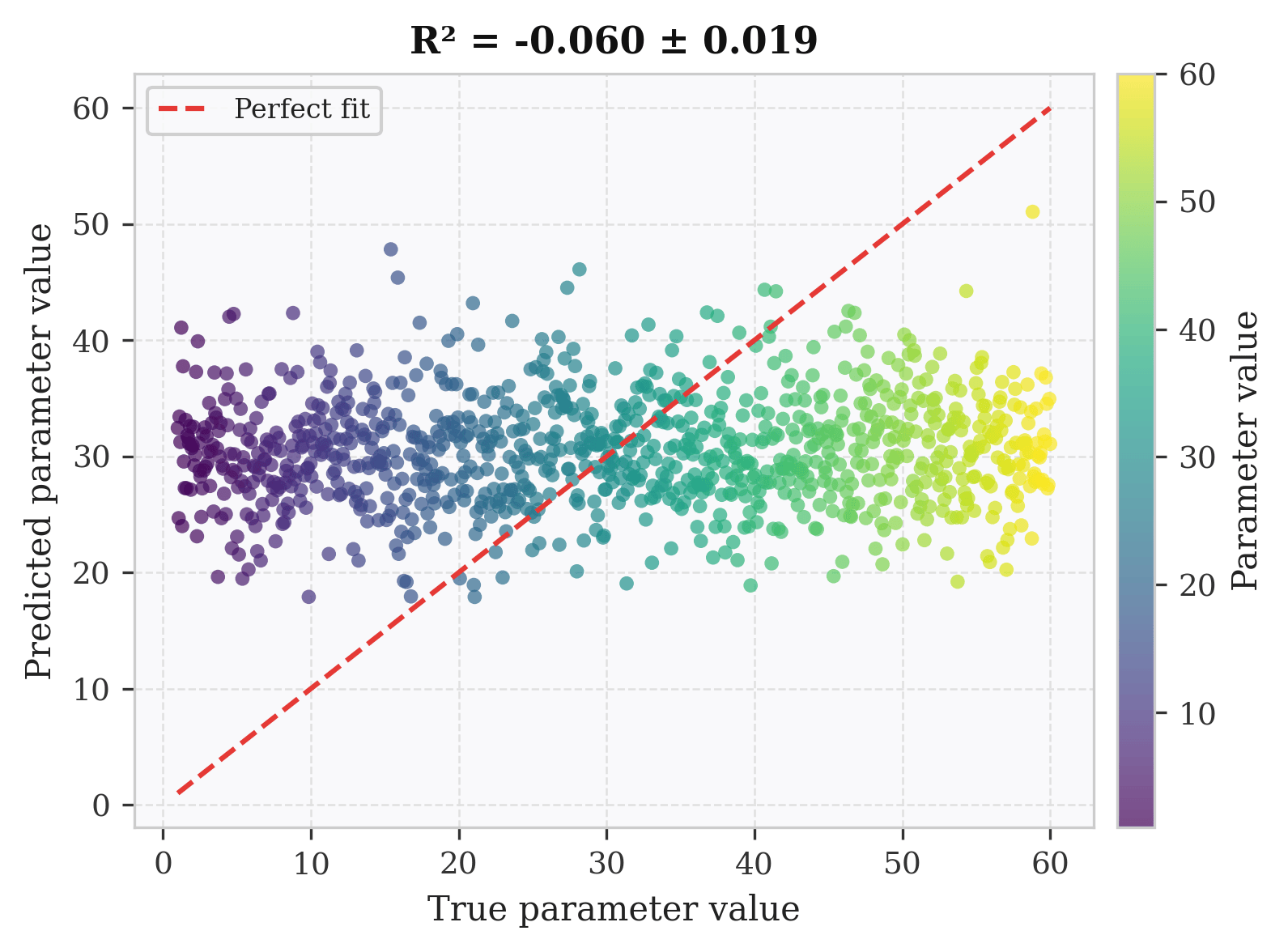}
        \caption{}
    \end{subfigure}

    \vspace{-0.5cm}

    % --- Row 2 (CSBrain) ---
    \makebox[0pt][r]{\raisebox{1.4cm}[0pt][0pt]{\rotatebox[origin=c]{90}{\textbf{CSBrain}}}\hspace{1em}}%
    \begin{subfigure}[t]{0.3\linewidth}
        \centering
        \includegraphics[width=\linewidth]{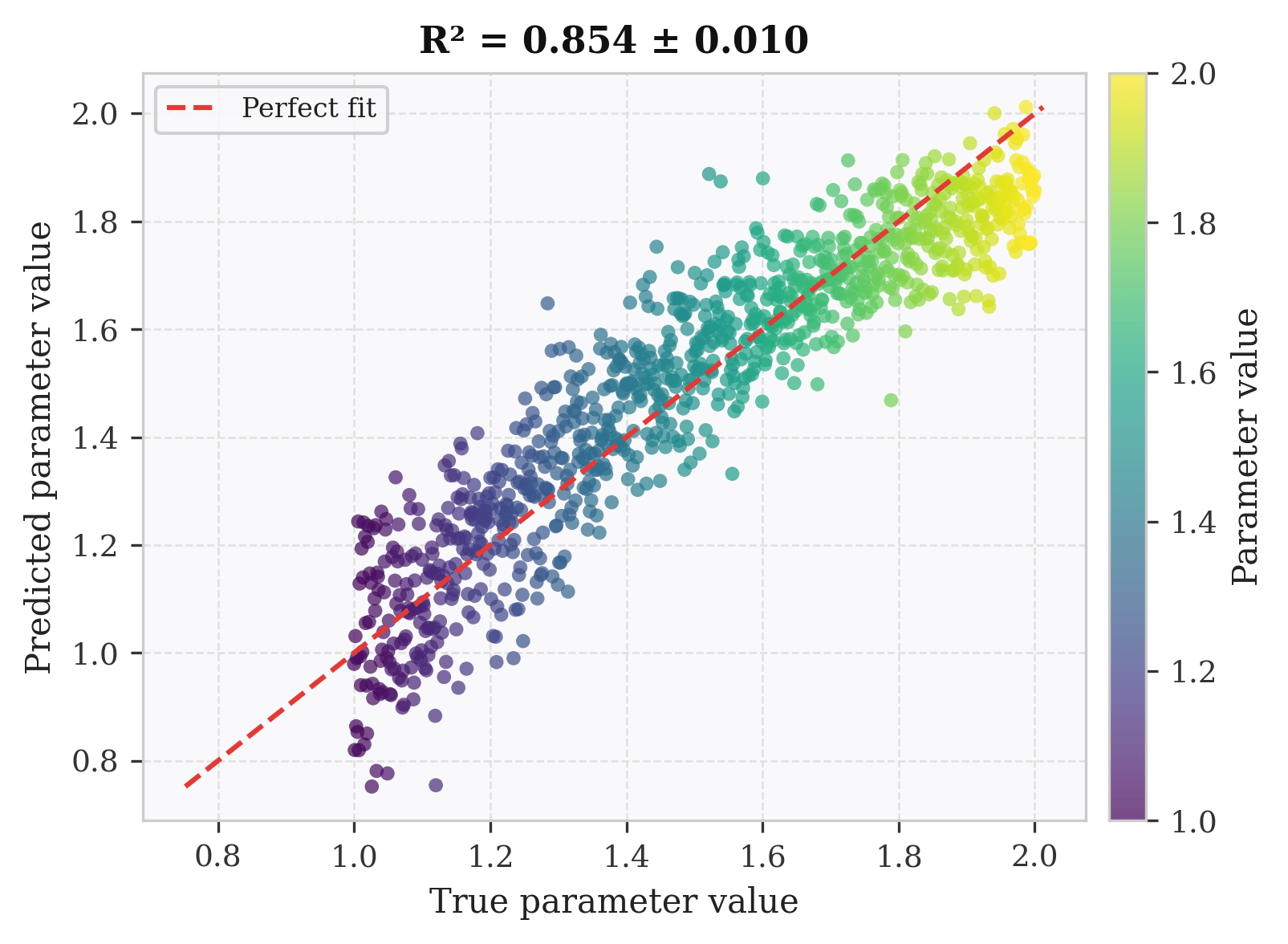}
        \caption{}
    \end{subfigure}
    \hfill
    \begin{subfigure}[t]{0.3\linewidth}
        \centering
        \includegraphics[width=\linewidth]{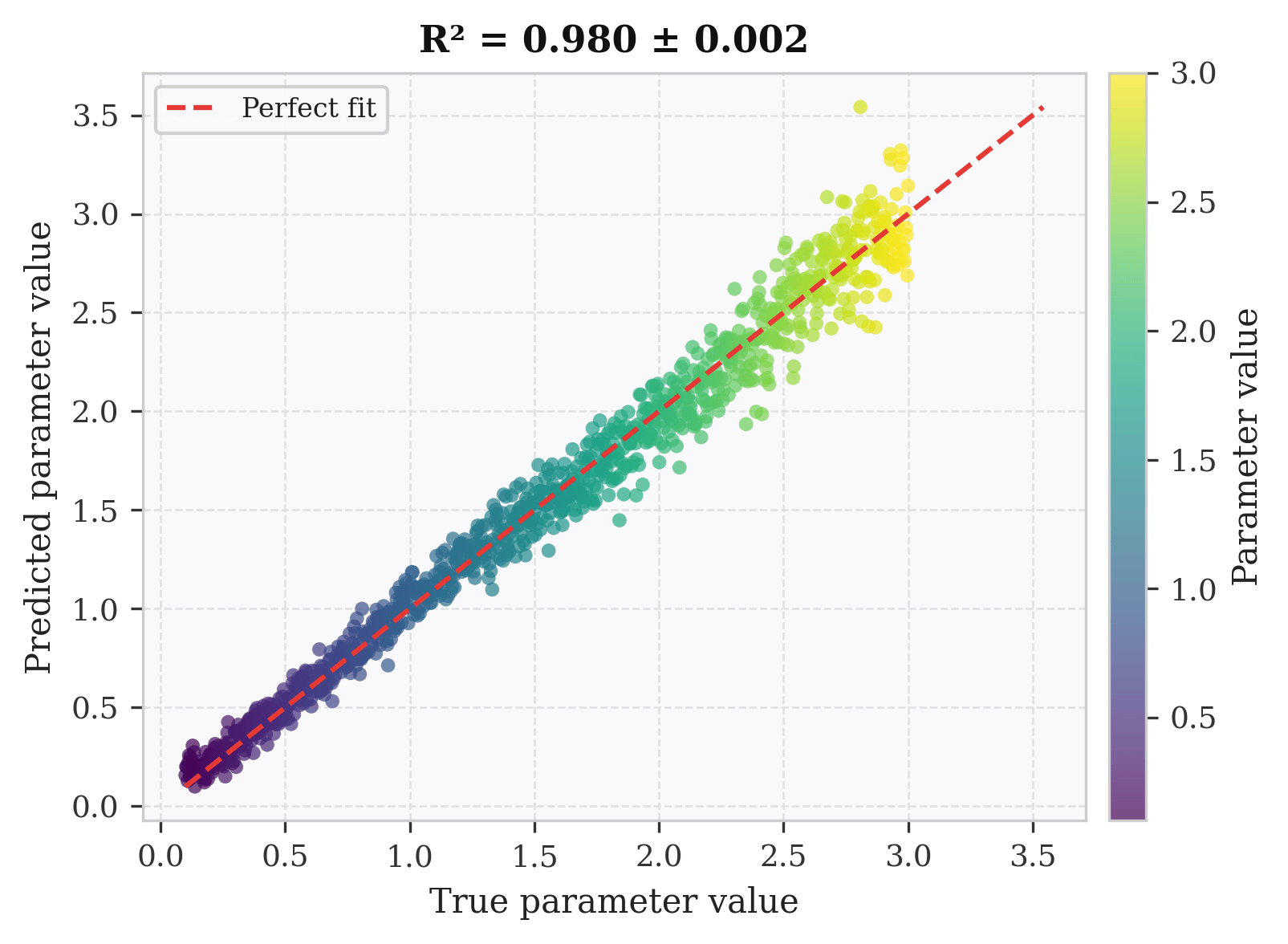}
        \caption{}
    \end{subfigure}
    \hfill
    \begin{subfigure}[t]{0.3\linewidth}
        \centering
        \includegraphics[width=\linewidth]{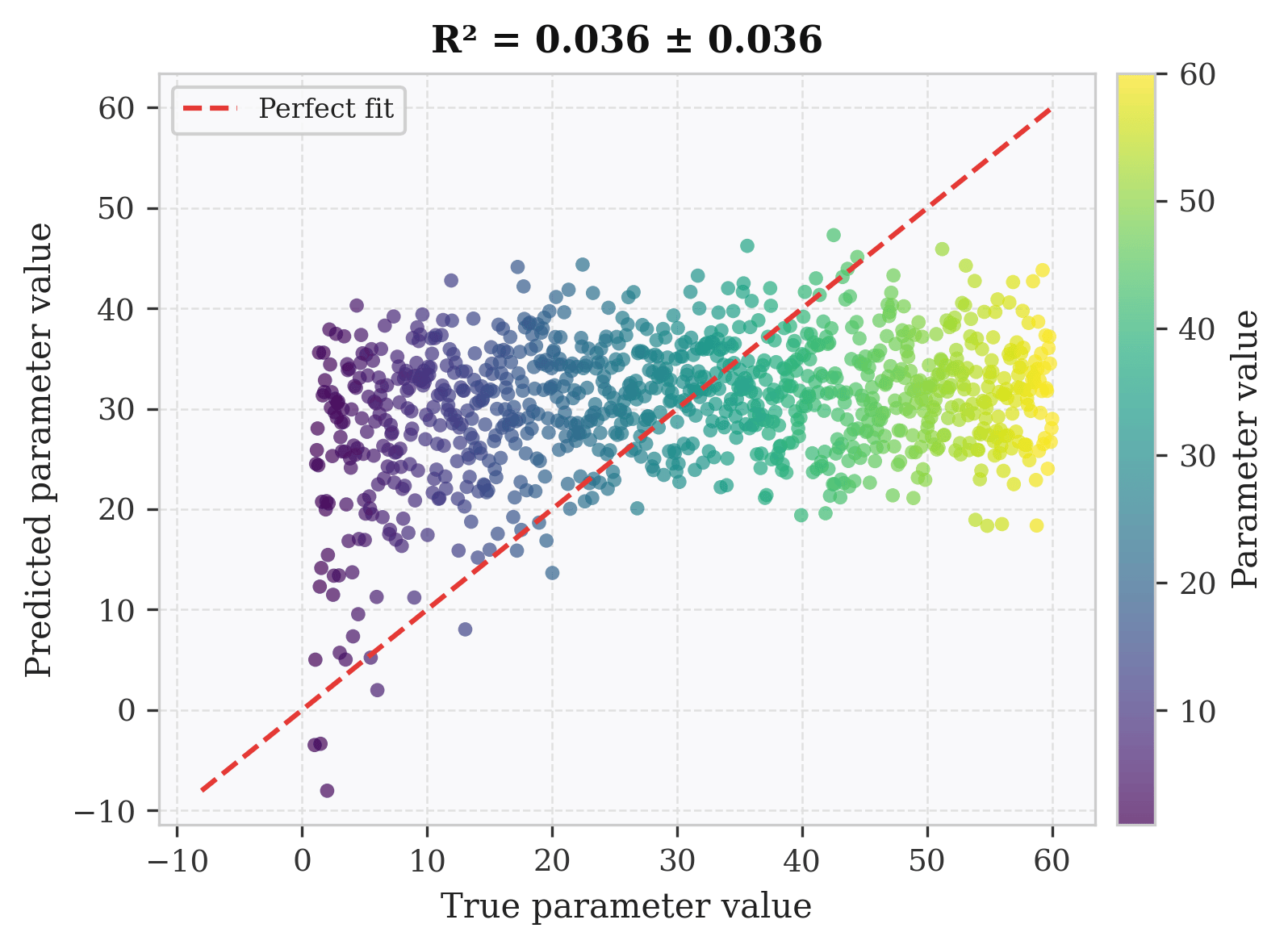}
        \caption{}
    \end{subfigure}

    \vspace{-0.5cm}

    % --- Row 3 (LaBraM) ---
    \makebox[0pt][r]{\raisebox{1.2cm}[0pt][0pt]{\rotatebox[origin=c]{90}{\textbf{LaBraM}}}\hspace{1em}}%
    \begin{subfigure}[t]{0.3\linewidth}
        \centering
        \includegraphics[width=\linewidth]{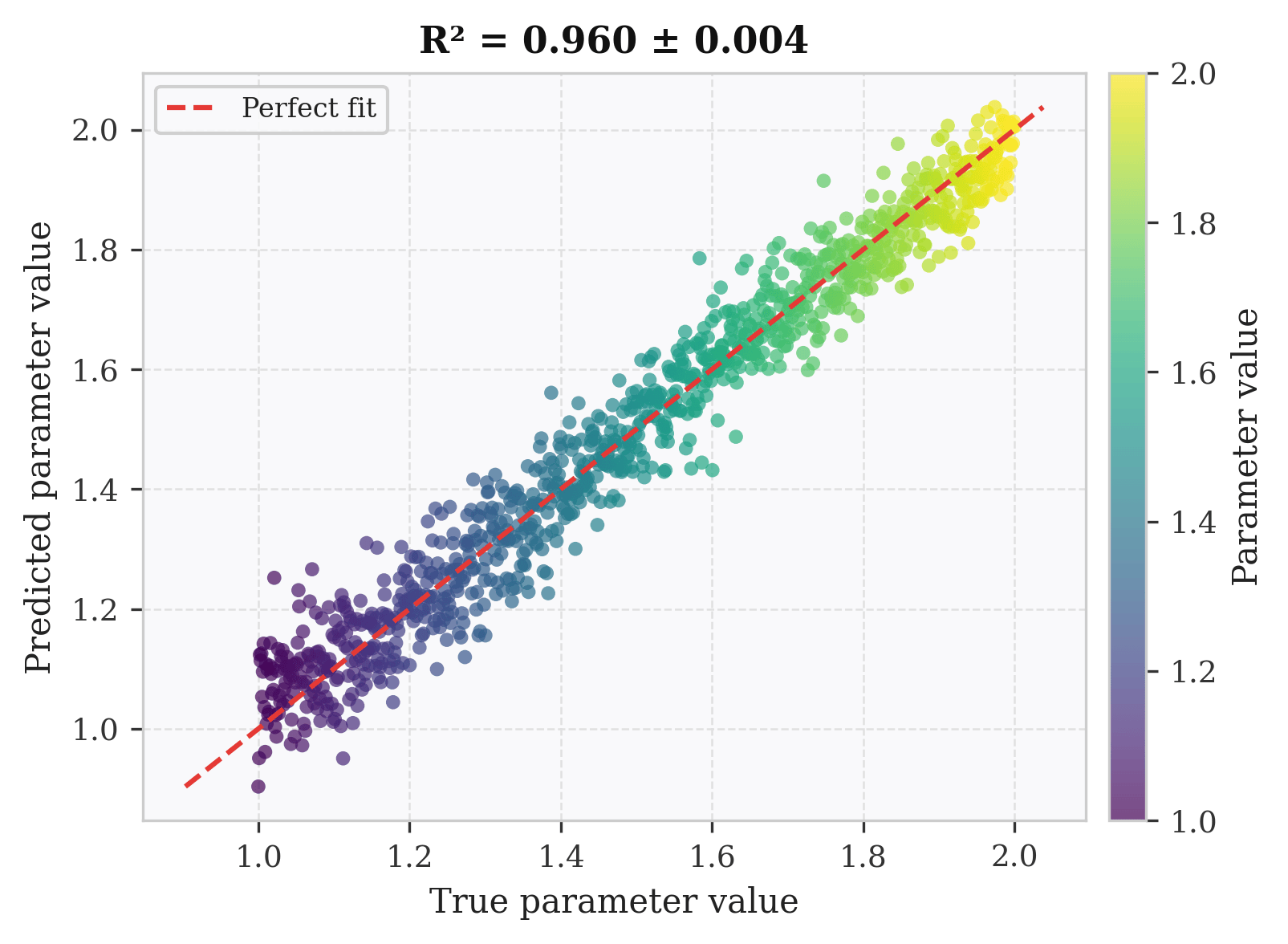}
        \caption{}
    \end{subfigure}
    \hfill
    \begin{subfigure}[t]{0.3\linewidth}
        \centering
        \includegraphics[width=\linewidth]{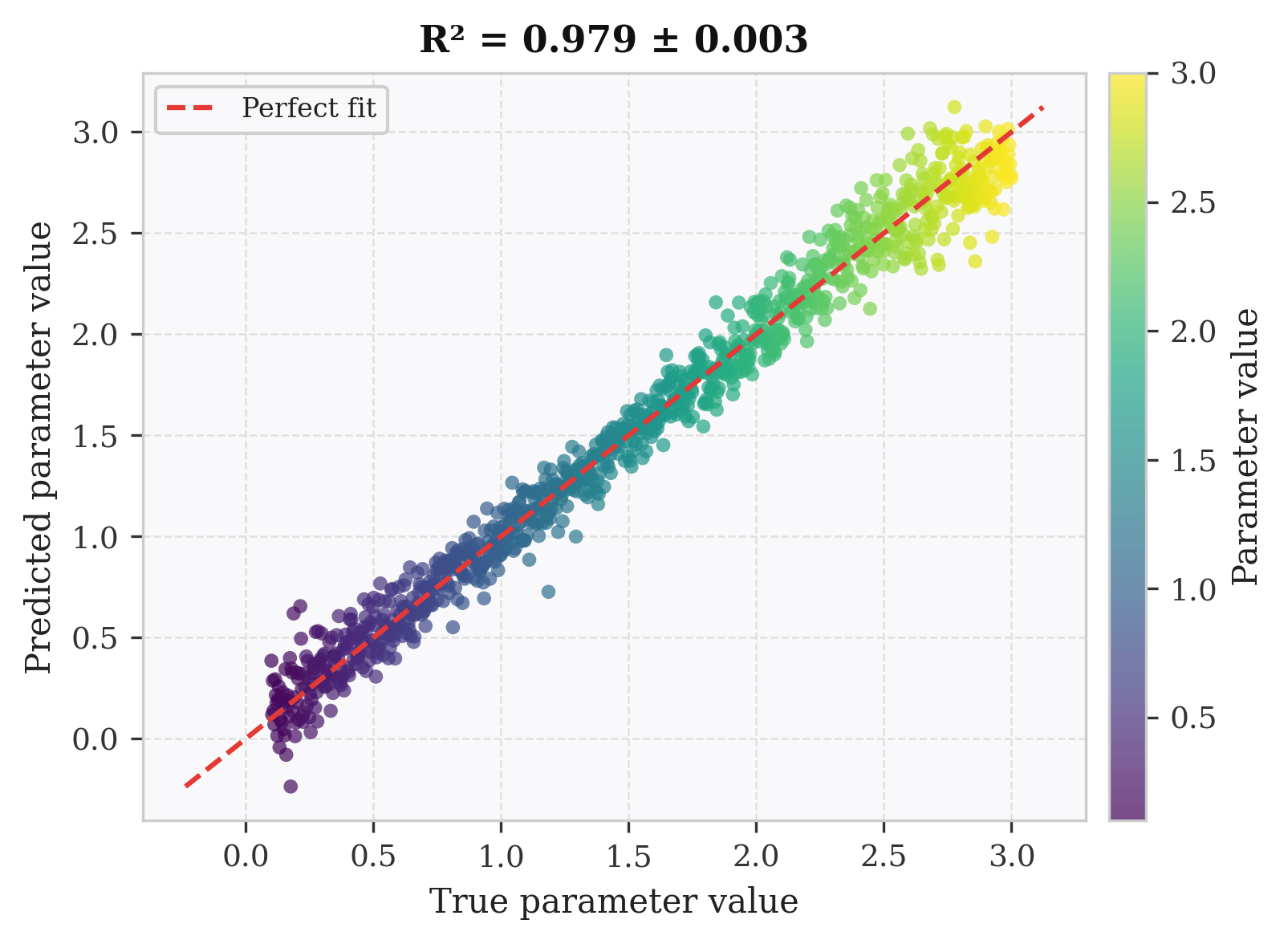}
        \caption{}
    \end{subfigure}
    \hfill
    \begin{subfigure}[t]{0.3\linewidth}
        \centering
        \includegraphics[width=\linewidth]{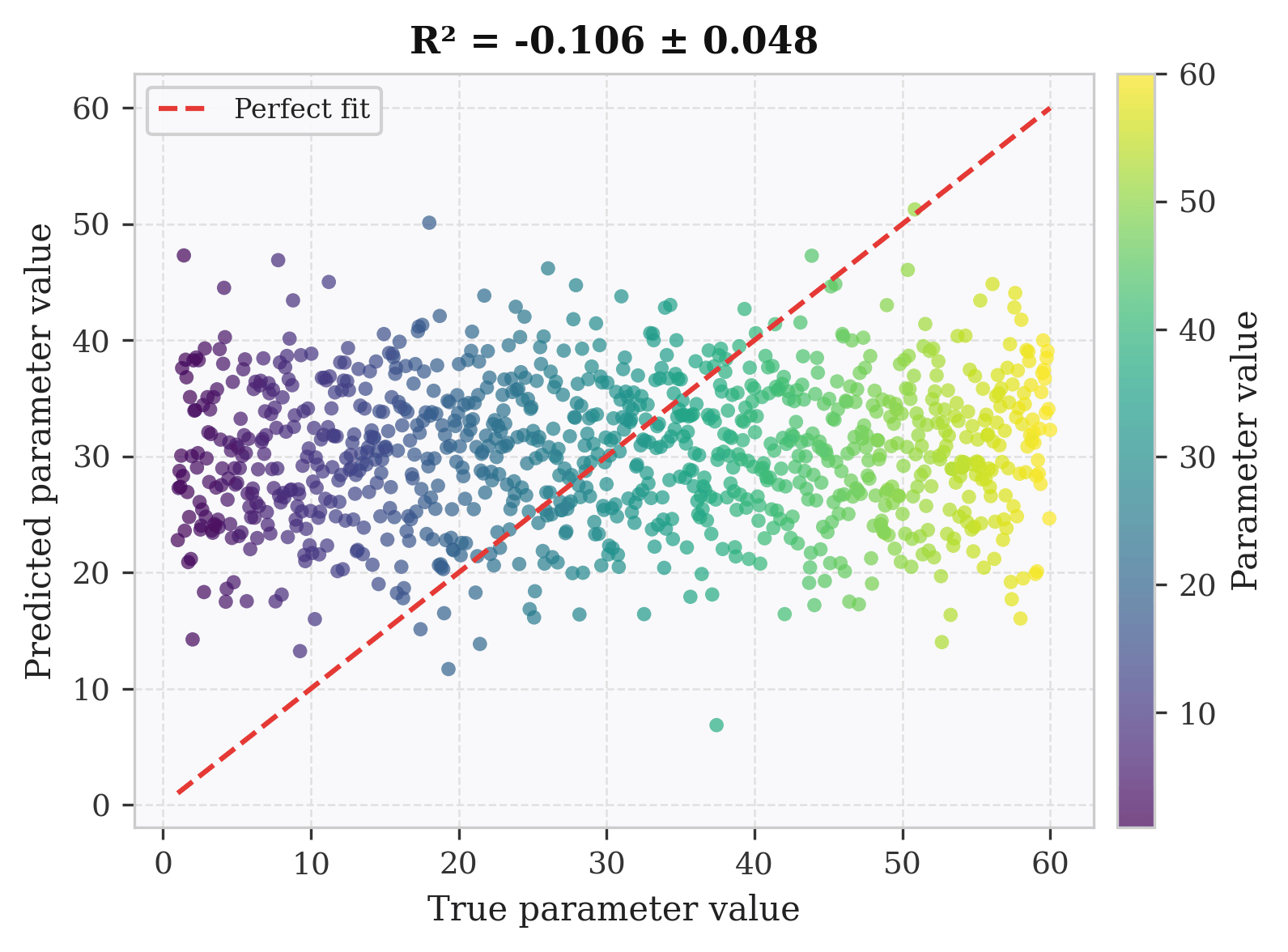}
        \caption{}
    \end{subfigure}

    \caption{
    Linear decodability for Fz channel across three foundation models (CBraMod, CSBrain, LaBraM) for Aperiodic Exponent ($\beta$), Aperiodic Offset ($A_{\text{ap}}$) and Oscillation frequency ($f_{\text{osc}}$).
    }
    \label{fig:Fz}
\end{figure}

\begin{figure}[H]
    \centering
    
    % --- Column Titles ---
    \makebox[0.19\linewidth]{\textbf{10Hz}} \hfill
    \makebox[0.19\linewidth]{\textbf{20Hz}} \hfill
    \makebox[0.19\linewidth]{\textbf{30Hz}} \hfill
    \makebox[0.19\linewidth]{\textbf{40Hz}} \hfill
    \makebox[0.19\linewidth]{\textbf{50Hz}}\\ \vspace{0.2em}

    \makebox[0pt][r]{\raisebox{1.2cm}[0pt][0pt]{\rotatebox[origin=c]{90}{\textbf{CSBrain}}}\hspace{1em}}%
    \begin{subfigure}[t]{0.19\linewidth}
        \centering
        \includegraphics[width=\linewidth]{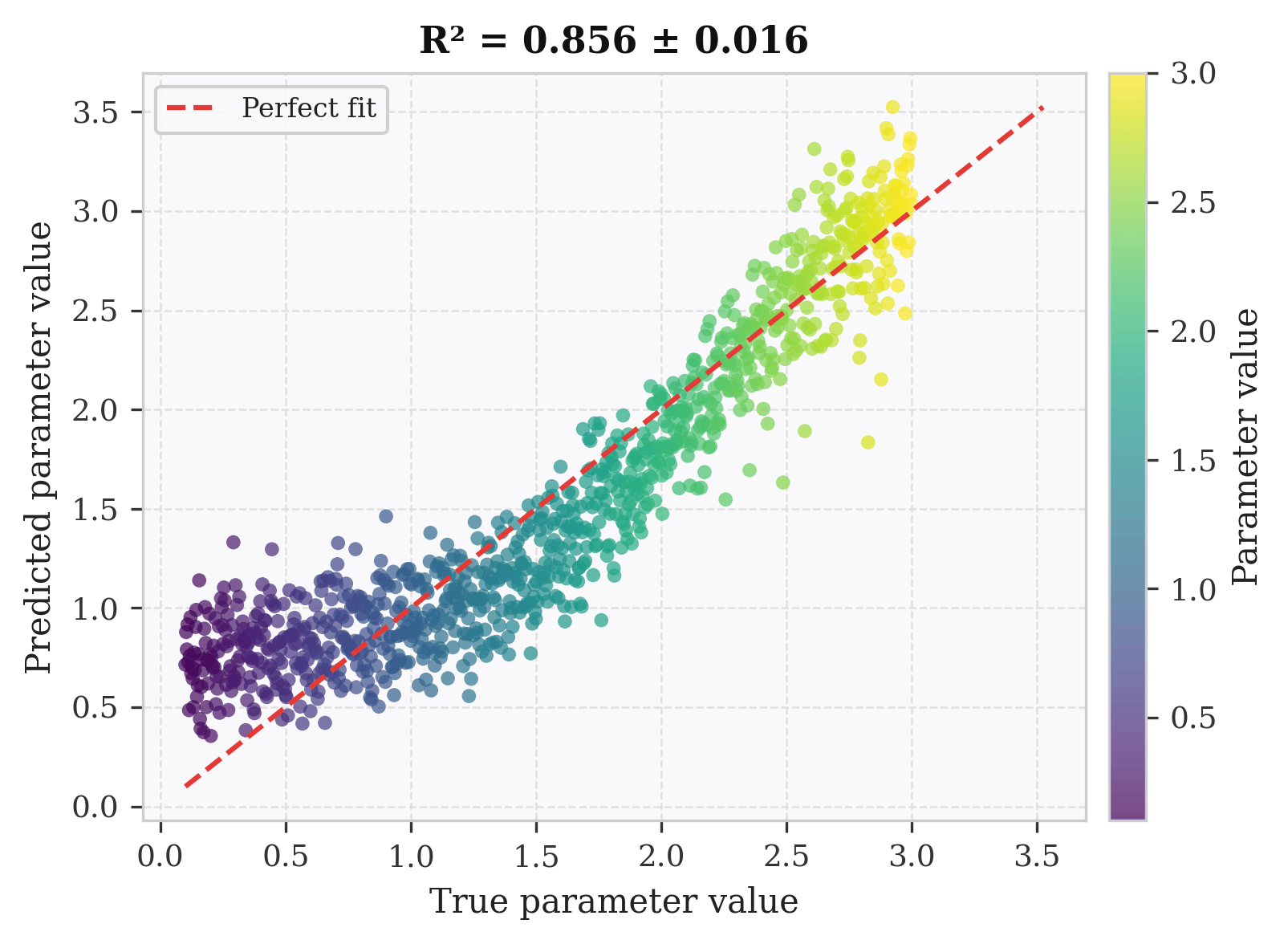}
        \caption{}
    \end{subfigure}%
    \hfill
    \begin{subfigure}[t]{0.19\linewidth}
        \centering
        \includegraphics[width=\linewidth]{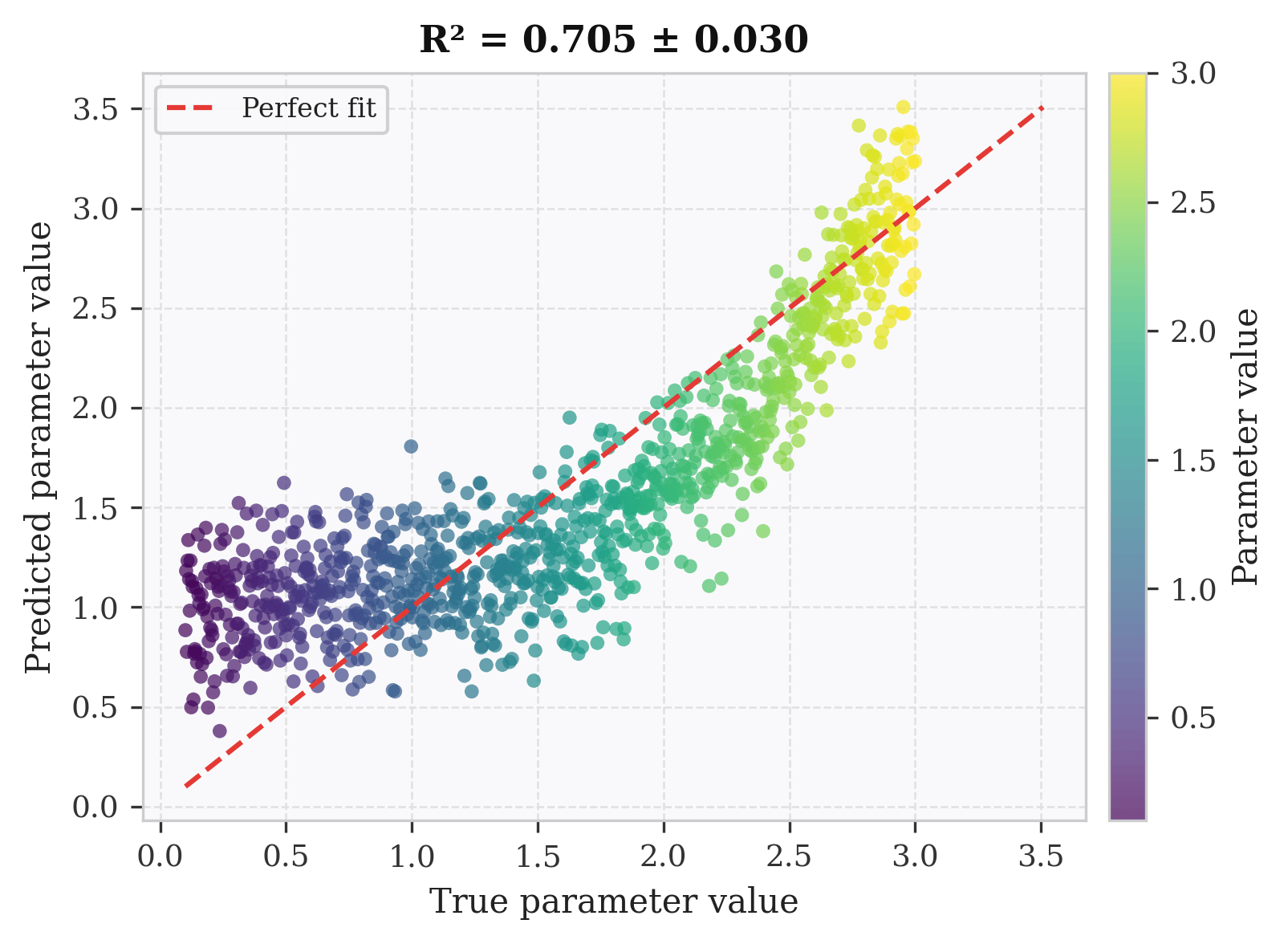}
        \caption{}
    \end{subfigure}%
    \hfill
    \begin{subfigure}[t]{0.19\linewidth}
        \centering
        \includegraphics[width=\linewidth]{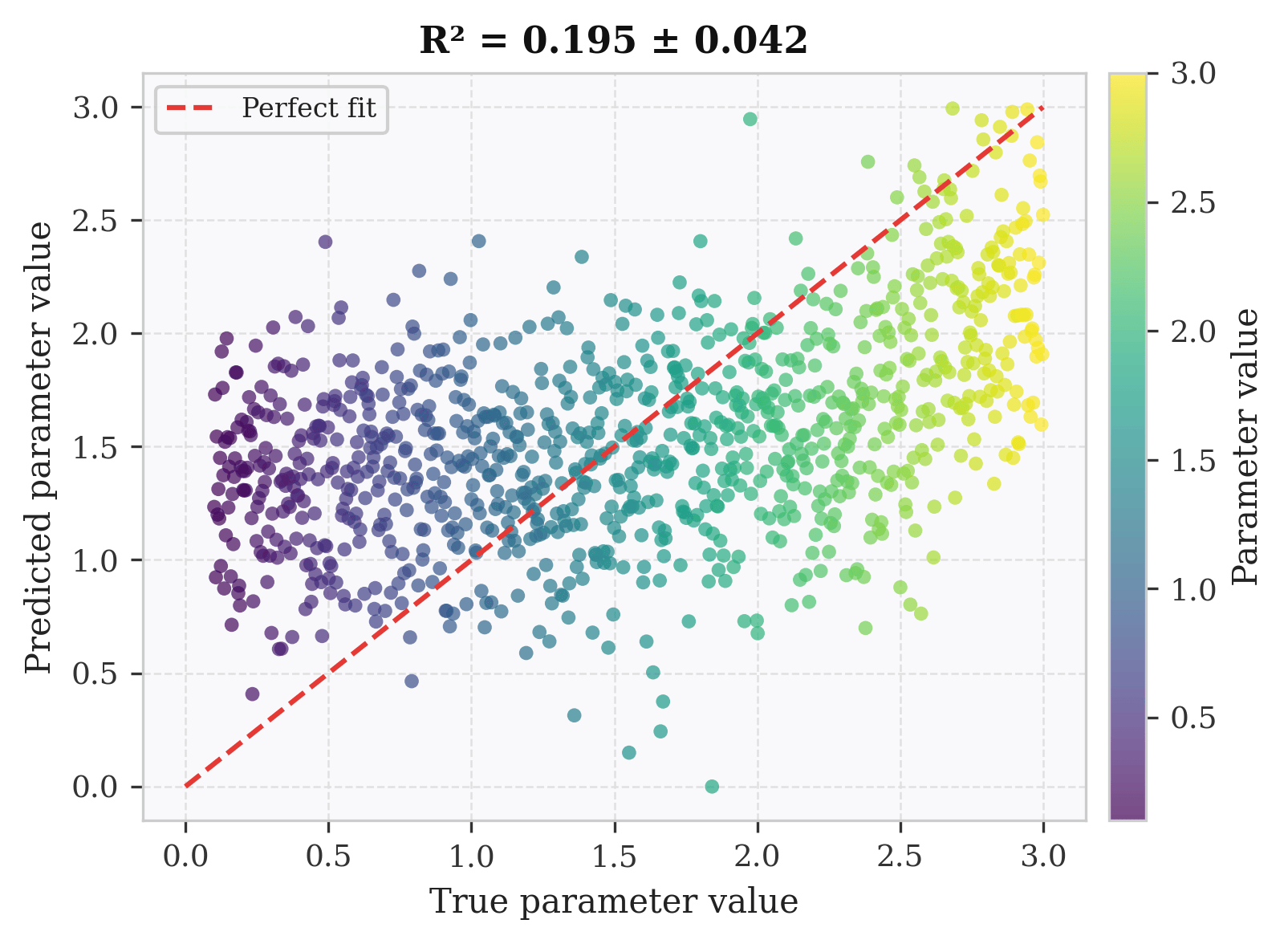}
        \caption{}
    \end{subfigure}%
    \hfill
    \begin{subfigure}[t]{0.19\linewidth}
        \centering
        \includegraphics[width=\linewidth]{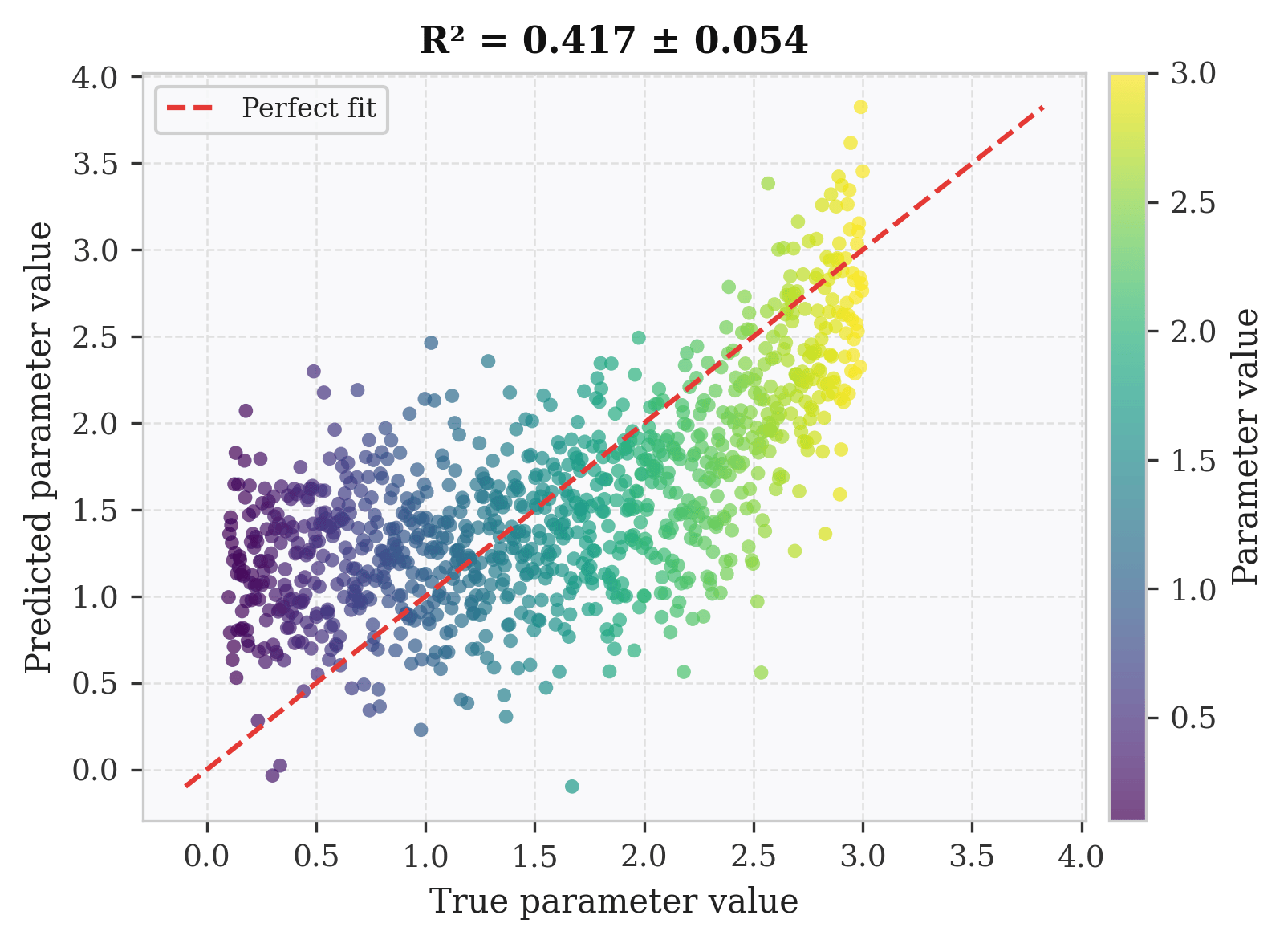}
        \caption{}
    \end{subfigure}
    \hfill
    \begin{subfigure}[t]{0.19\linewidth}
        \centering
        \includegraphics[width=\linewidth]{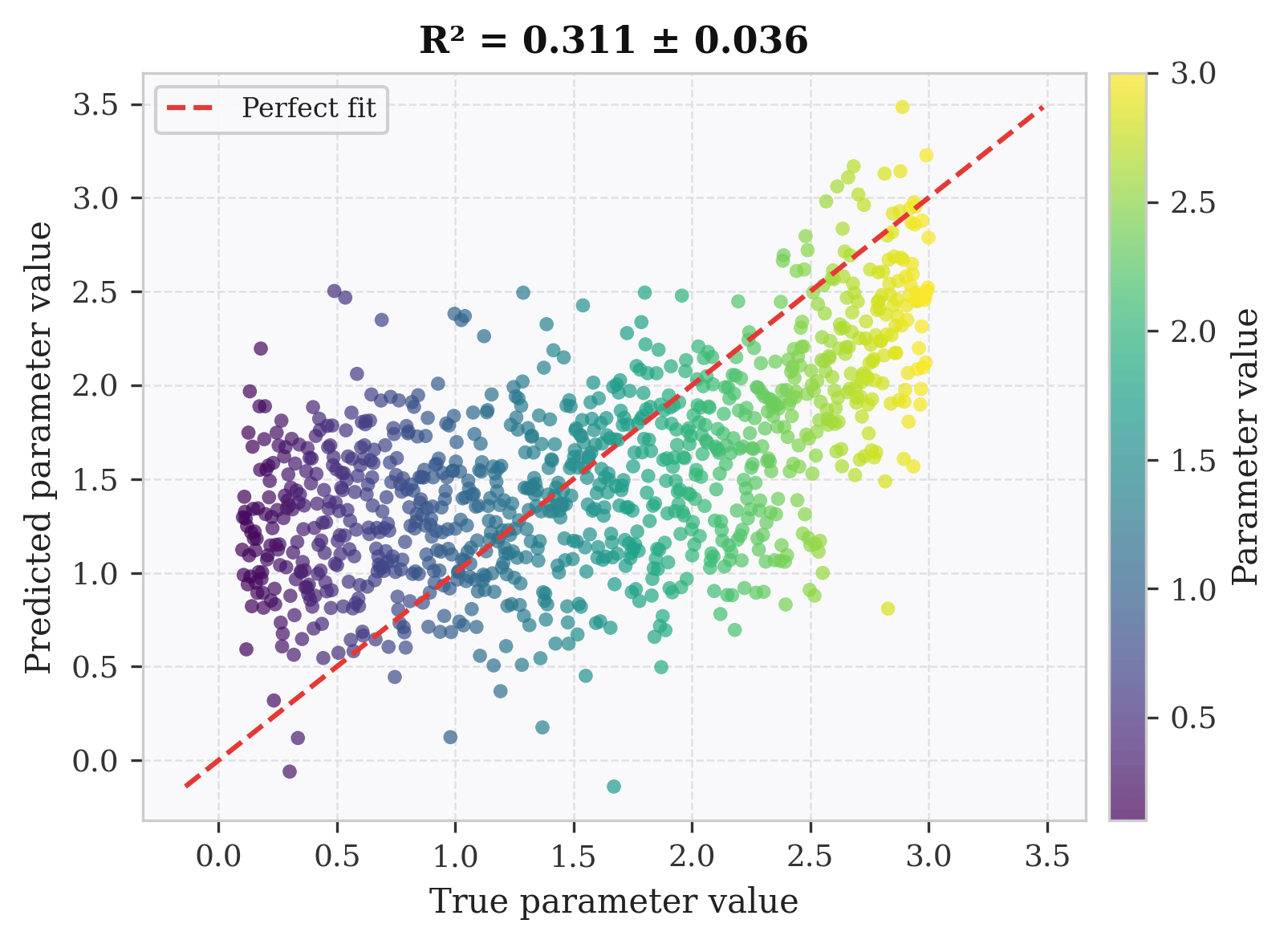}
        \caption{}
    \end{subfigure}

    \vspace{-0.2em}

    \makebox[0pt][r]{\raisebox{1.2cm}[0pt][0pt]{\rotatebox[origin=c]{90}{\textbf{LaBraM}}}\hspace{1em}}%
    \begin{subfigure}[t]{0.19\linewidth}
        \centering
        \includegraphics[width=\linewidth]{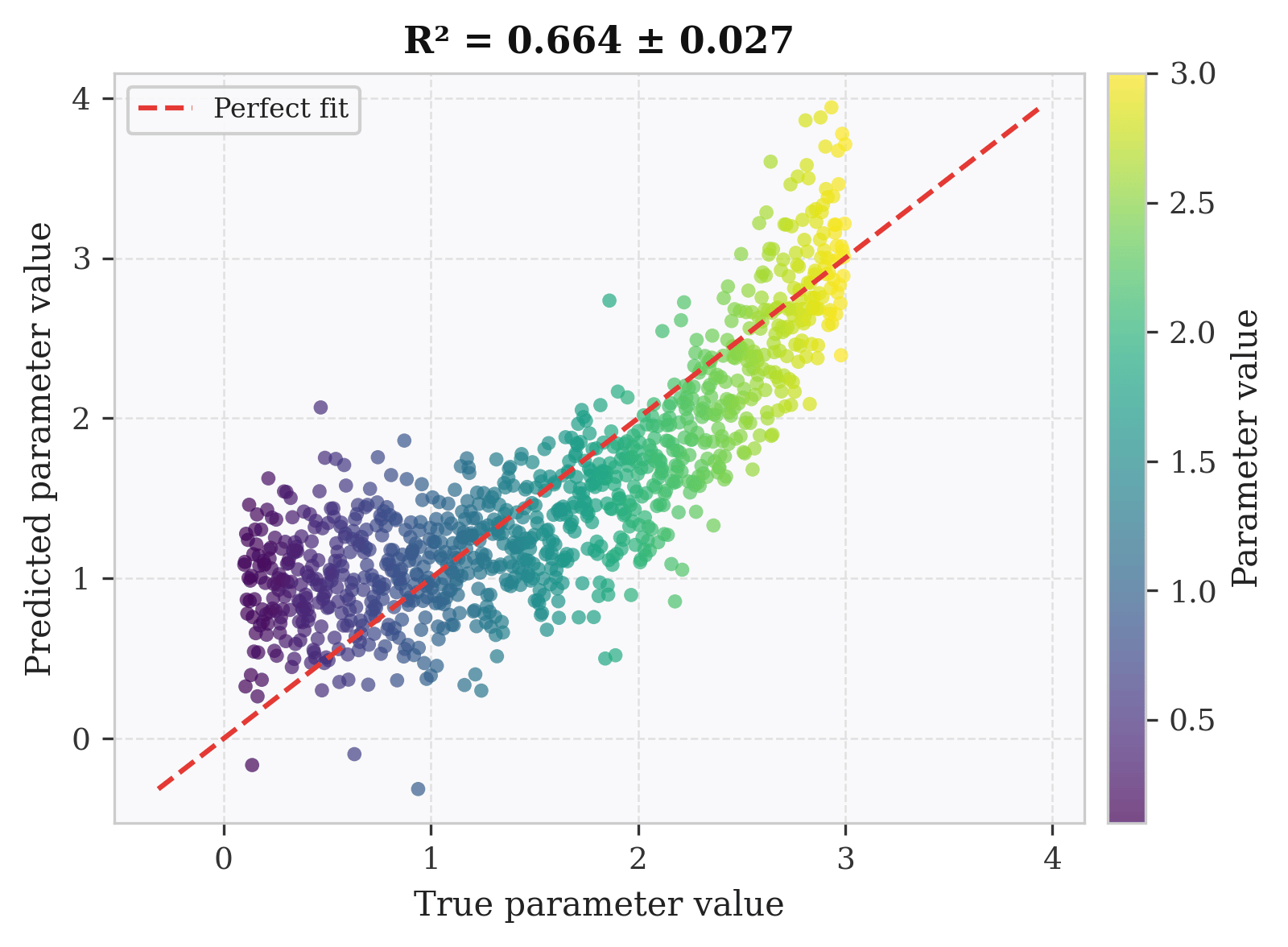}
        \caption{}
    \end{subfigure}%
    \hfill
    \begin{subfigure}[t]{0.19\linewidth}
        \centering
        \includegraphics[width=\linewidth]{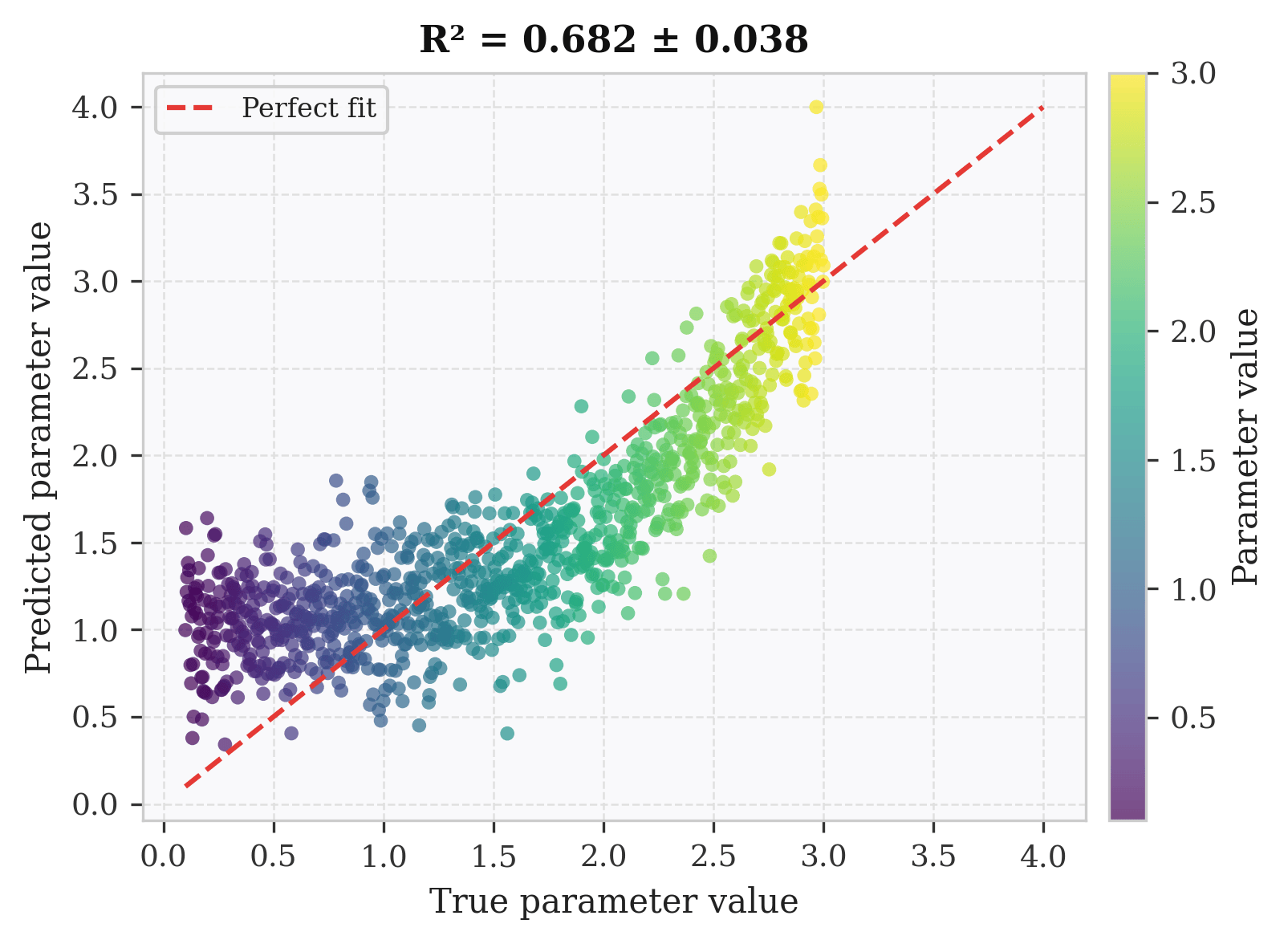}
        \caption{}
    \end{subfigure}%
    \hfill
    \begin{subfigure}[t]{0.19\linewidth}
        \centering
        \includegraphics[width=\linewidth]{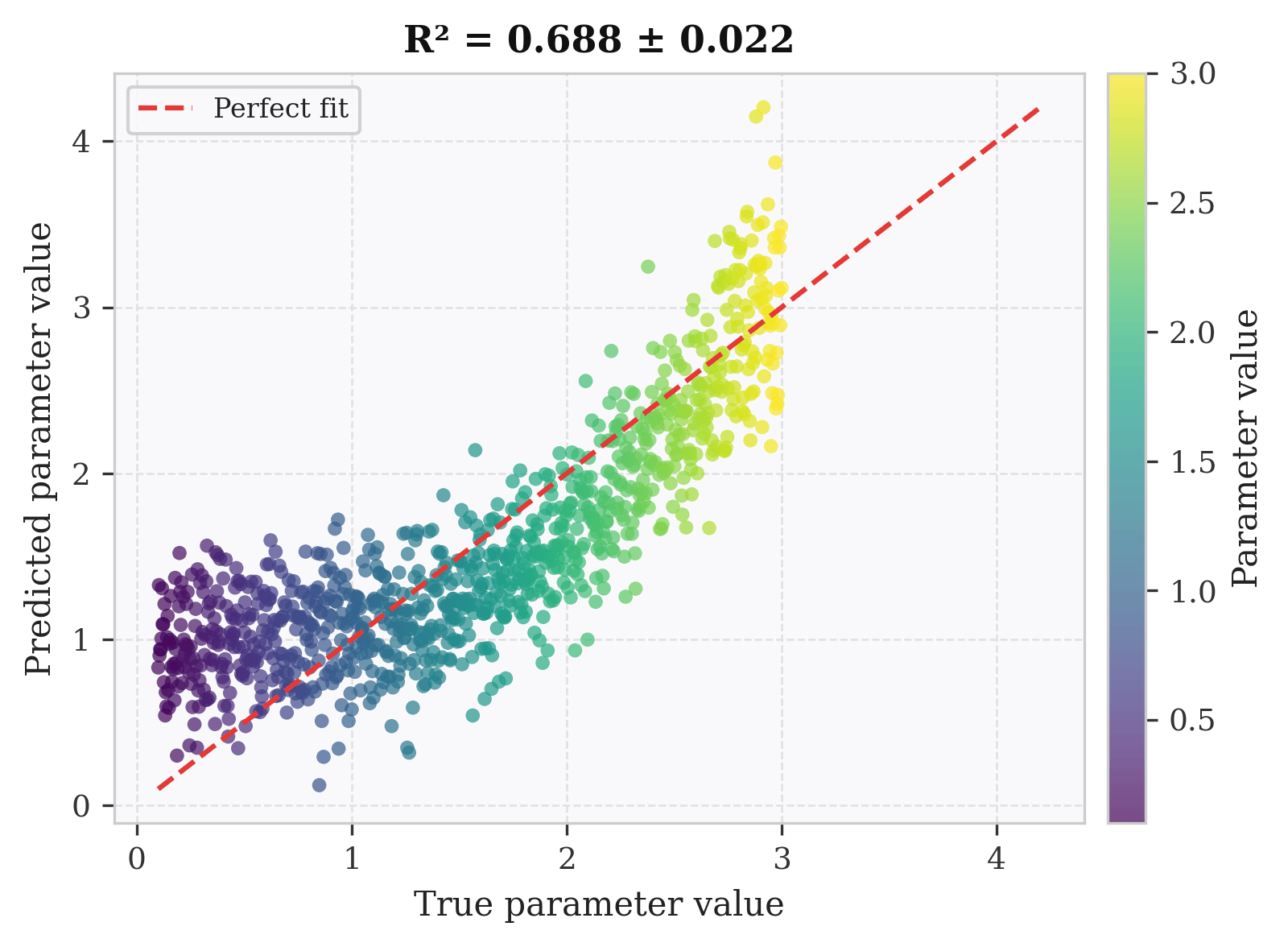}
        \caption{}
    \end{subfigure}%
    \hfill
    \begin{subfigure}[t]{0.19\linewidth}
        \centering
        \includegraphics[width=\linewidth]{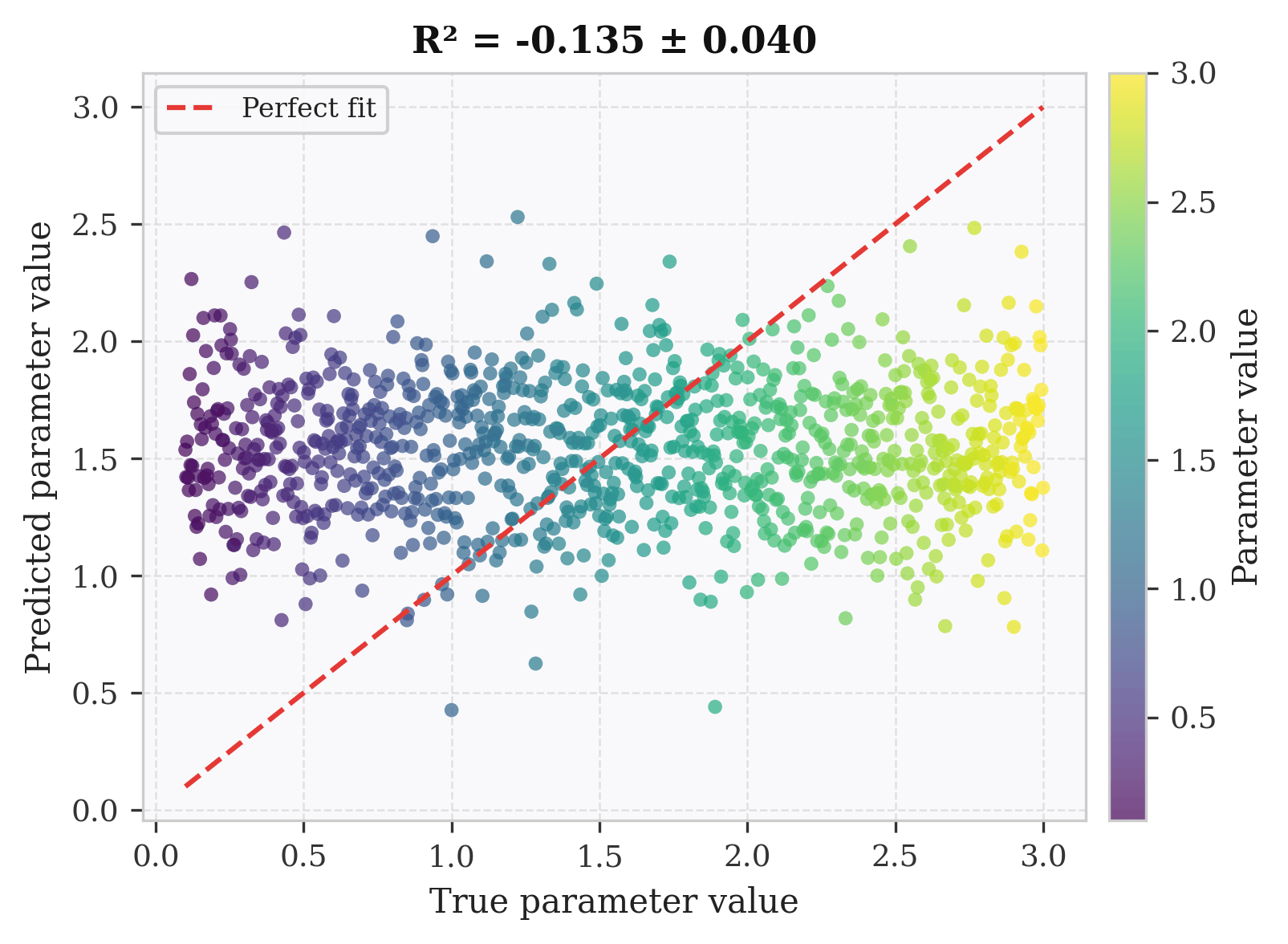}
        \caption{}
    \end{subfigure}
    \hfill
    \begin{subfigure}[t]{0.19\linewidth}
        \centering
        \includegraphics[width=\linewidth]{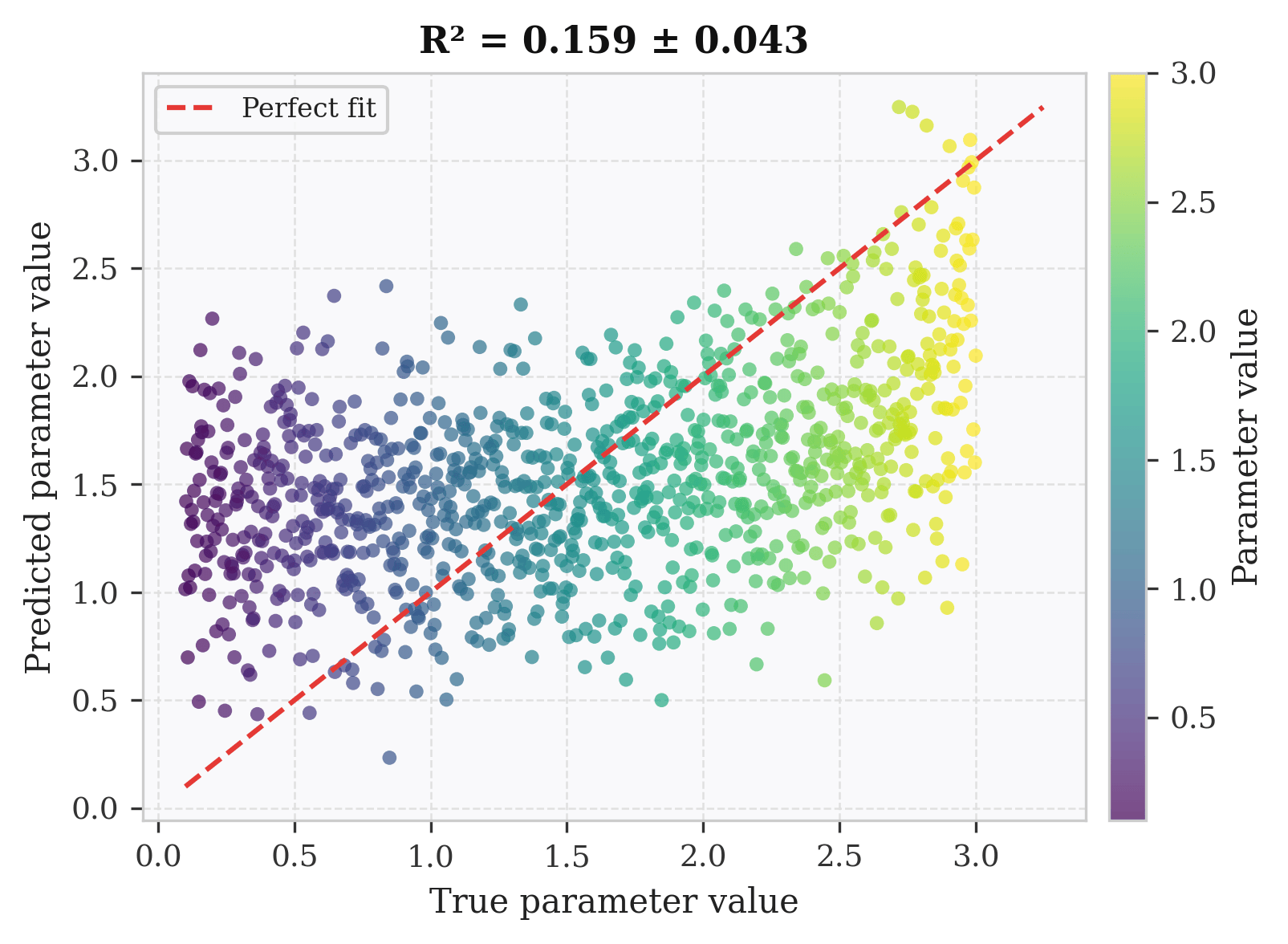}
        \caption{}
    \end{subfigure}

    \caption{
        Linear decodability comparison of Fz channel across oscillatory frequencies ($f_{\text{osc}}$) with varying power of the oscillation ($A_{\text{osc}}$) for CSBrain and LaBraM model.
    }
    \label{fig:fz_oscfreq_power}
\end{figure}

\begin{figure}[H]
    \centering
    
    % --- Column Titles ---
    \makebox[0.3\linewidth]{$\bm{\beta}$} \hfill
    \makebox[0.3\linewidth]{$\bm{A_{\text{ap}}}$} \hfill
    \makebox[0.3\linewidth]{$\bm{f_{\text{osc}}}$} \\ \vspace{0.2em}

    % --- Row 1 (CBraMod) ---
    \makebox[0pt][r]{\raisebox{1.4cm}[0pt][0pt]{\rotatebox[origin=c]{90}{\textbf{CBraMod}}}\hspace{1em}}%
    \begin{subfigure}[t]{0.3\linewidth}
        \centering
        \includegraphics[width=\linewidth]{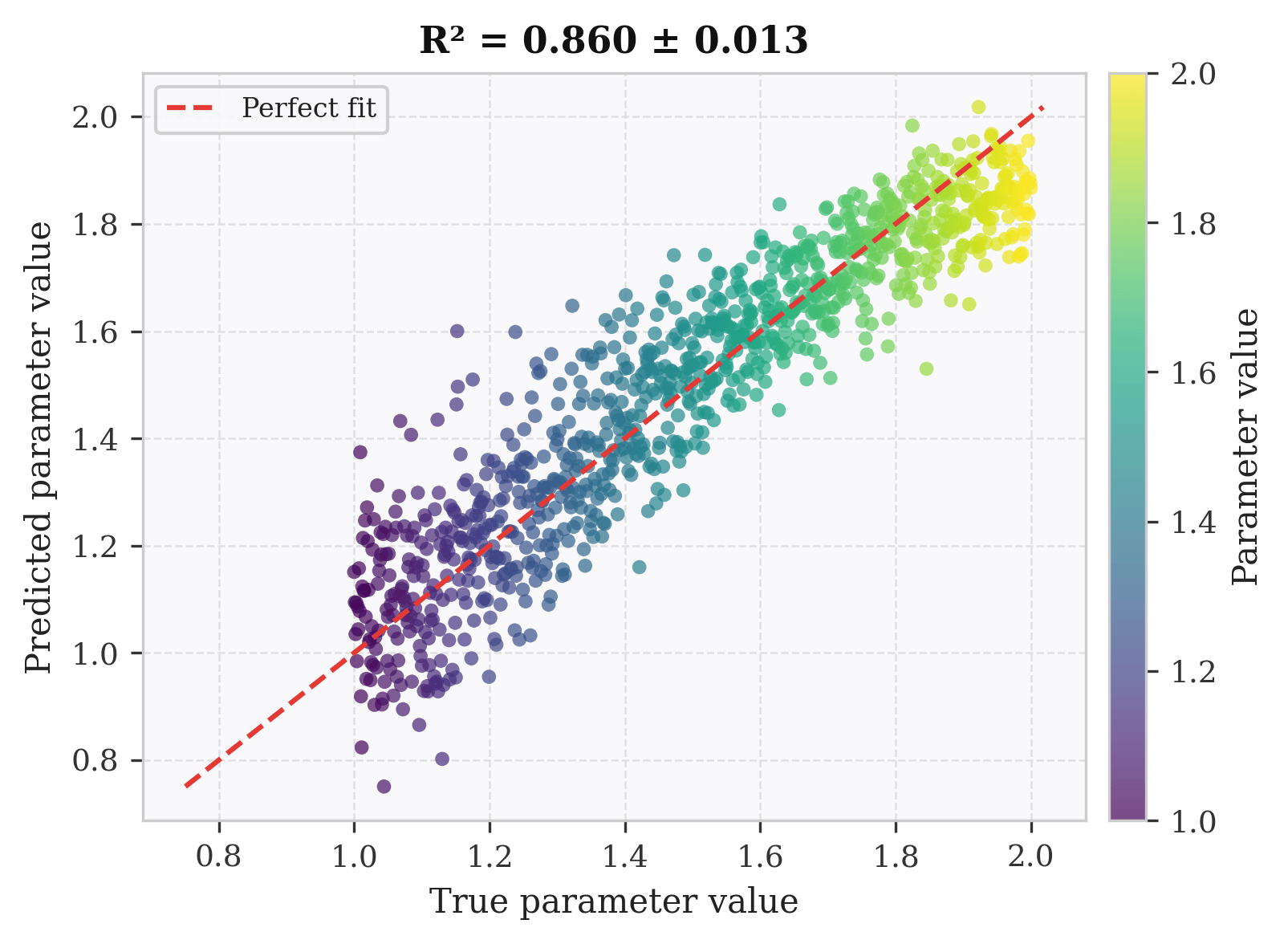}
        \caption{}
    \end{subfigure}
    \hfill
    \begin{subfigure}[t]{0.3\linewidth}
        \centering
        \includegraphics[width=\linewidth]{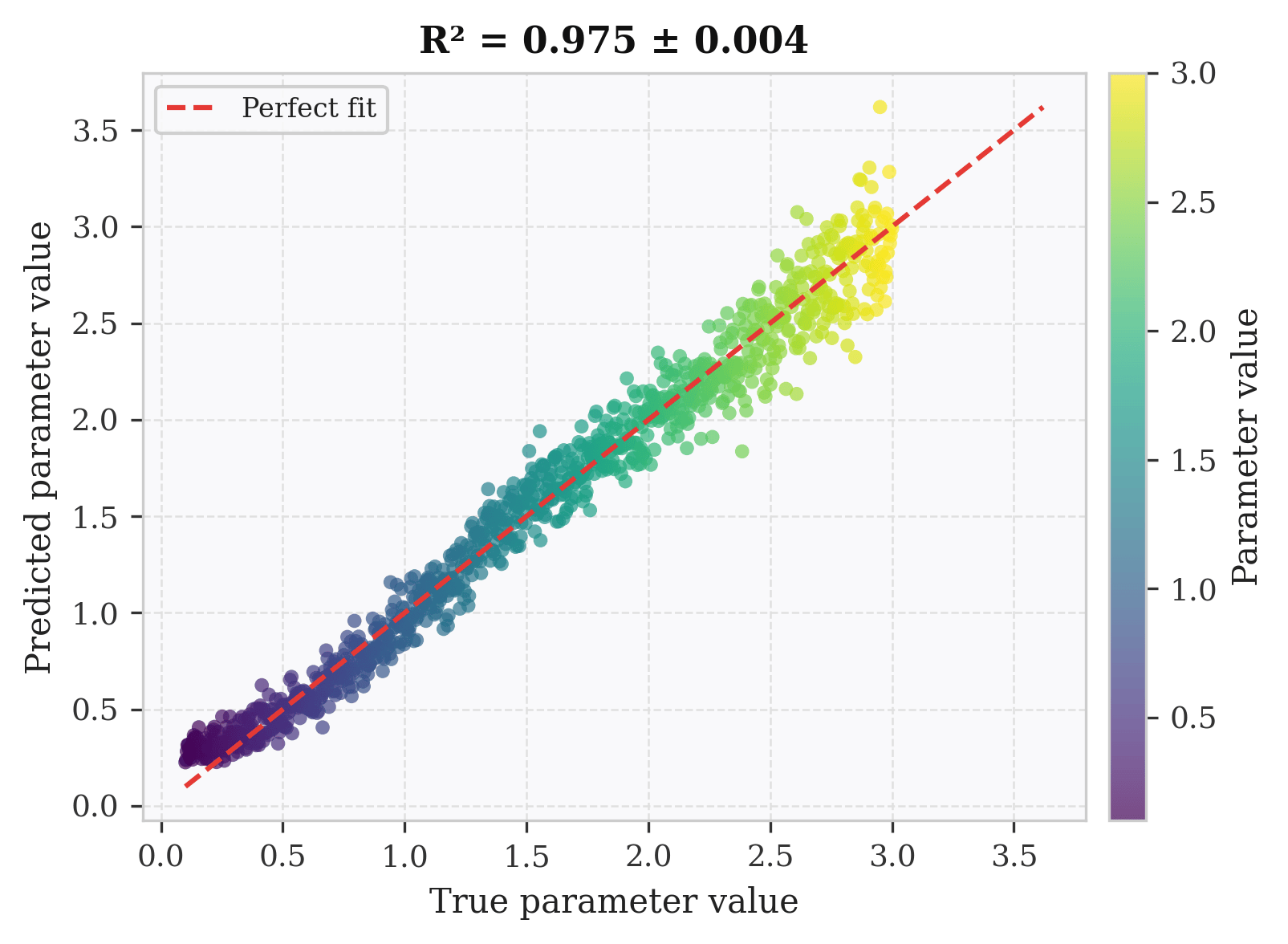}
        \caption{}
    \end{subfigure}
    \hfill
    \begin{subfigure}[t]{0.3\linewidth}
        \centering
        \includegraphics[width=\linewidth]{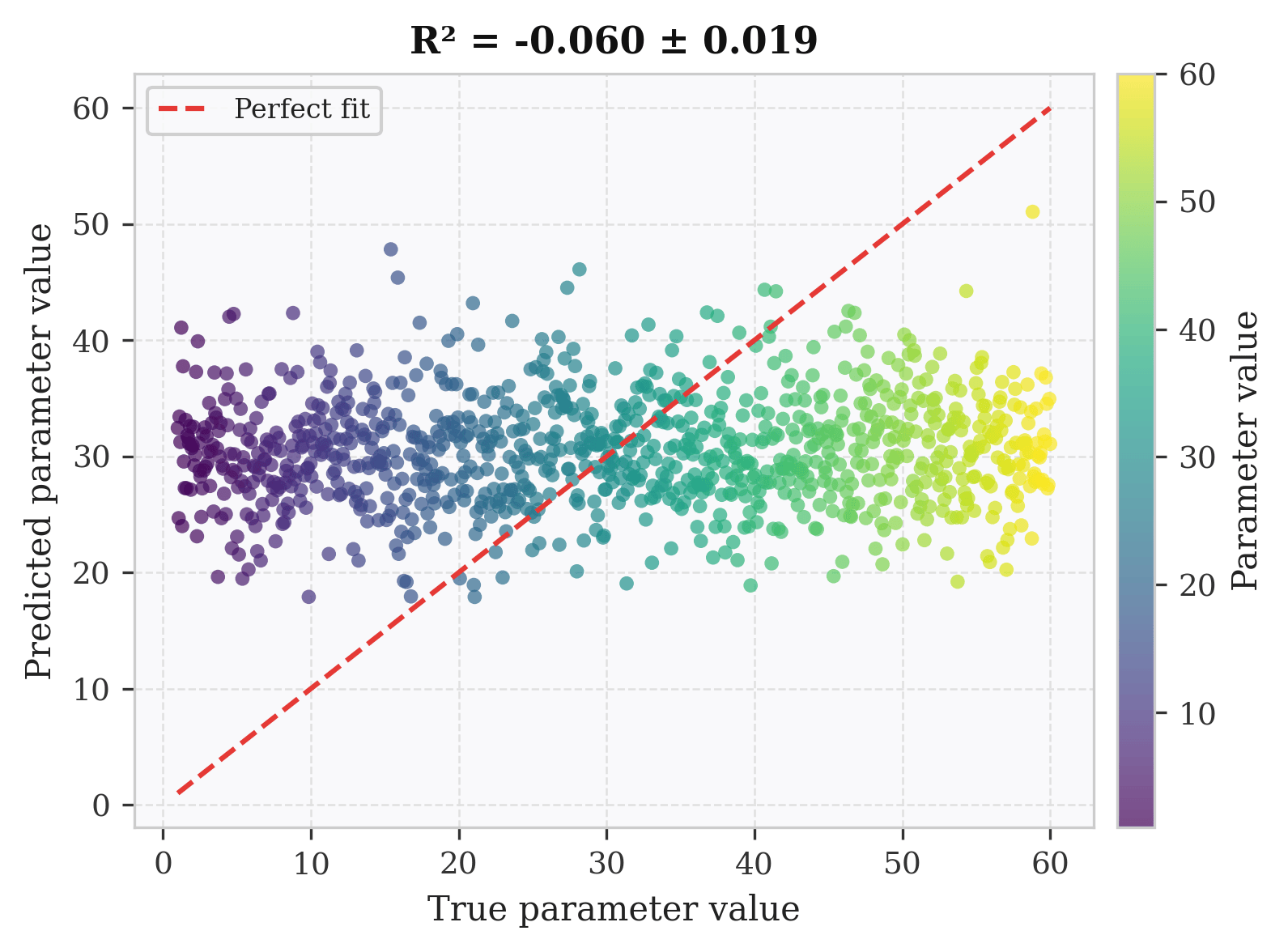}
        \caption{}
    \end{subfigure}

    \vspace{-0.5cm}

    % --- Row 2 (CSBrain) ---
    \makebox[0pt][r]{\raisebox{1.4cm}[0pt][0pt]{\rotatebox[origin=c]{90}{\textbf{CSBrain}}}\hspace{1em}}%
    \begin{subfigure}[t]{0.3\linewidth}
        \centering
        \includegraphics[width=\linewidth]{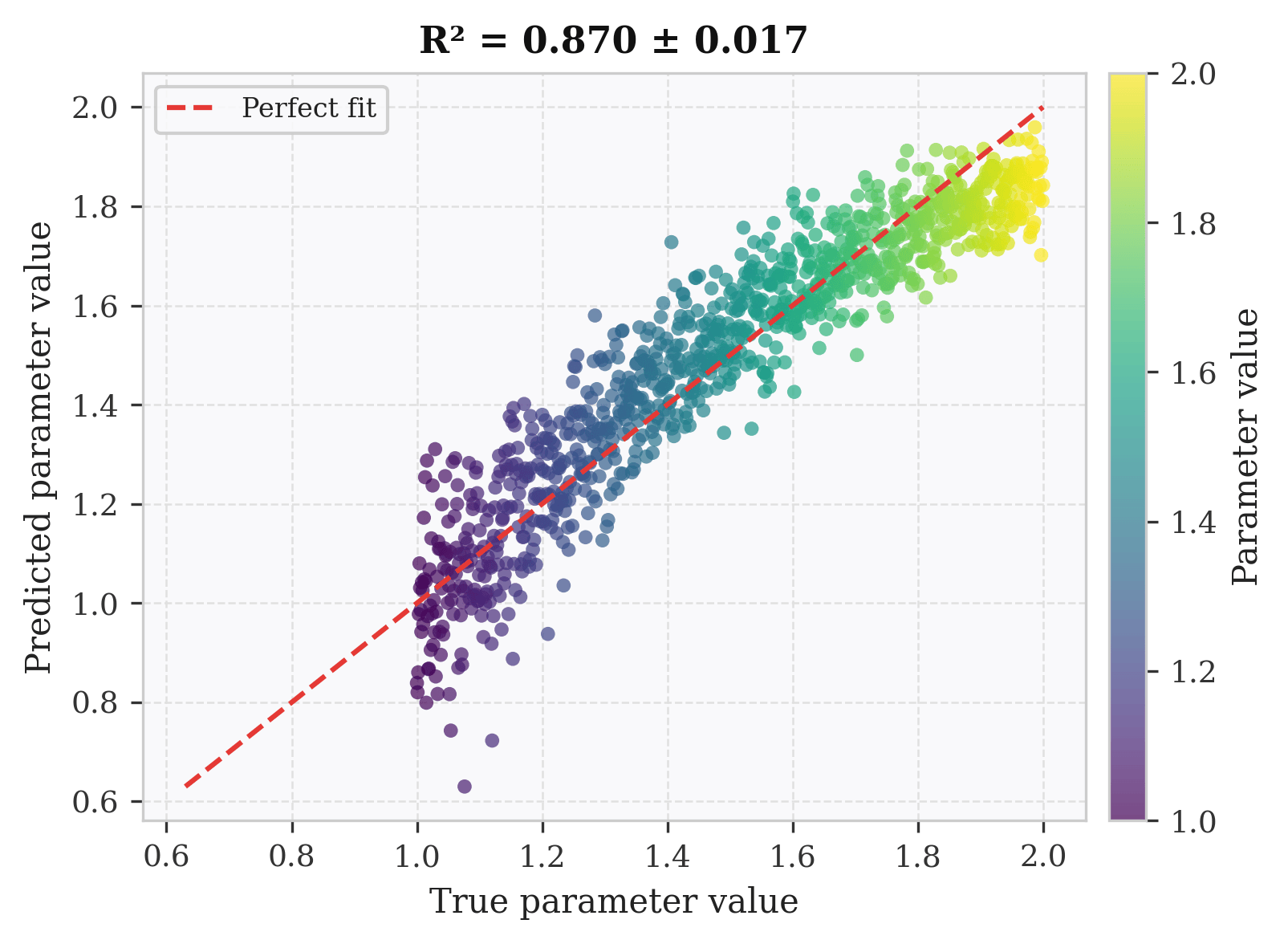}
        \caption{}
    \end{subfigure}
    \hfill
    \begin{subfigure}[t]{0.3\linewidth}
        \centering
        \includegraphics[width=\linewidth]{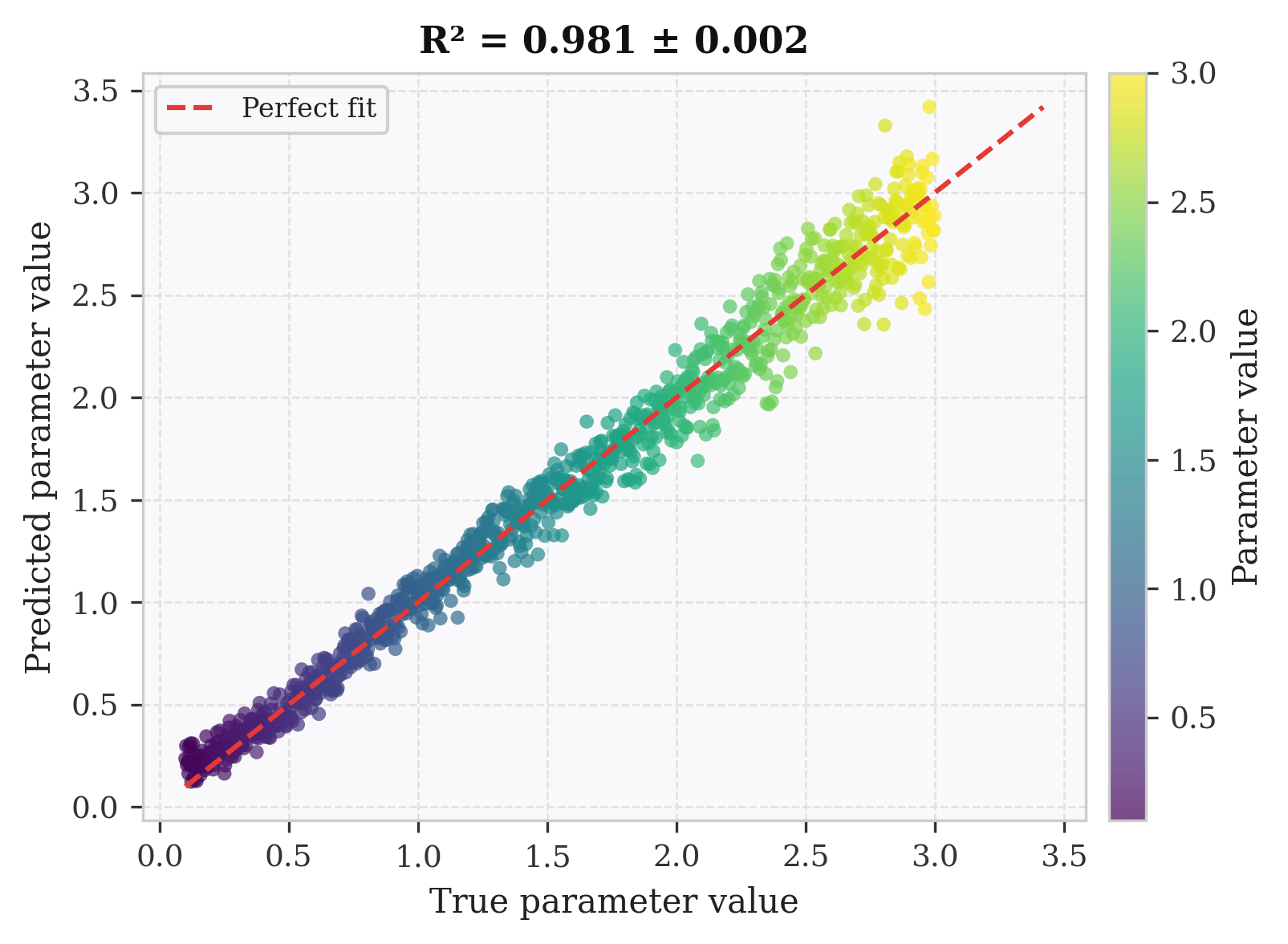}
        \caption{}
    \end{subfigure}
    \hfill
    \begin{subfigure}[t]{0.3\linewidth}
        \centering
        \includegraphics[width=\linewidth]{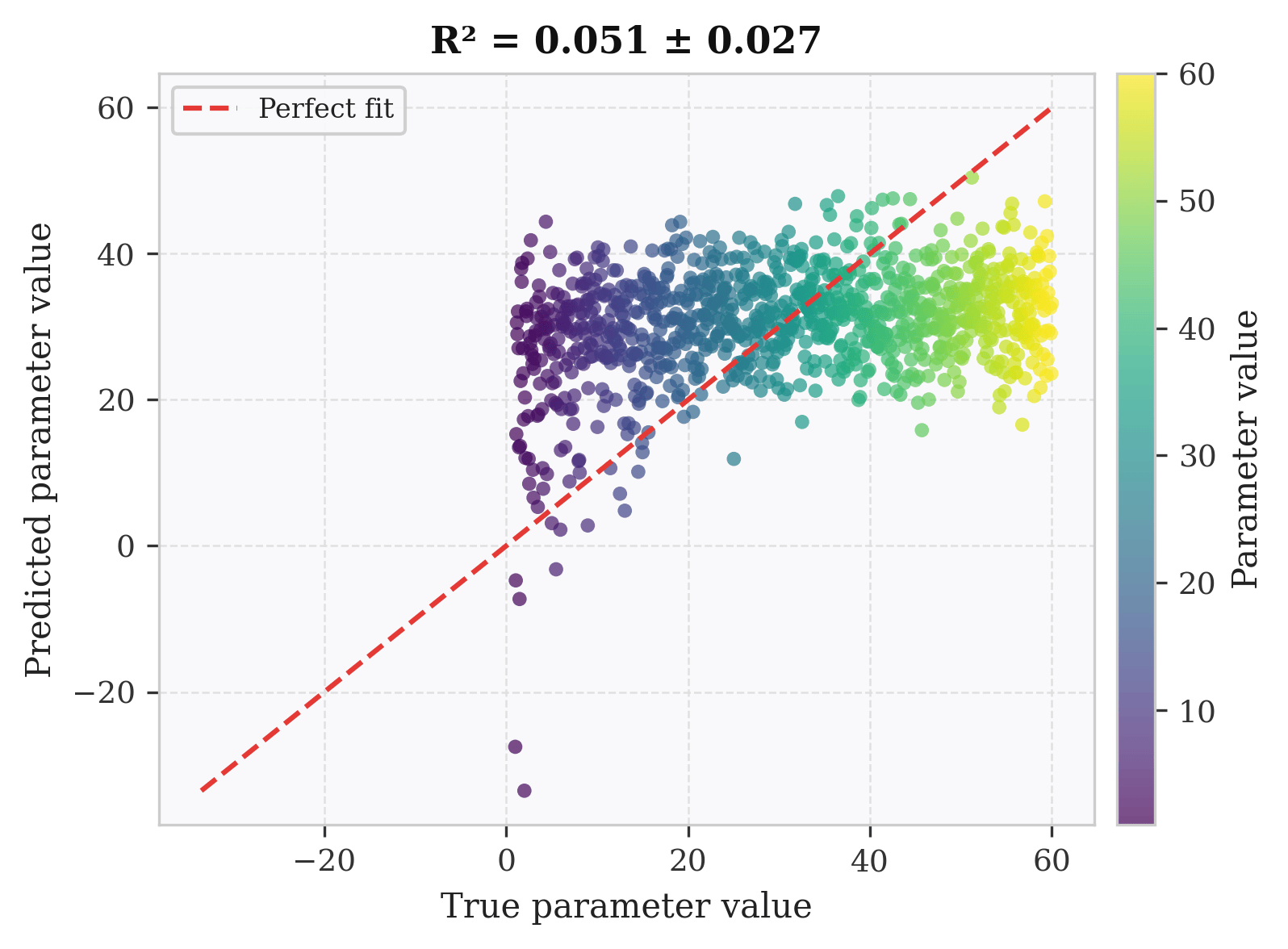}
        \caption{}
    \end{subfigure}

    \vspace{-0.5cm}

    % --- Row 3 (LaBraM) ---
    \makebox[0pt][r]{\raisebox{1.2cm}[0pt][0pt]{\rotatebox[origin=c]{90}{\textbf{LaBraM}}}\hspace{1em}}%
    \begin{subfigure}[t]{0.3\linewidth}
        \centering
        \includegraphics[width=\linewidth]{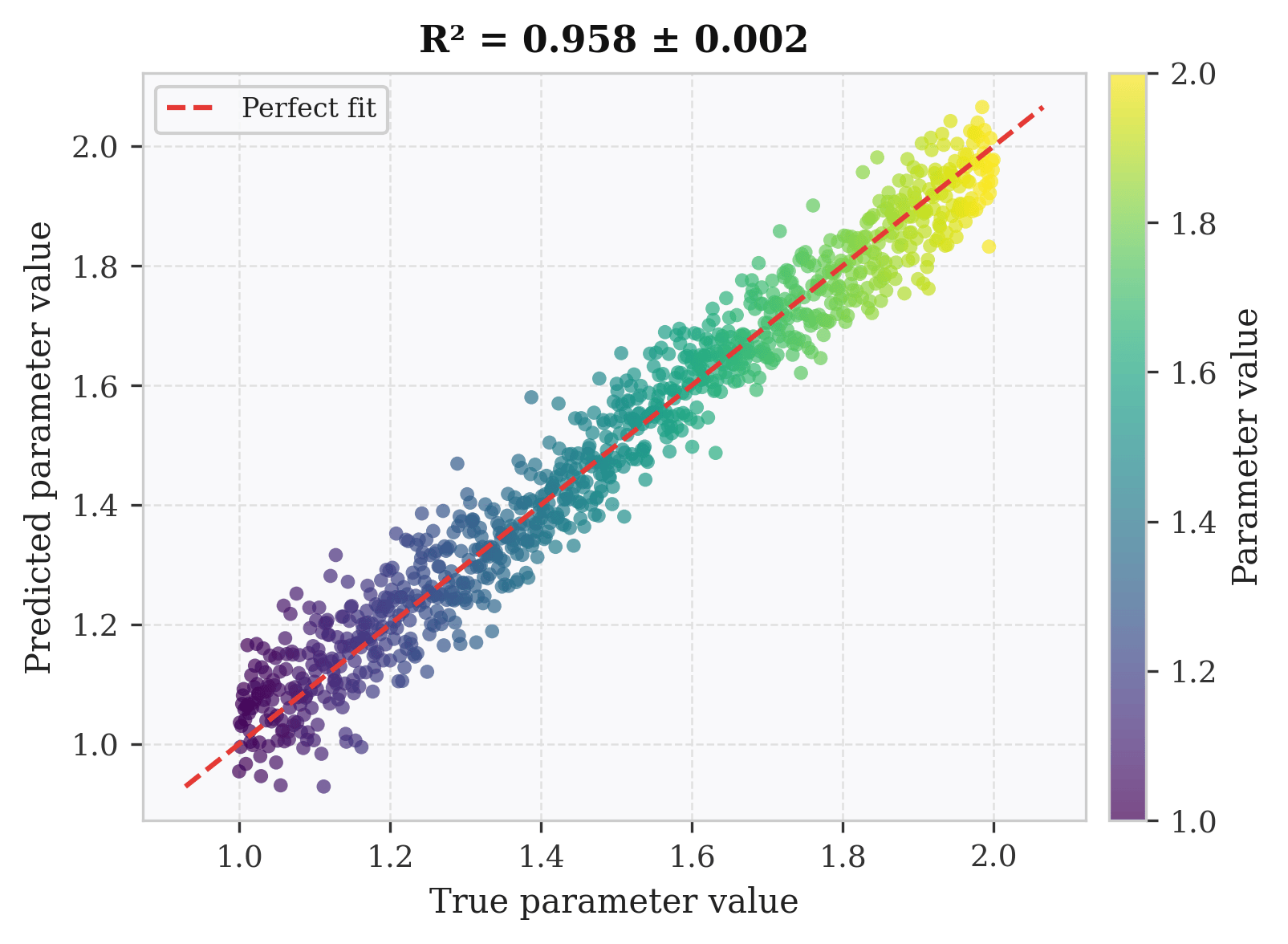}
        \caption{}
    \end{subfigure}
    \hfill
    \begin{subfigure}[t]{0.3\linewidth}
        \centering
        \includegraphics[width=\linewidth]{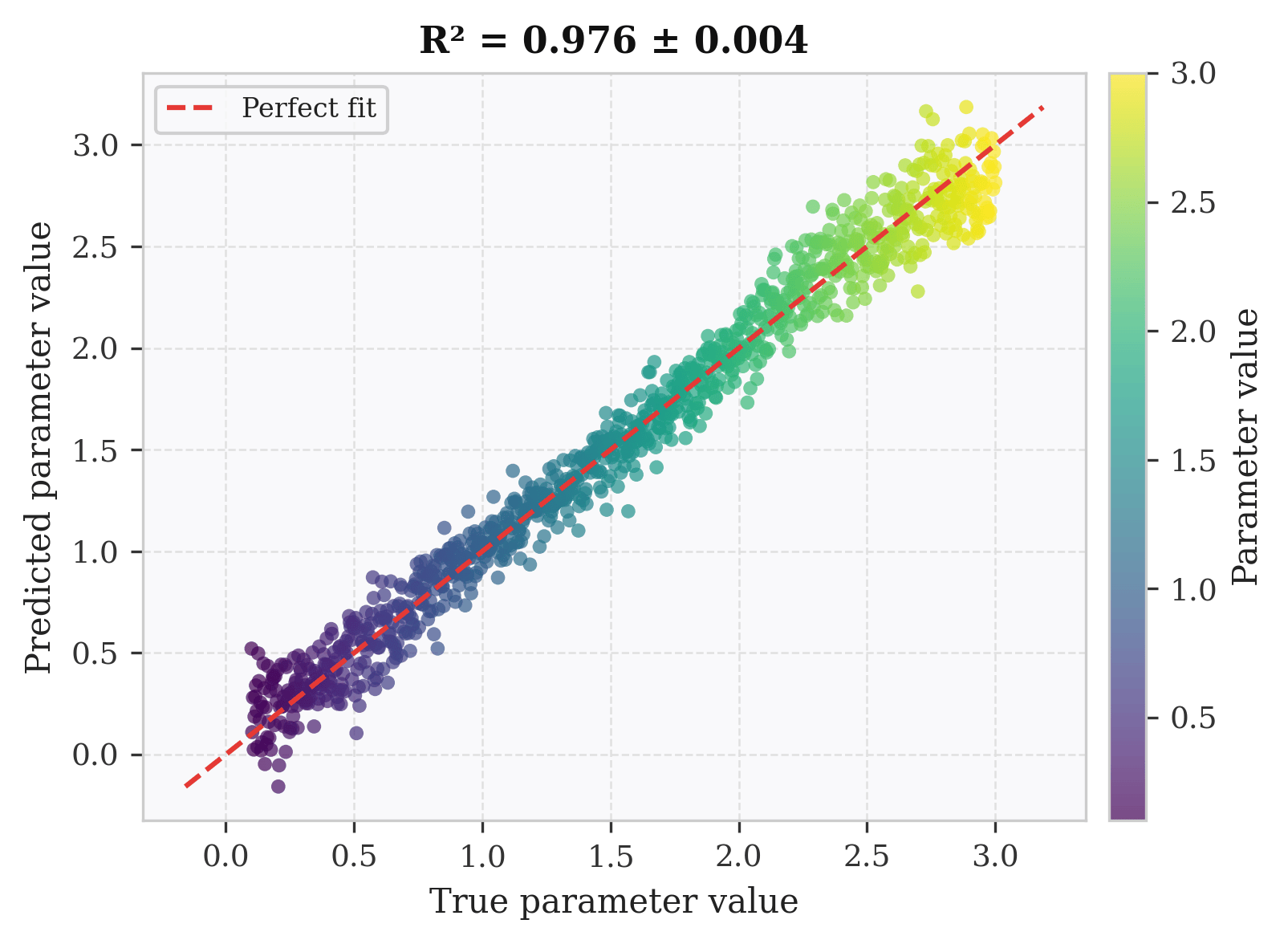}
        \caption{}
    \end{subfigure}
    \hfill
    \begin{subfigure}[t]{0.3\linewidth}
        \centering
        \includegraphics[width=\linewidth]{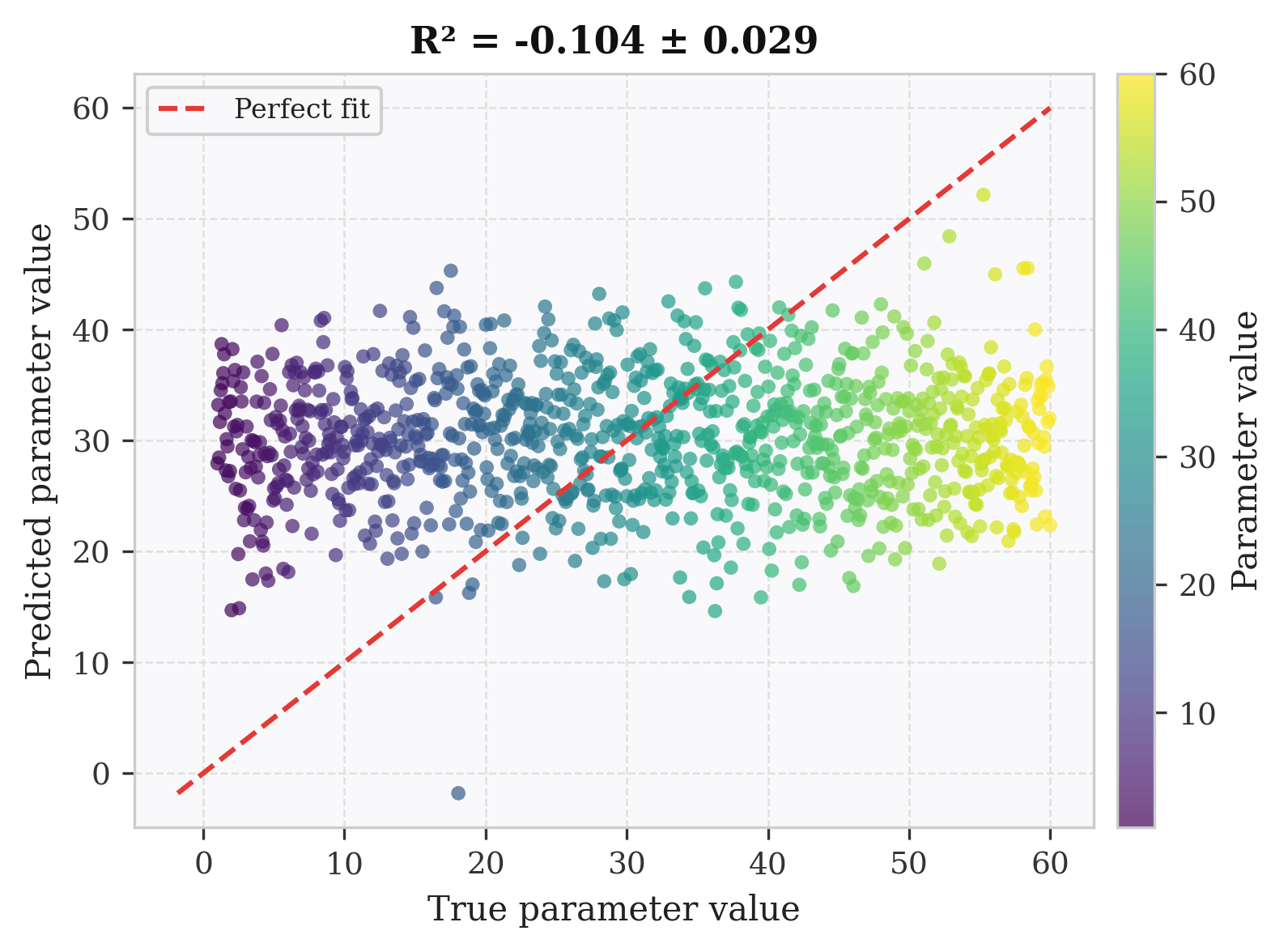}
        \caption{}
    \end{subfigure}

    \caption{
    Linear decodability for Pz channel across three foundation models (CBraMod, CSBrain, LaBraM) for Aperiodic Exponent ($\beta$), Aperiodic Offset ($A_{\text{ap}}$) and Oscillation frequency ($f_{\text{osc}}$).
    }
    \label{fig:Pz}
\end{figure}

\begin{figure}[H]
    \centering
    
    % --- Column Titles ---
    \makebox[0.19\linewidth]{\textbf{10Hz}} \hfill
    \makebox[0.19\linewidth]{\textbf{20Hz}} \hfill
    \makebox[0.19\linewidth]{\textbf{30Hz}} \hfill
    \makebox[0.19\linewidth]{\textbf{40Hz}} \hfill
    \makebox[0.19\linewidth]{\textbf{50Hz}}\\ \vspace{0.2em}

    \makebox[0pt][r]{\raisebox{1.2cm}[0pt][0pt]{\rotatebox[origin=c]{90}{\textbf{CSBrain}}}\hspace{1em}}%
    \begin{subfigure}[t]{0.19\linewidth}
        \centering
        \includegraphics[width=\linewidth]{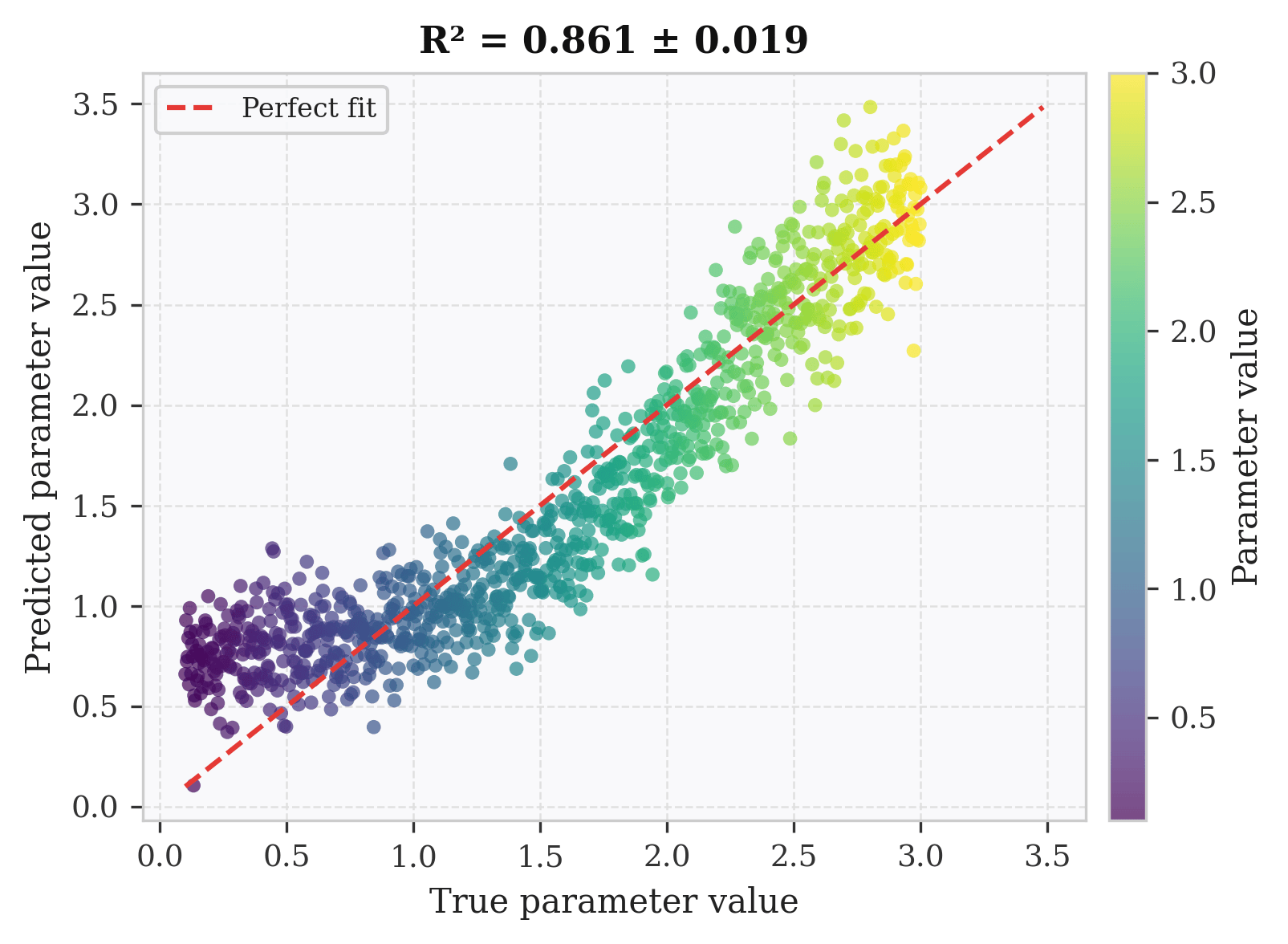}
        \caption{}
    \end{subfigure}%
    \hfill
    \begin{subfigure}[t]{0.19\linewidth}
        \centering
        \includegraphics[width=\linewidth]{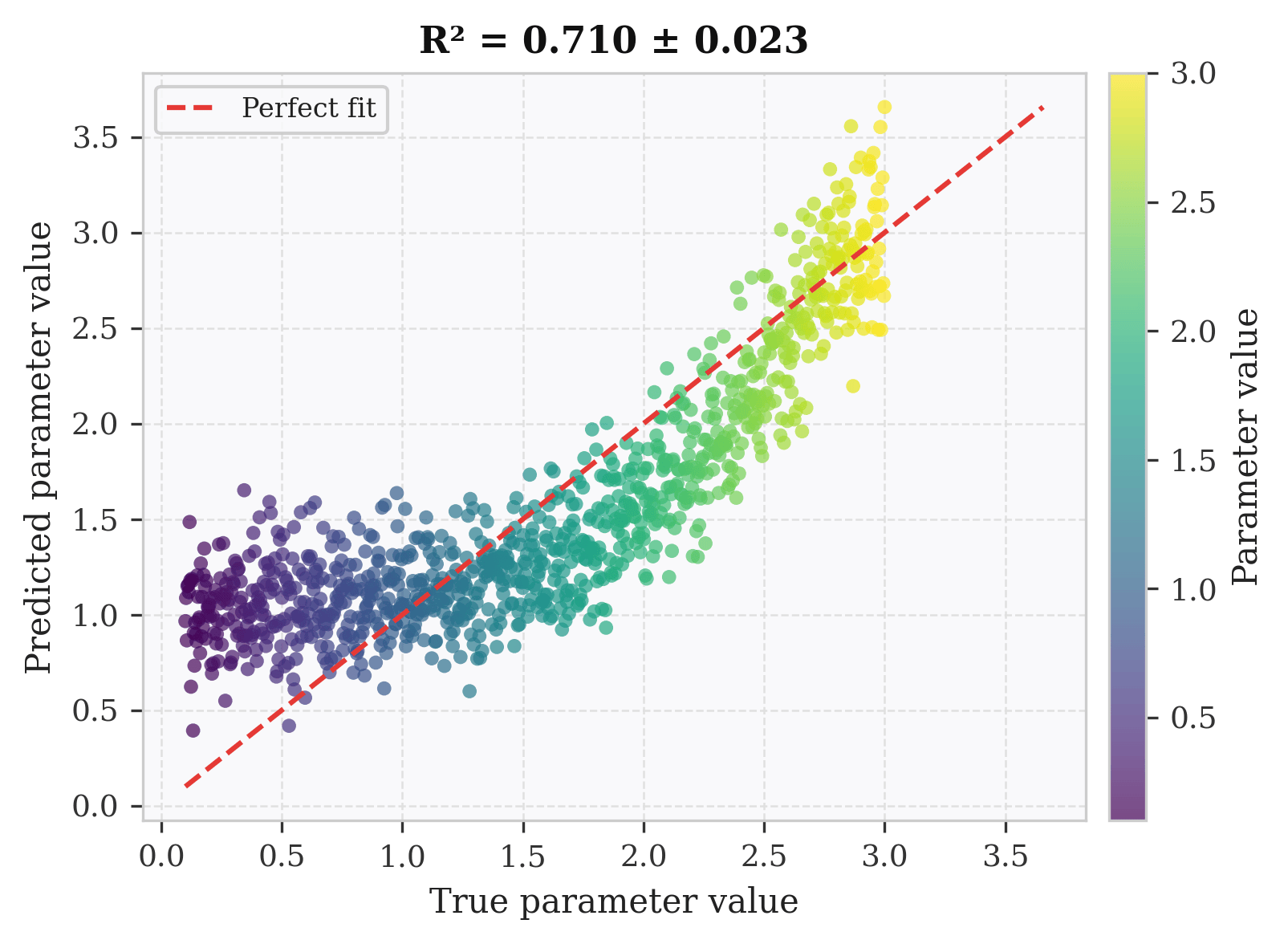}
        \caption{}
    \end{subfigure}%
    \hfill
    \begin{subfigure}[t]{0.19\linewidth}
        \centering
        \includegraphics[width=\linewidth]{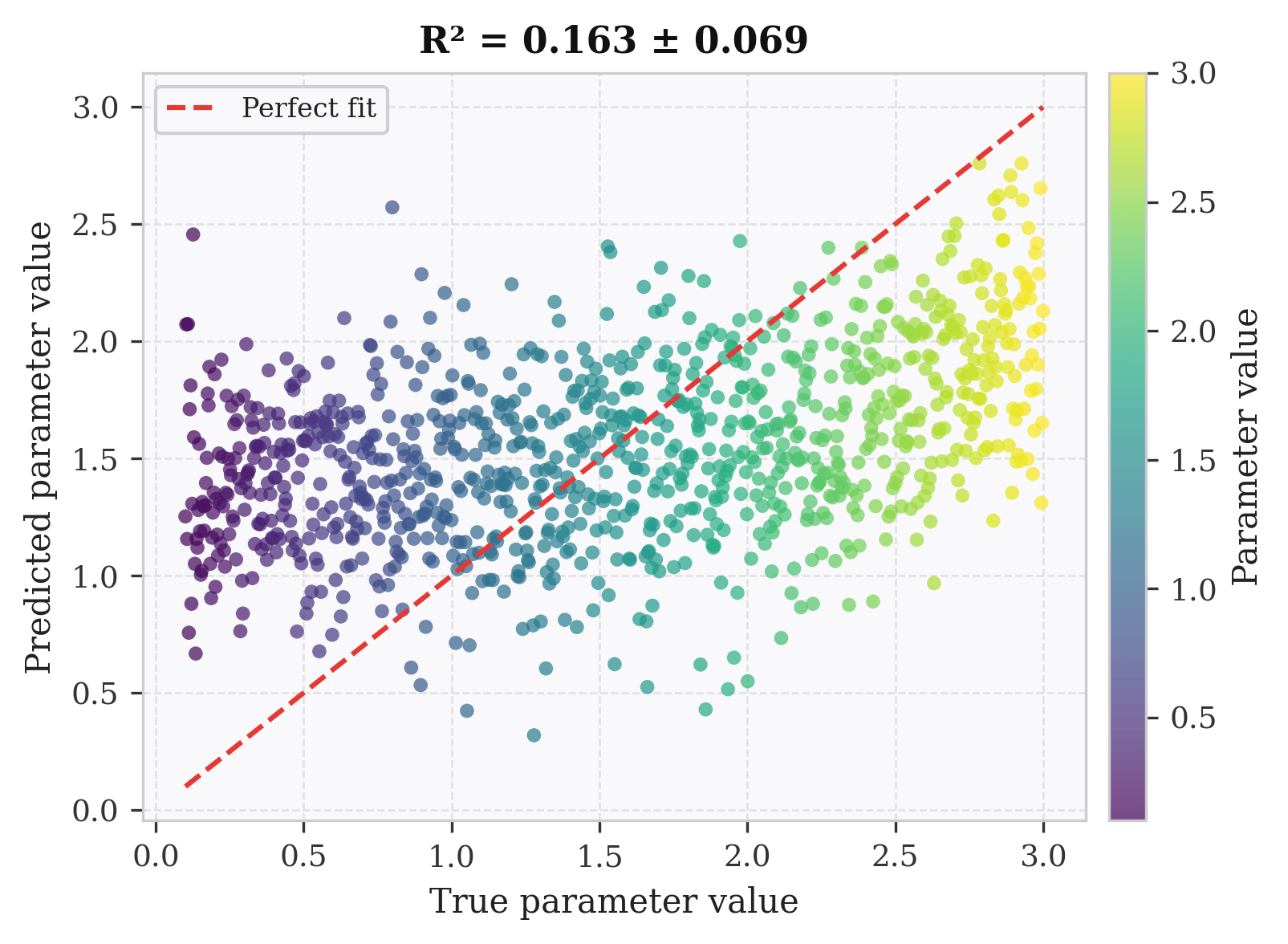}
        \caption{}
    \end{subfigure}%
    \hfill
    \begin{subfigure}[t]{0.19\linewidth}
        \centering
        \includegraphics[width=\linewidth]{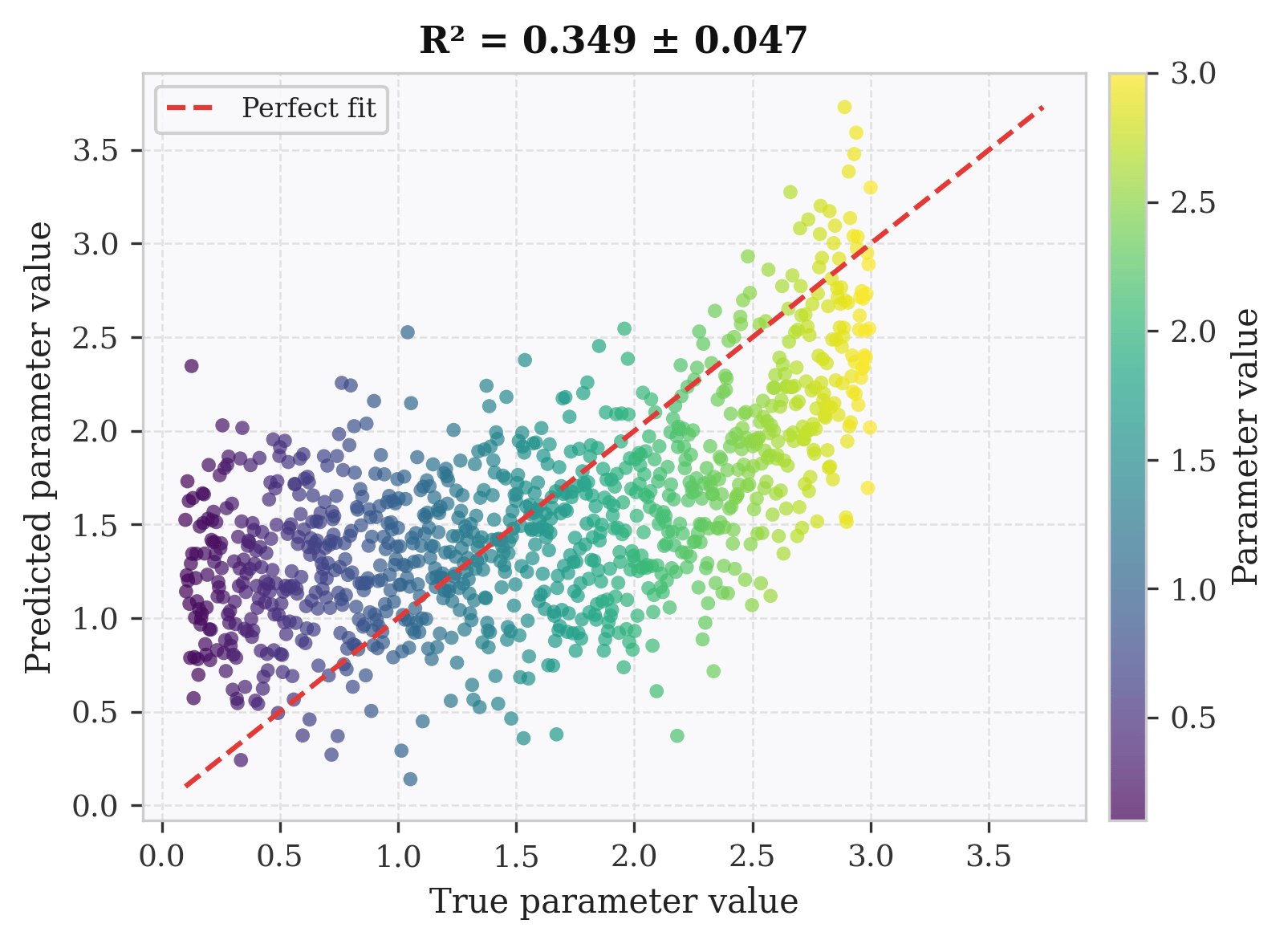}
        \caption{}
    \end{subfigure}
    \hfill
    \begin{subfigure}[t]{0.19\linewidth}
        \centering
        \includegraphics[width=\linewidth]{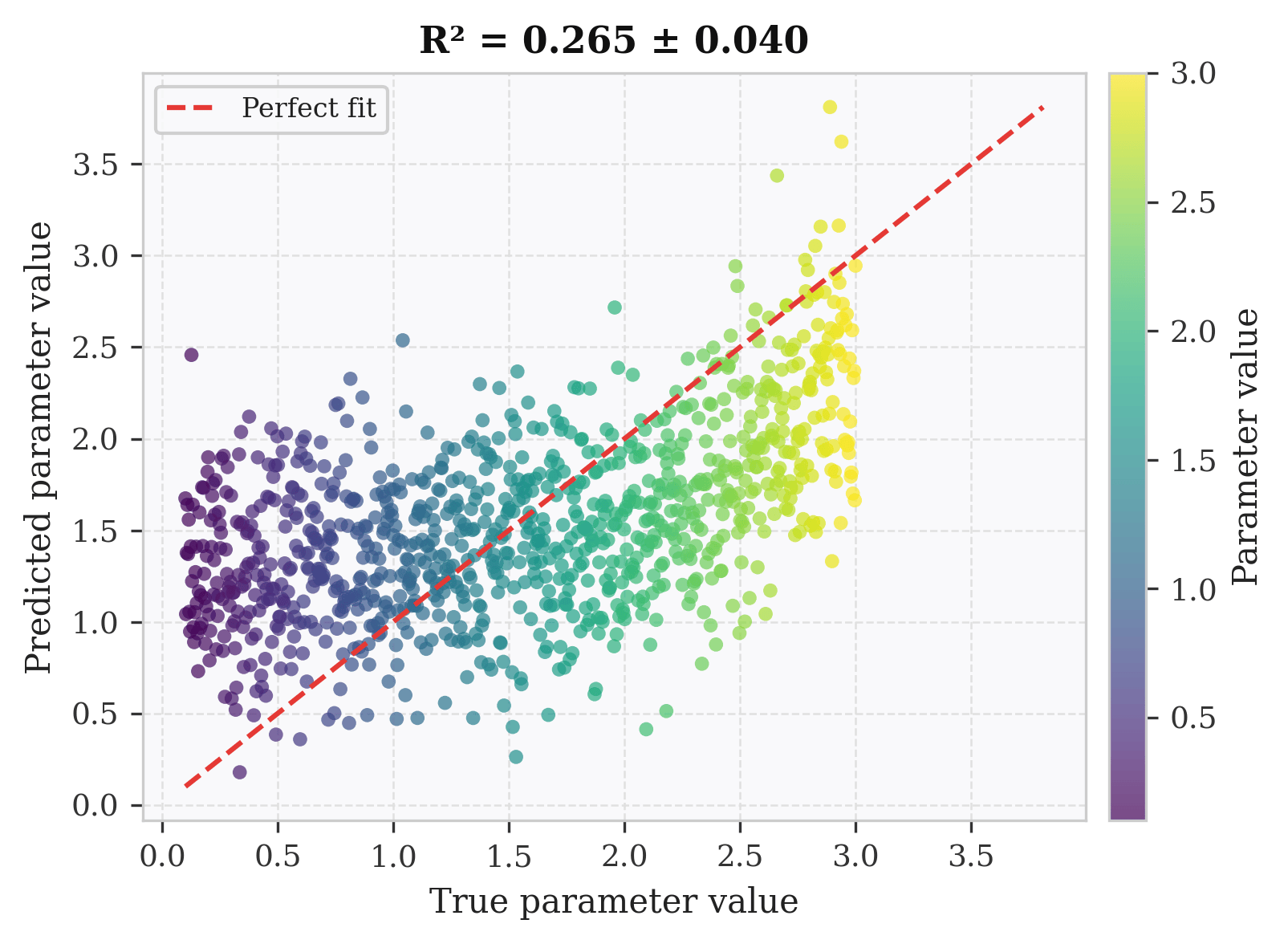}
        \caption{}
    \end{subfigure}

    \vspace{-0.2em}

    \makebox[0pt][r]{\raisebox{1.2cm}[0pt][0pt]{\rotatebox[origin=c]{90}{\textbf{LaBraM}}}\hspace{1em}}%
    \begin{subfigure}[t]{0.19\linewidth}
        \centering
        \includegraphics[width=\linewidth]{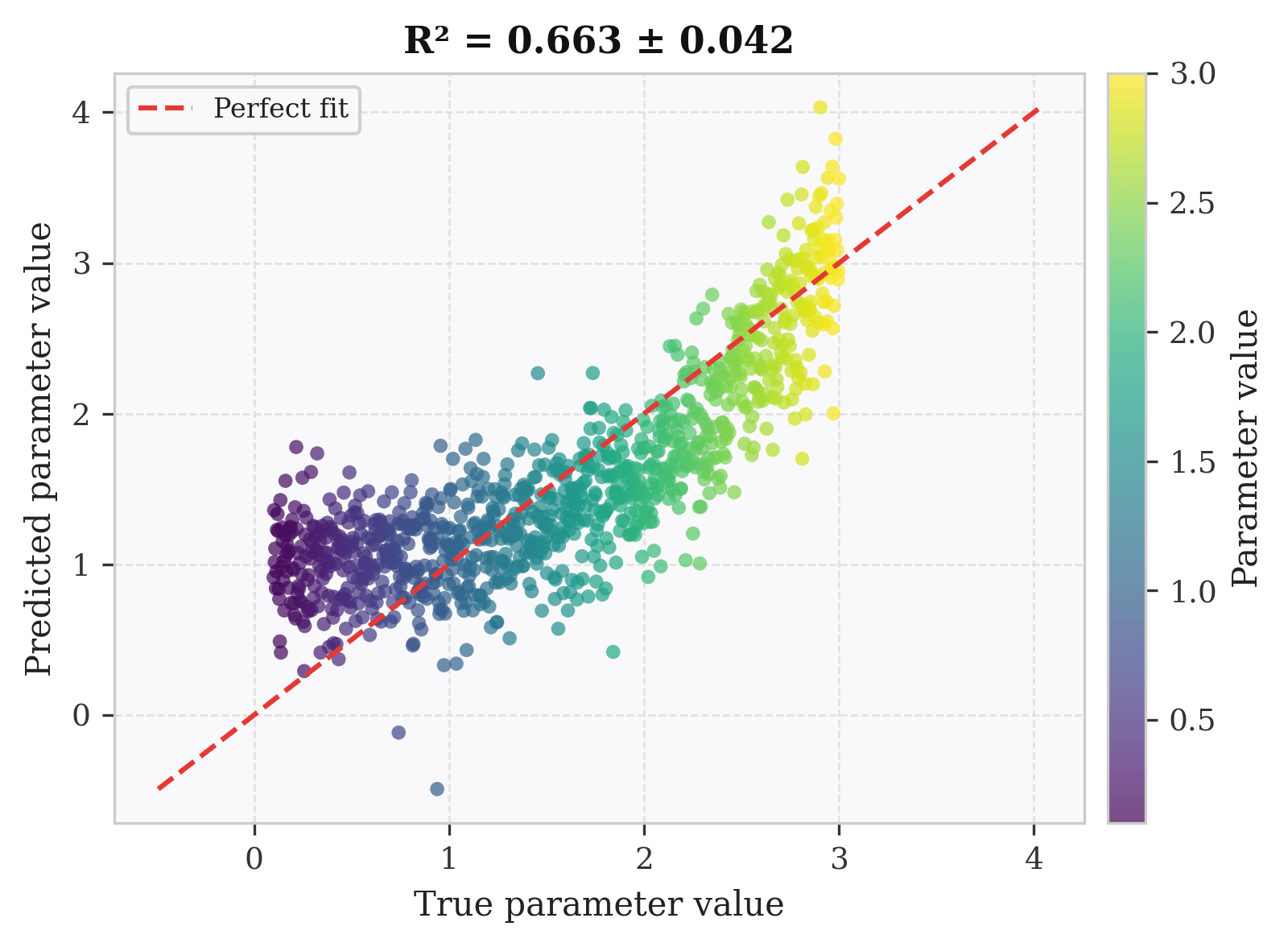}
        \caption{}
    \end{subfigure}%
    \hfill
    \begin{subfigure}[t]{0.19\linewidth}
        \centering
        \includegraphics[width=\linewidth]{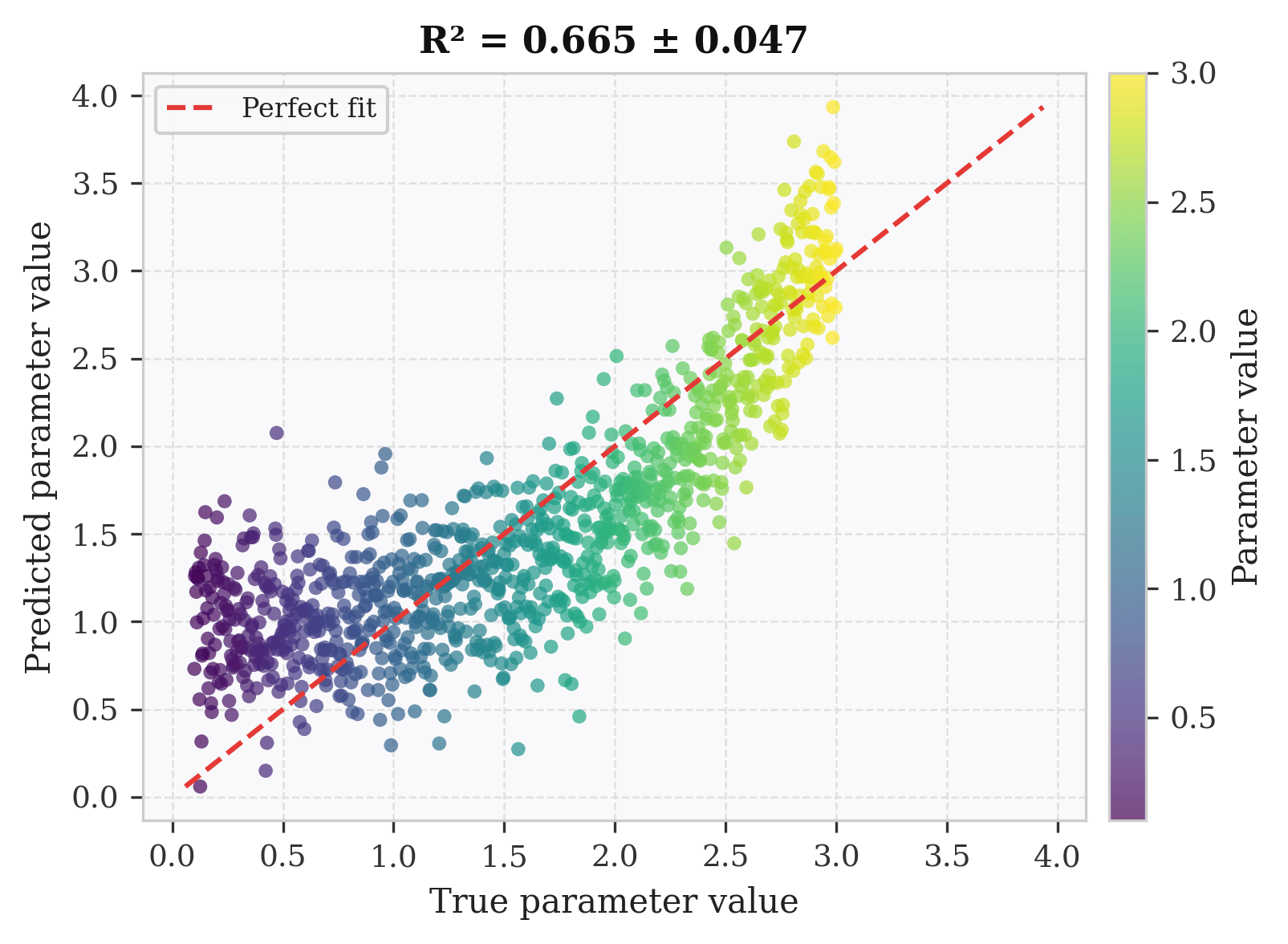}
        \caption{}
    \end{subfigure}%
    \hfill
    \begin{subfigure}[t]{0.19\linewidth}
        \centering
        \includegraphics[width=\linewidth]{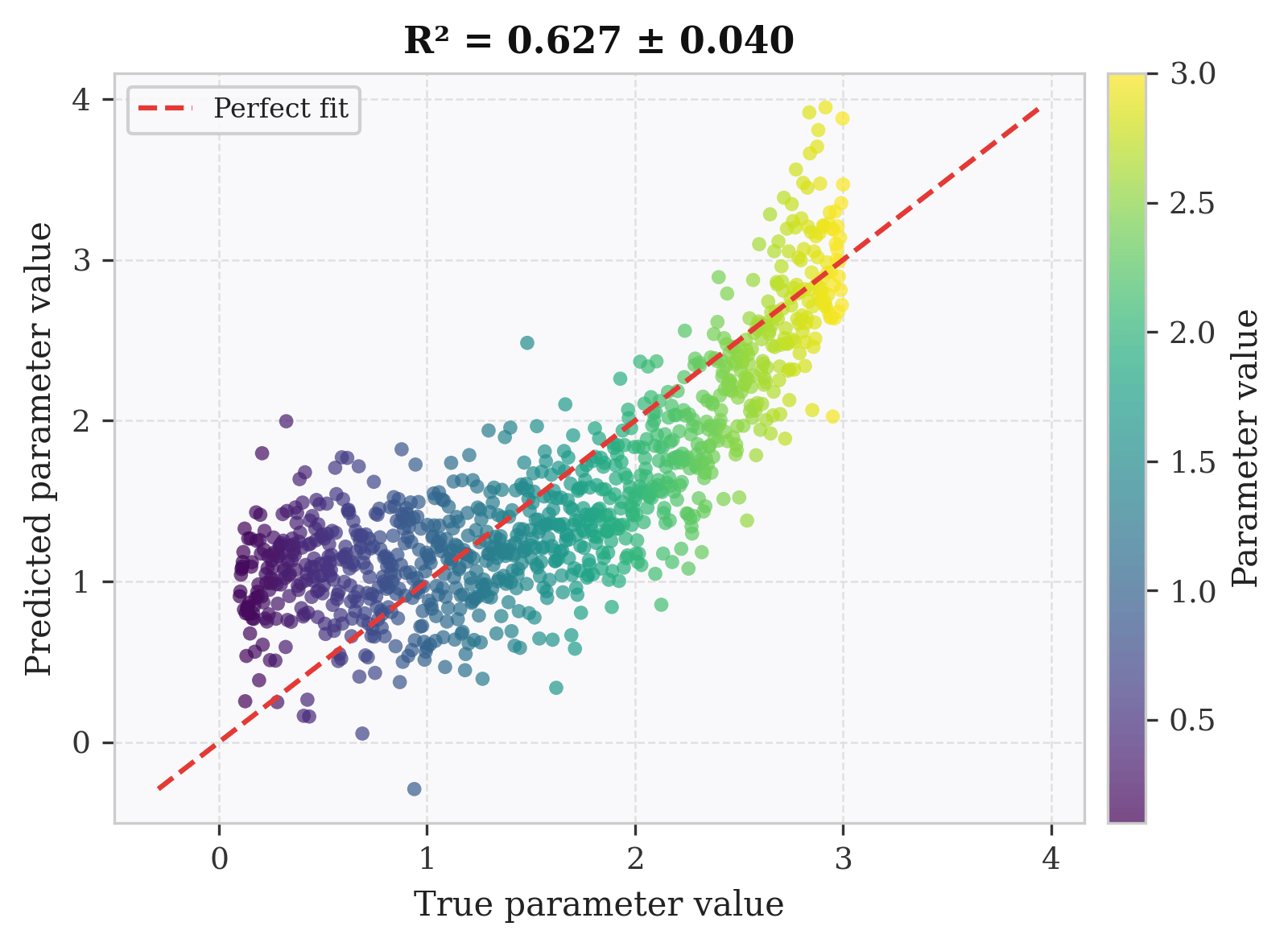}
        \caption{}
    \end{subfigure}%
    \hfill
    \begin{subfigure}[t]{0.19\linewidth}
        \centering
        \includegraphics[width=\linewidth]{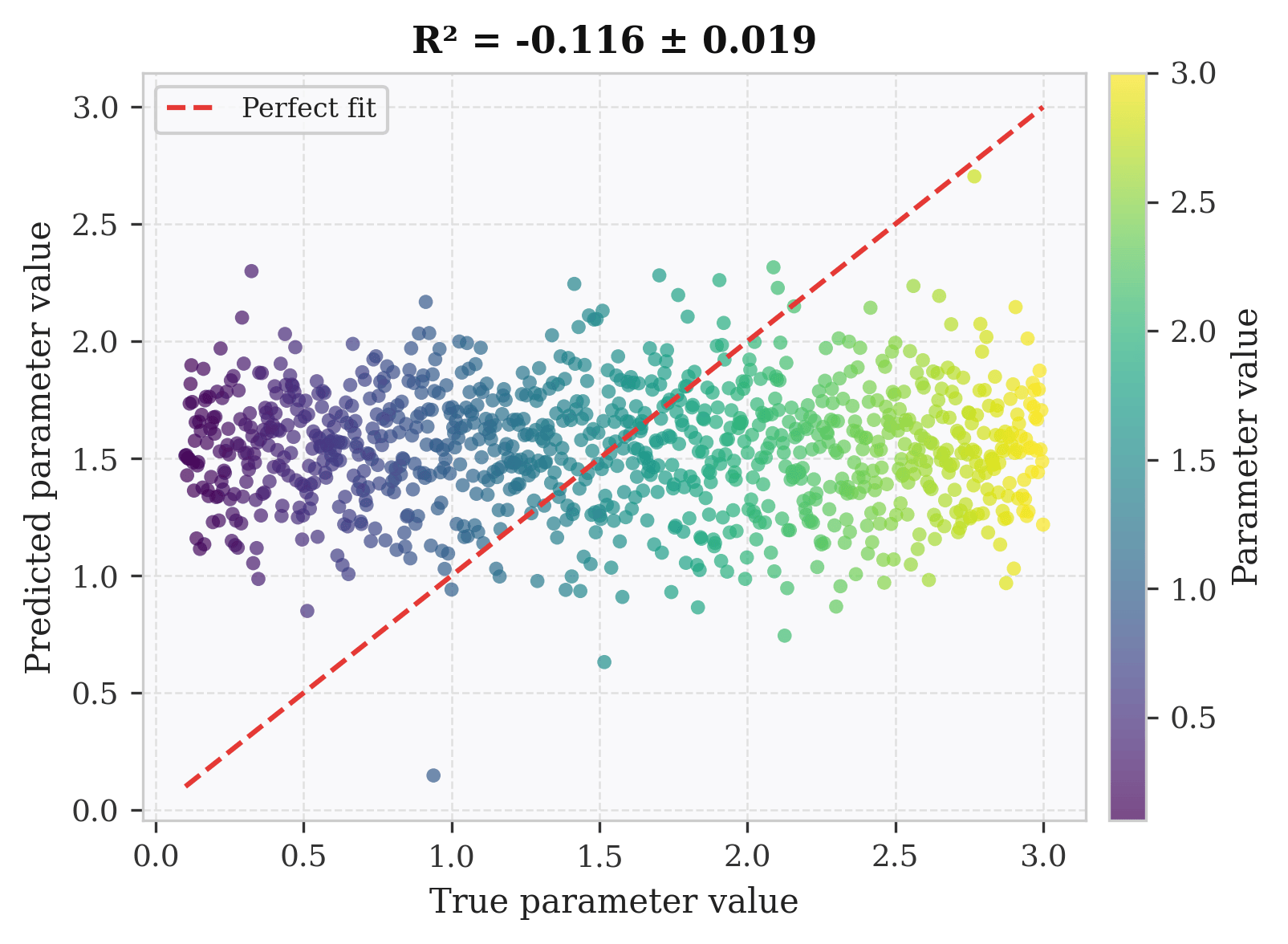}
        \caption{}
    \end{subfigure}
    \hfill
    \begin{subfigure}[t]{0.19\linewidth}
        \centering
        \includegraphics[width=\linewidth]{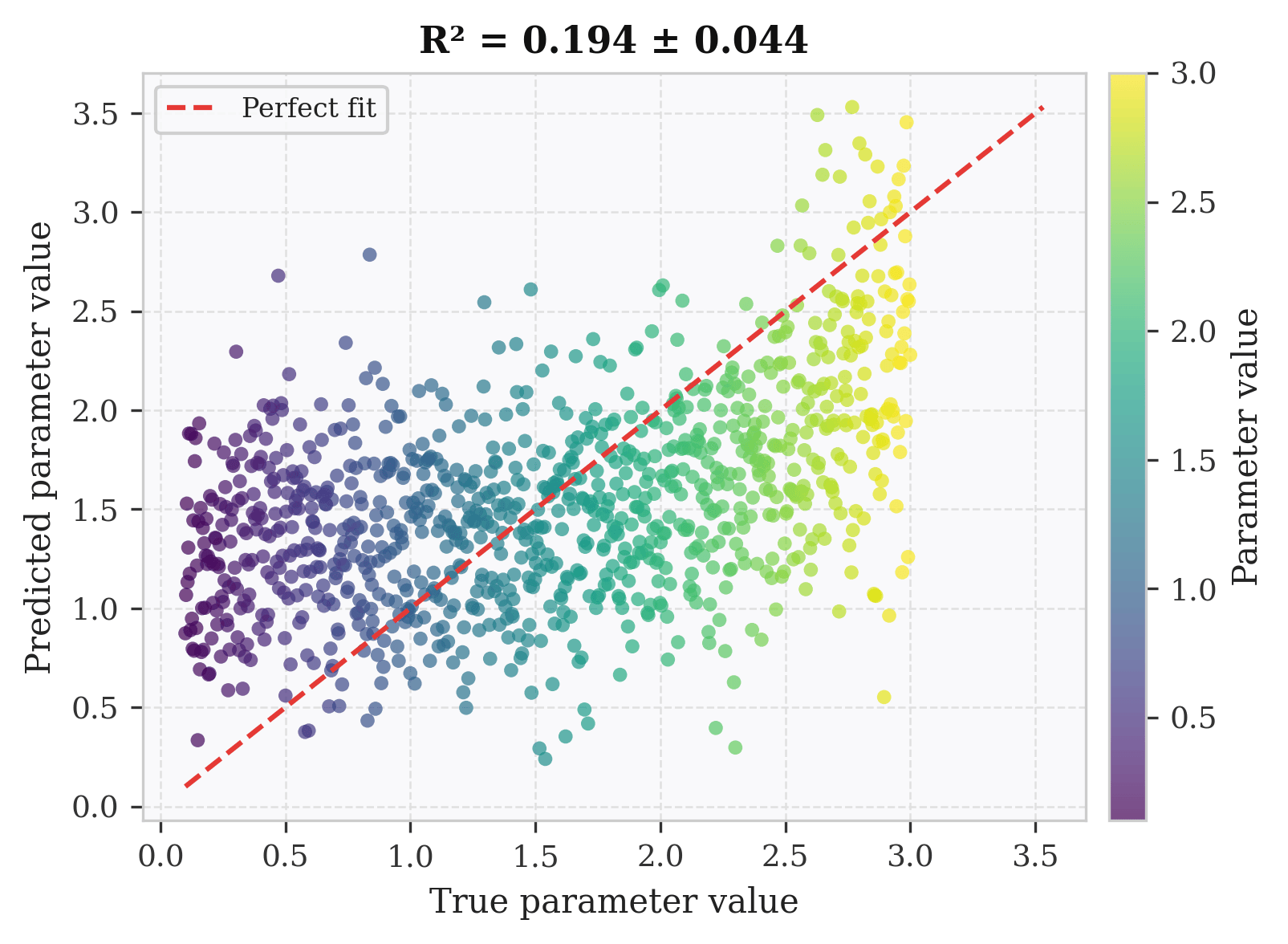}
        \caption{}
    \end{subfigure}

    \caption{
        Linear decodability comparison of Pz channel across oscillatory frequencies ($f_{\text{osc}}$) with varying power of the oscillation ($A_{\text{osc}}$) for CSBrain and LaBraM model.
    }
    \label{fig:pz_oscfreq_power}
\end{figure}

\begin{figure}[H]
    \centering
    
    % --- Column Titles ---
    \makebox[0.3\linewidth]{$\bm{\beta}$} \hfill
    \makebox[0.3\linewidth]{$\bm{A_{\text{ap}}}$} \hfill
    \makebox[0.3\linewidth]{$\bm{f_{\text{osc}}}$} \\ \vspace{0.2em}

    % --- Row 1 (CBraMod) ---
    \makebox[0pt][r]{\raisebox{1.4cm}[0pt][0pt]{\rotatebox[origin=c]{90}{\textbf{CBraMod}}}\hspace{1em}}%
    \begin{subfigure}[t]{0.3\linewidth}
        \centering
        \includegraphics[width=\linewidth]{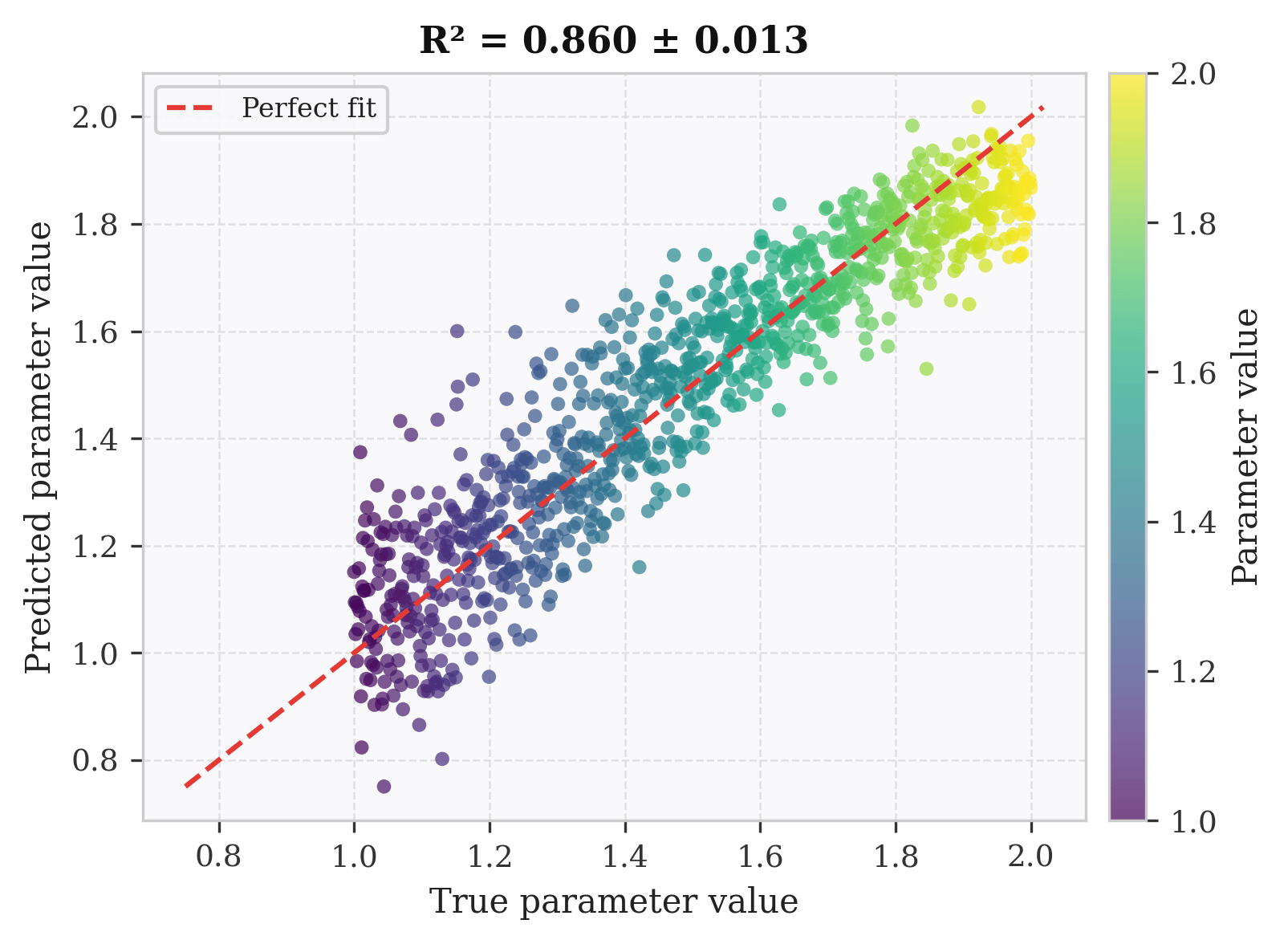}
        \caption{}
    \end{subfigure}
    \hfill
    \begin{subfigure}[t]{0.3\linewidth}
        \centering
        \includegraphics[width=\linewidth]{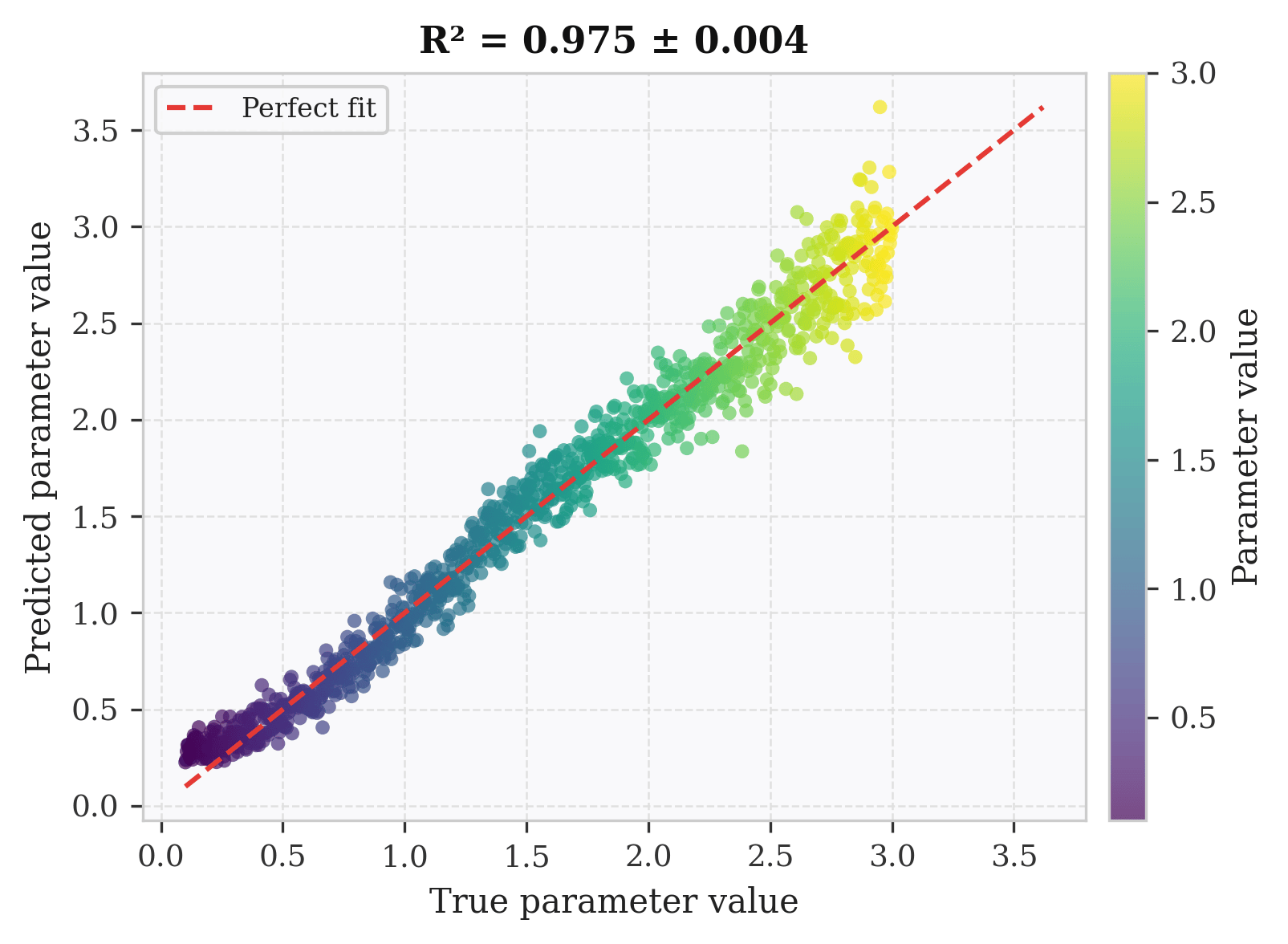}
        \caption{}
    \end{subfigure}
    \hfill
    \begin{subfigure}[t]{0.3\linewidth}
        \centering
        \includegraphics[width=\linewidth]{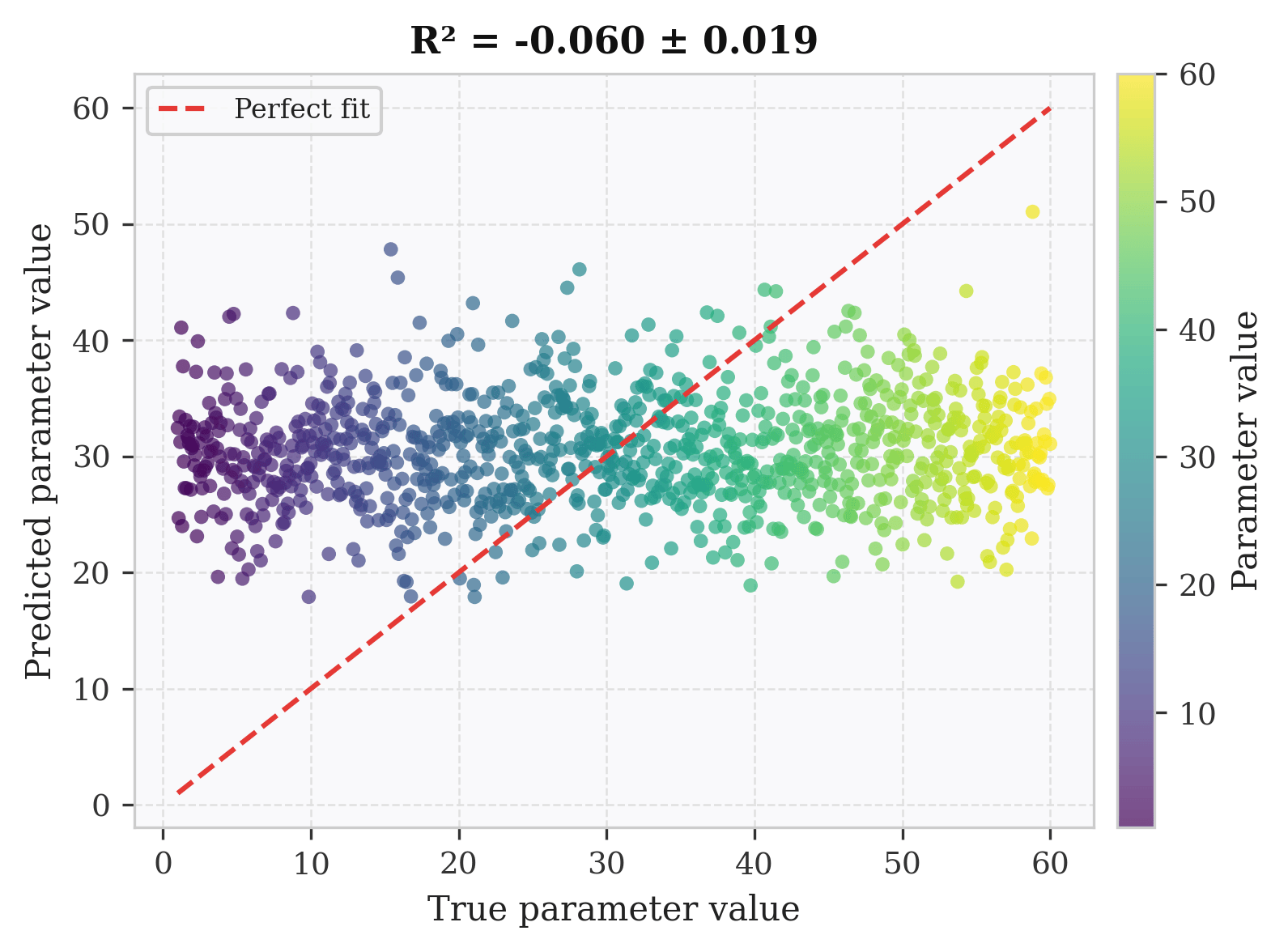}
        \caption{}
    \end{subfigure}

    \vspace{-0.5cm}

    % --- Row 2 (CSBrain) ---
    \makebox[0pt][r]{\raisebox{1.4cm}[0pt][0pt]{\rotatebox[origin=c]{90}{\textbf{CSBrain}}}\hspace{1em}}%
    \begin{subfigure}[t]{0.3\linewidth}
        \centering
        \includegraphics[width=\linewidth]{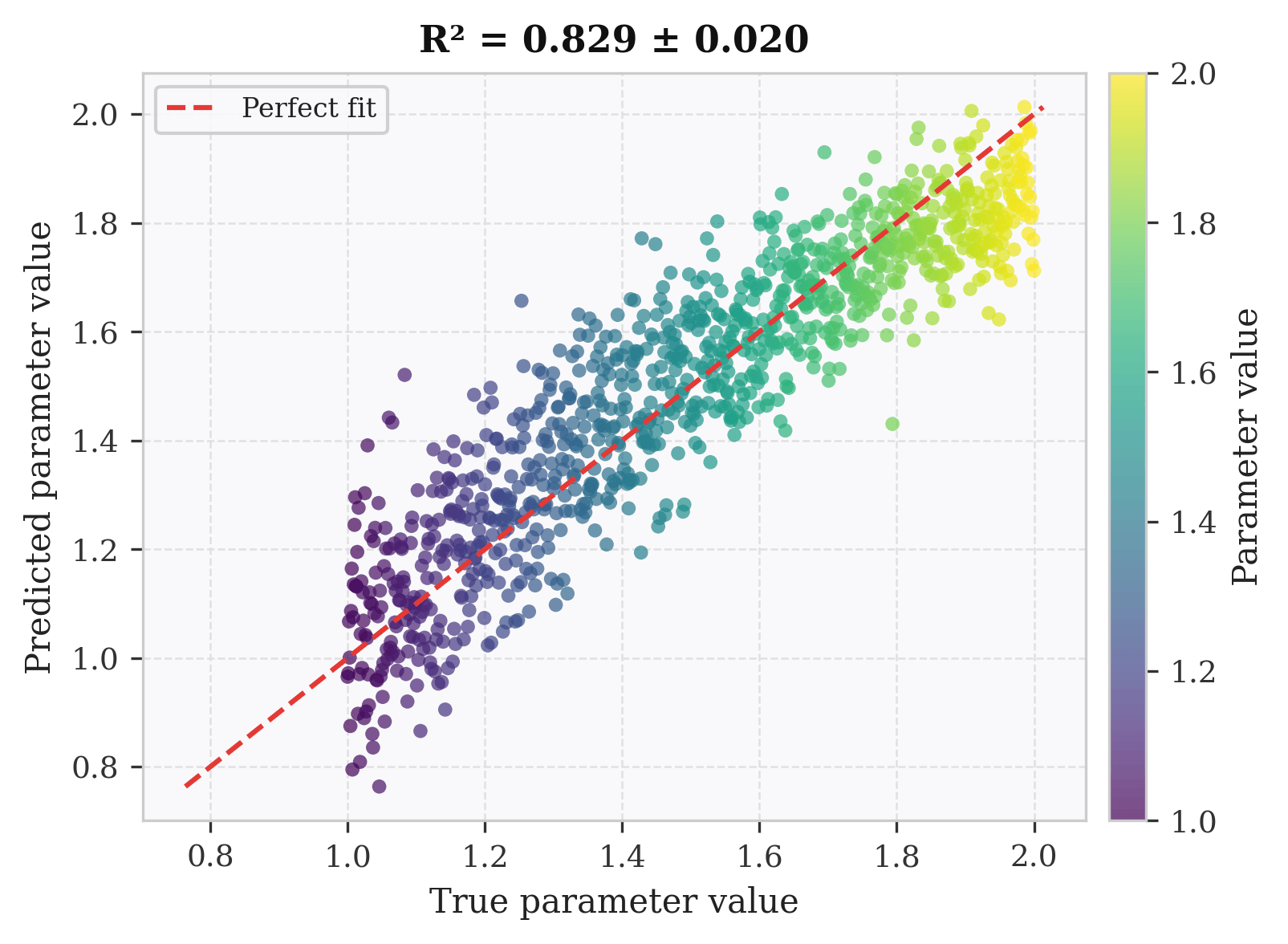}
        \caption{}
    \end{subfigure}
    \hfill
    \begin{subfigure}[t]{0.3\linewidth}
        \centering
        \includegraphics[width=\linewidth]{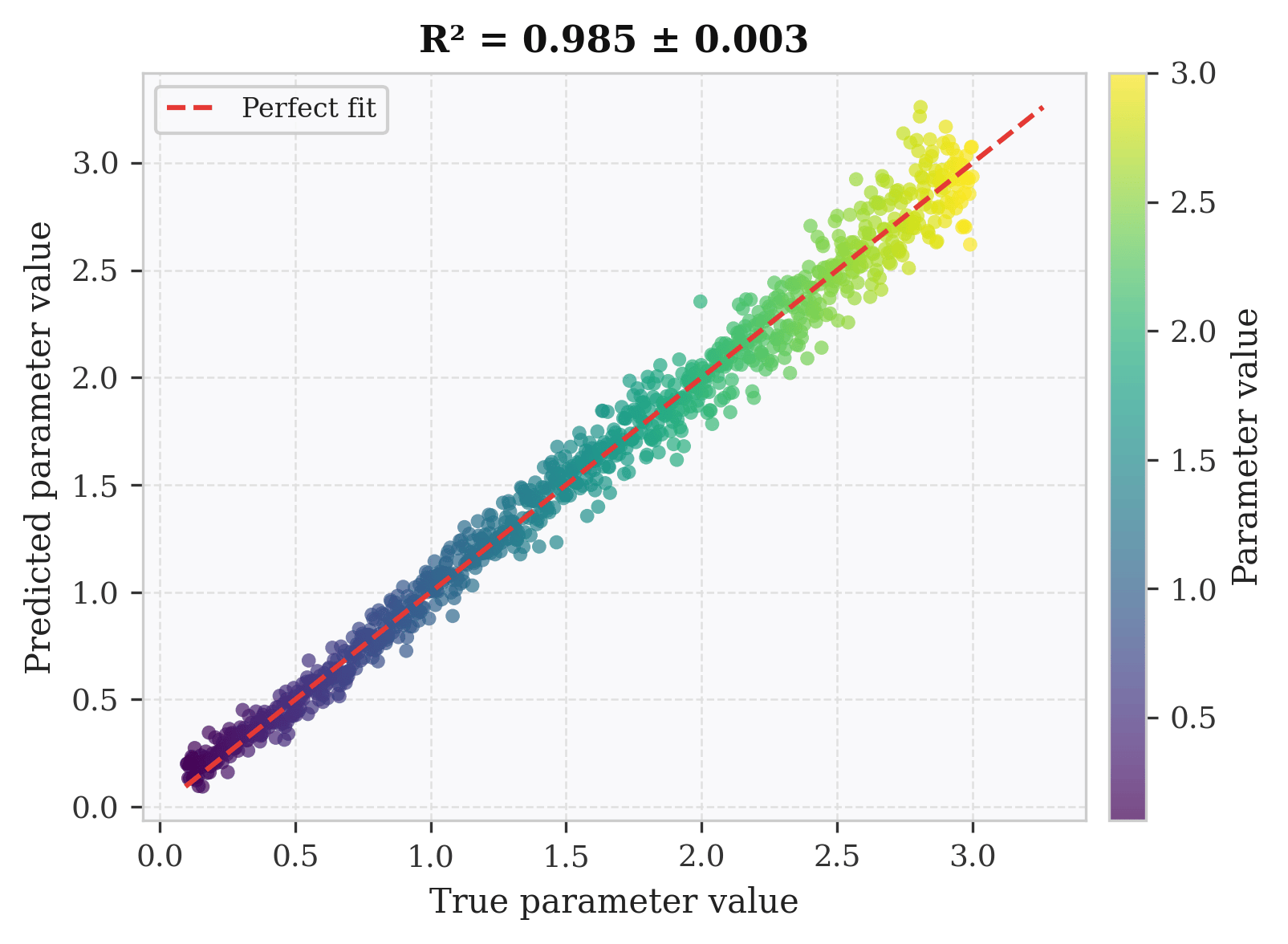}
        \caption{}
    \end{subfigure}
    \hfill
    \begin{subfigure}[t]{0.3\linewidth}
        \centering
        \includegraphics[width=\linewidth]{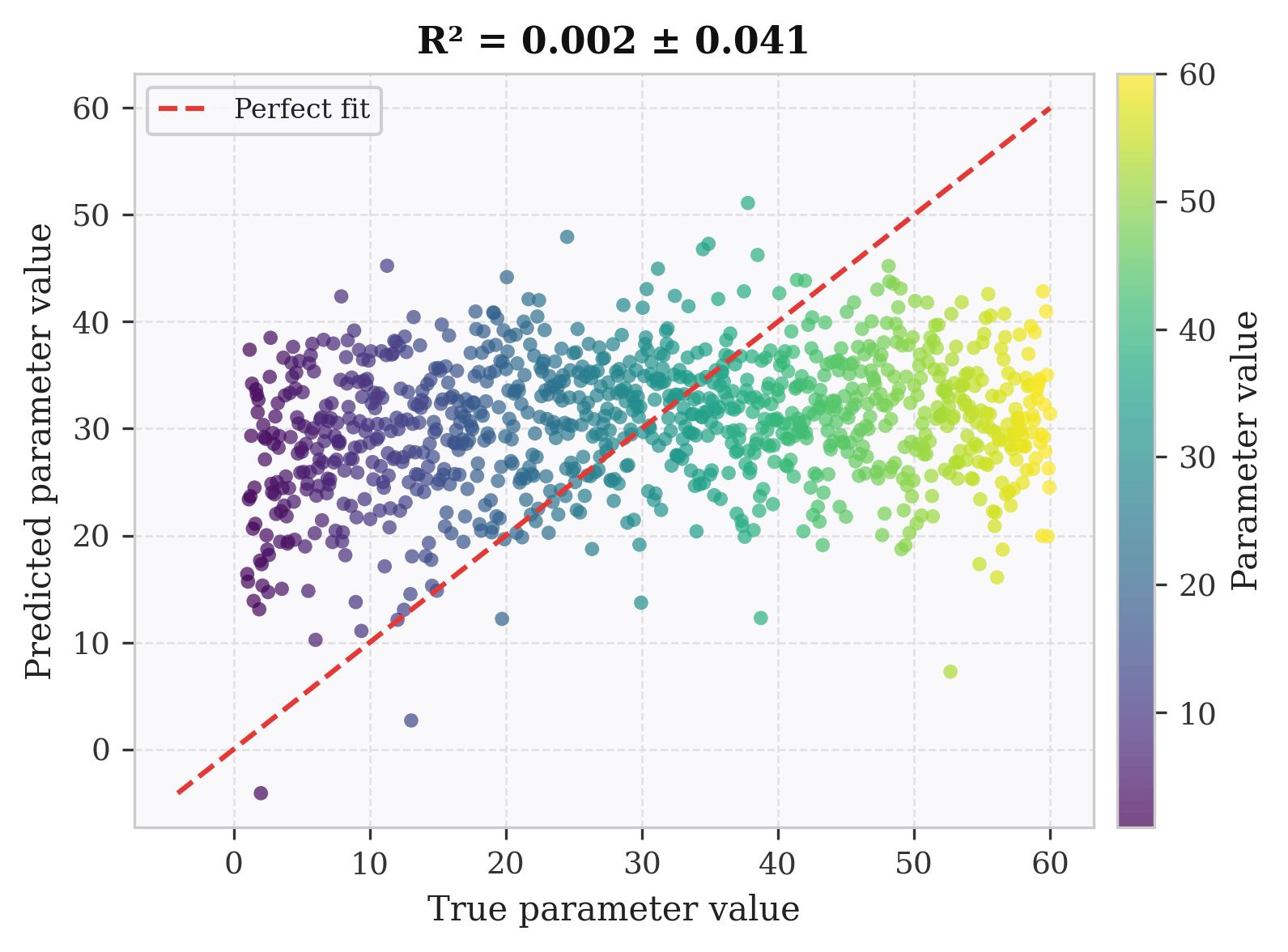}
        \caption{}
    \end{subfigure}

    \vspace{-0.5cm}

    % --- Row 3 (LaBraM) ---
    \makebox[0pt][r]{\raisebox{1.2cm}[0pt][0pt]{\rotatebox[origin=c]{90}{\textbf{LaBraM}}}\hspace{1em}}%
    \begin{subfigure}[t]{0.3\linewidth}
        \centering
        \includegraphics[width=\linewidth]{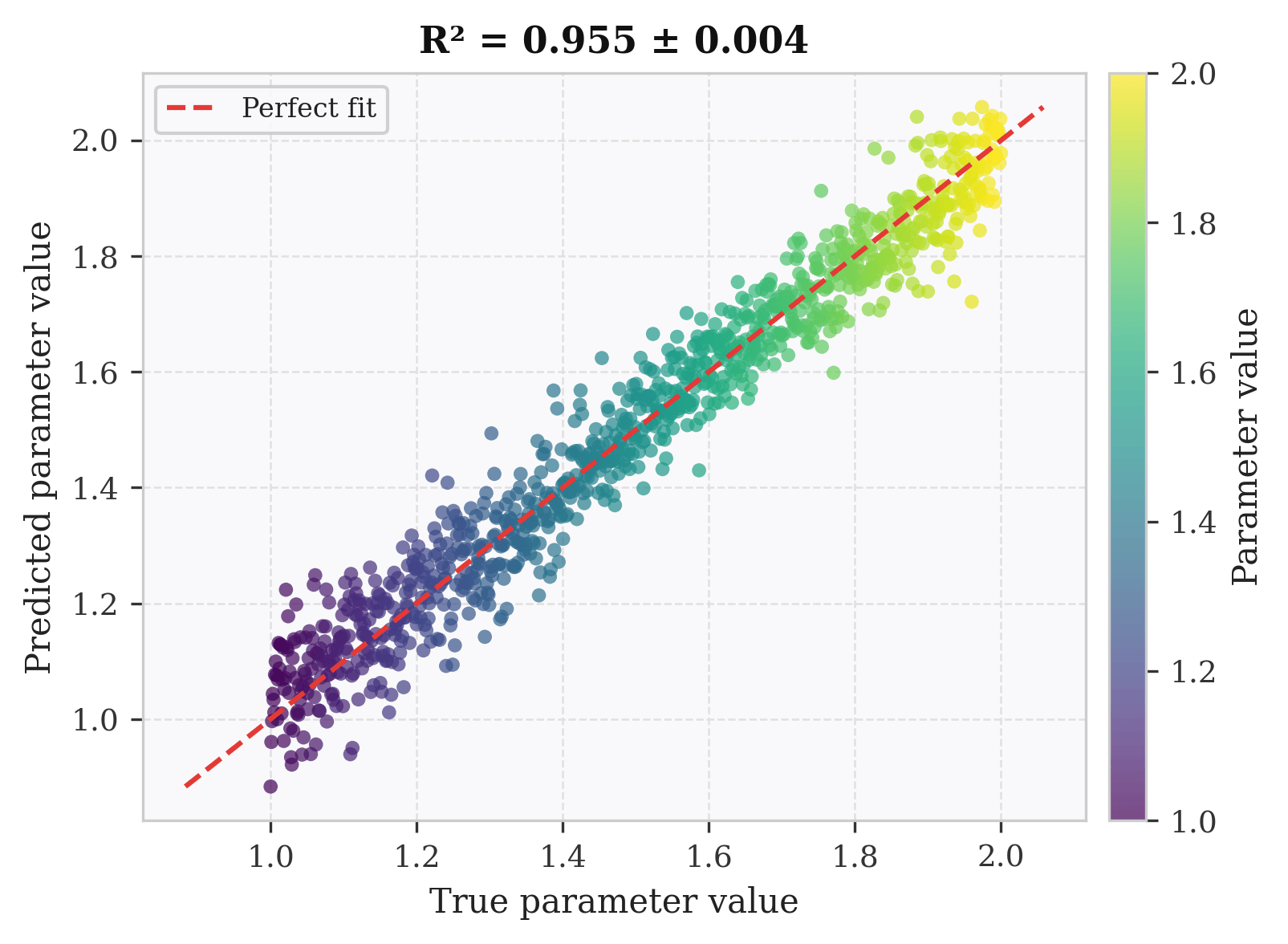}
        \caption{}
    \end{subfigure}
    \hfill
    \begin{subfigure}[t]{0.3\linewidth}
        \centering
        \includegraphics[width=\linewidth]{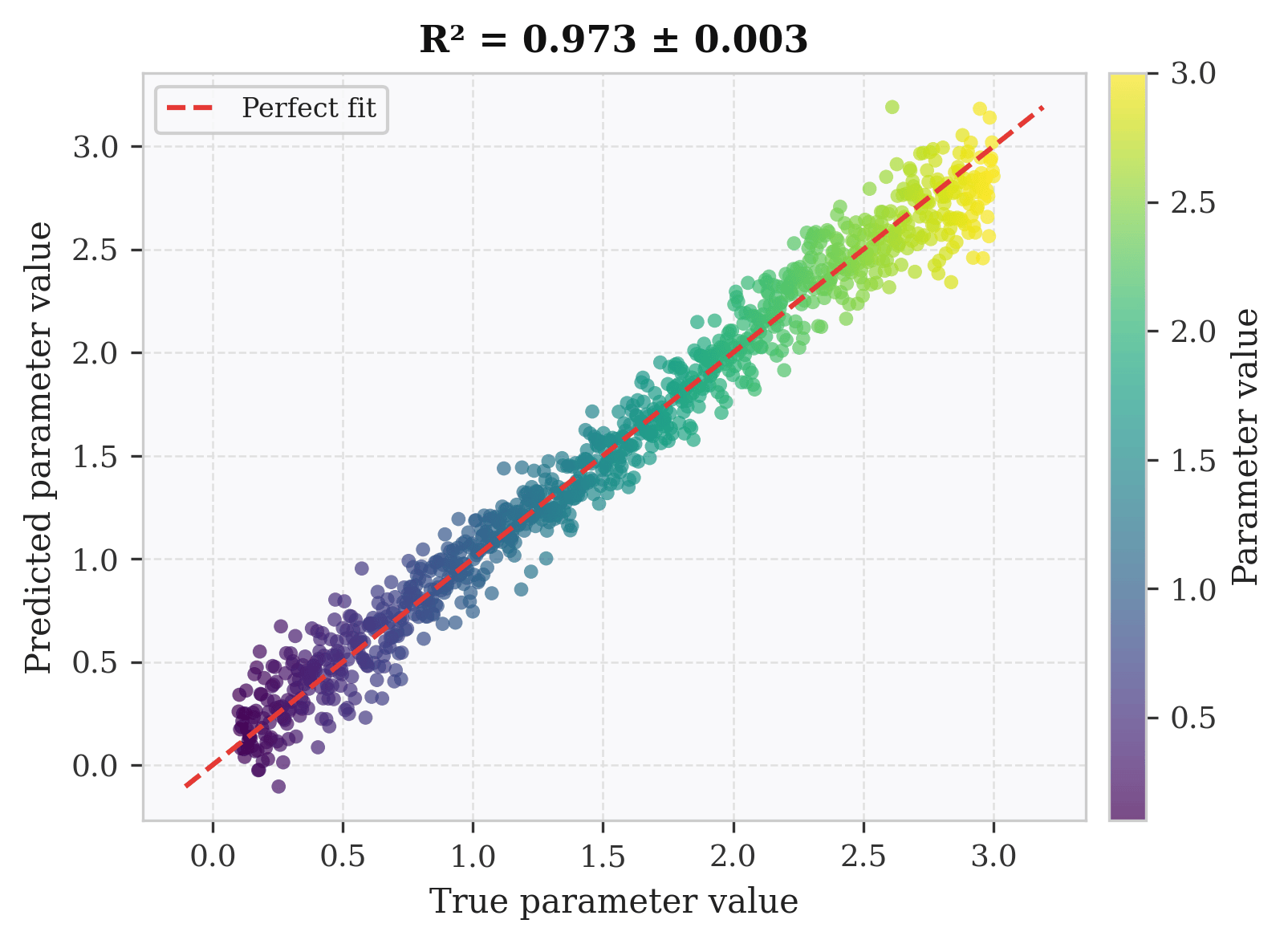}
        \caption{}
    \end{subfigure}
    \hfill
    \begin{subfigure}[t]{0.3\linewidth}
        \centering
        \includegraphics[width=\linewidth]{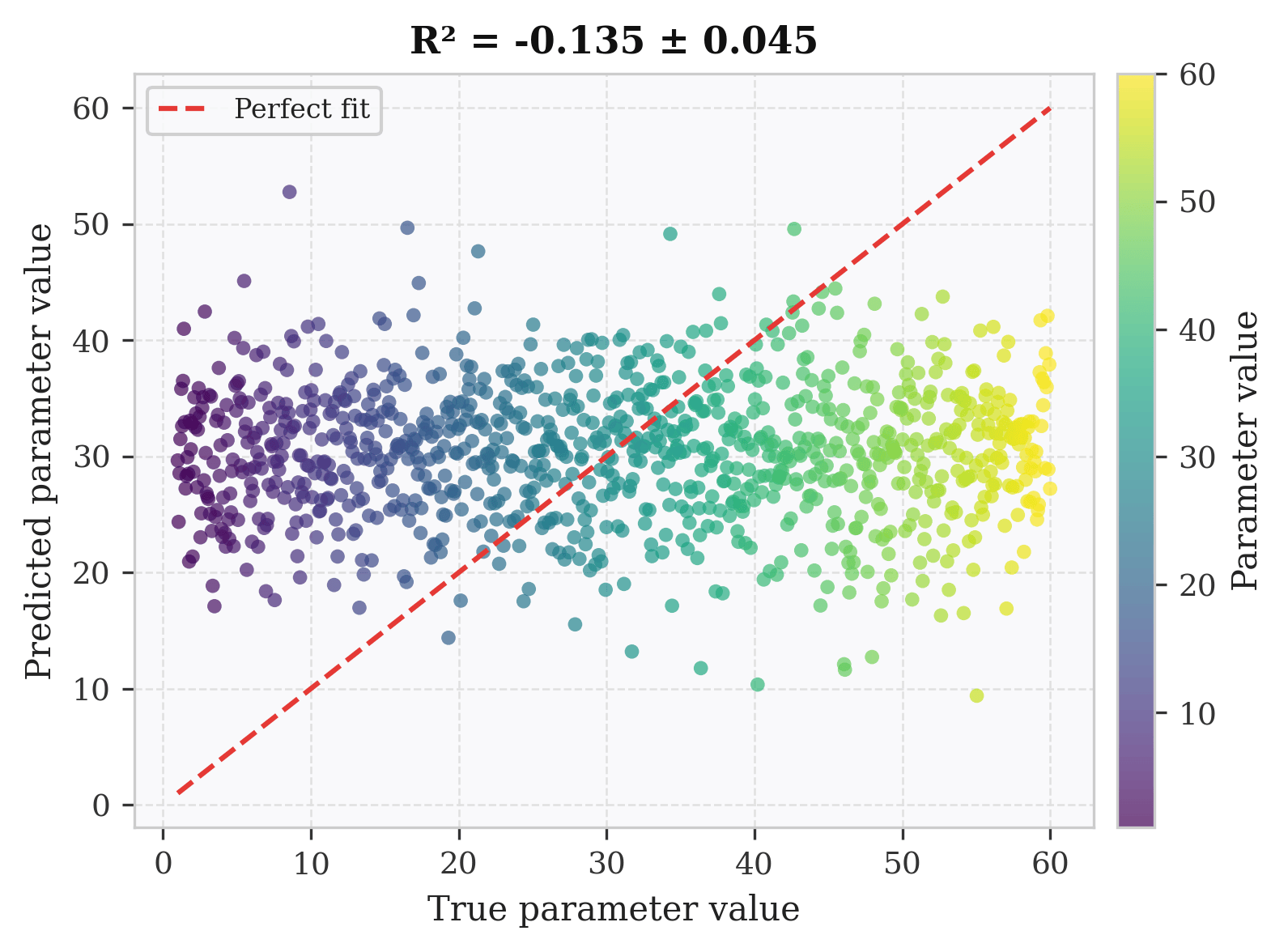}
        \caption{}
    \end{subfigure}

    \caption{
    Linear decodability for Oz channel across three foundation models (CBraMod, CSBrain, LaBraM) for Aperiodic Exponent ($\beta$), Aperiodic Offset ($A_{\text{ap}}$) and Oscillation frequency ($f_{\text{osc}}$).
    }
    \label{fig:Oz}
\end{figure}

\begin{figure}[H]
    \centering
    
    % --- Column Titles ---
    \makebox[0.19\linewidth]{\textbf{10Hz}} \hfill
    \makebox[0.19\linewidth]{\textbf{20Hz}} \hfill
    \makebox[0.19\linewidth]{\textbf{30Hz}} \hfill
    \makebox[0.19\linewidth]{\textbf{40Hz}} \hfill
    \makebox[0.19\linewidth]{\textbf{50Hz}}\\ \vspace{0.2em}

    \makebox[0pt][r]{\raisebox{1.2cm}[0pt][0pt]{\rotatebox[origin=c]{90}{\textbf{CSBrain}}}\hspace{1em}}%
    \begin{subfigure}[t]{0.19\linewidth}
        \centering
        \includegraphics[width=\linewidth]{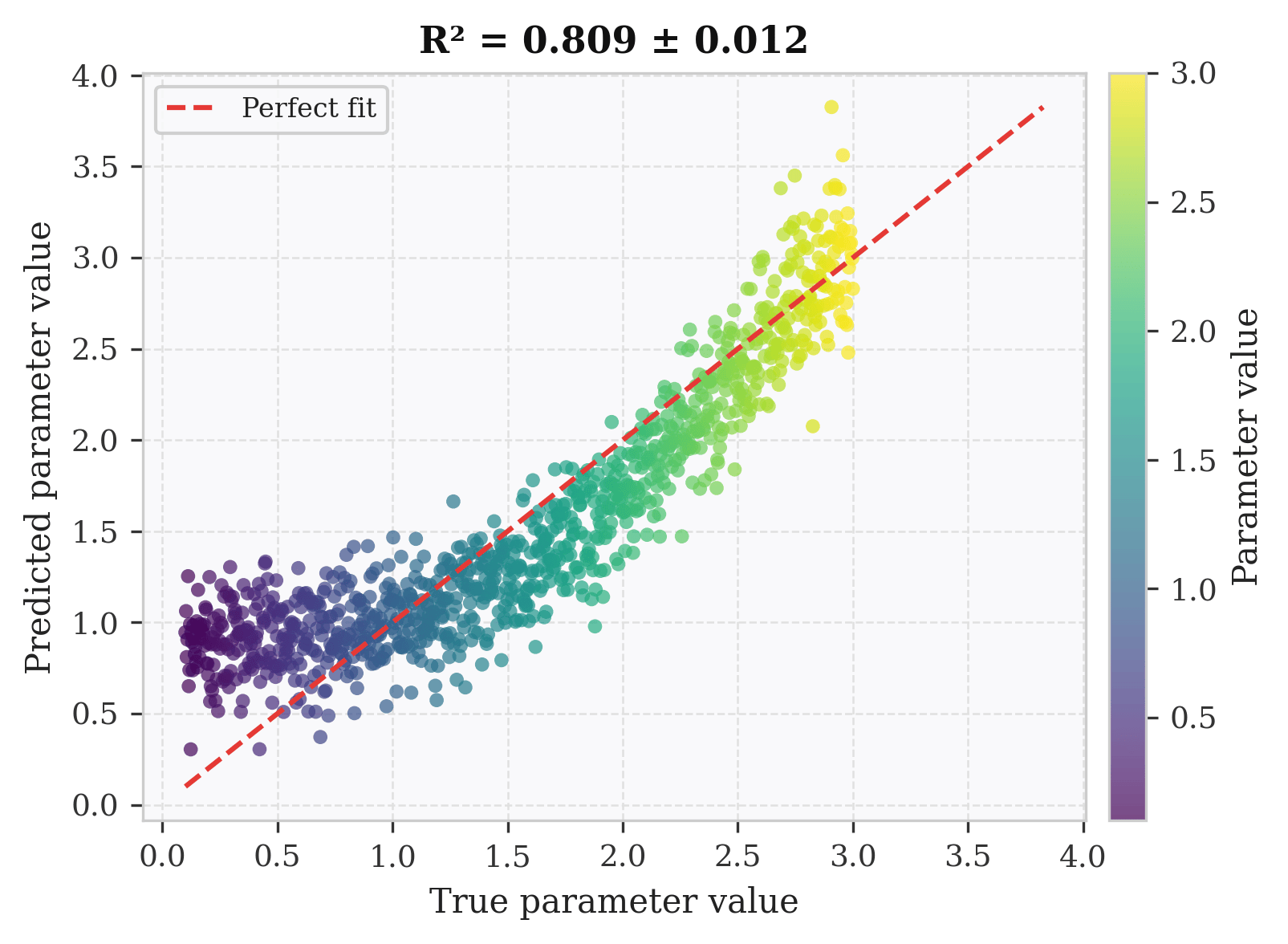}
        \caption{}
    \end{subfigure}%
    \hfill
    \begin{subfigure}[t]{0.19\linewidth}
        \centering
        \includegraphics[width=\linewidth]{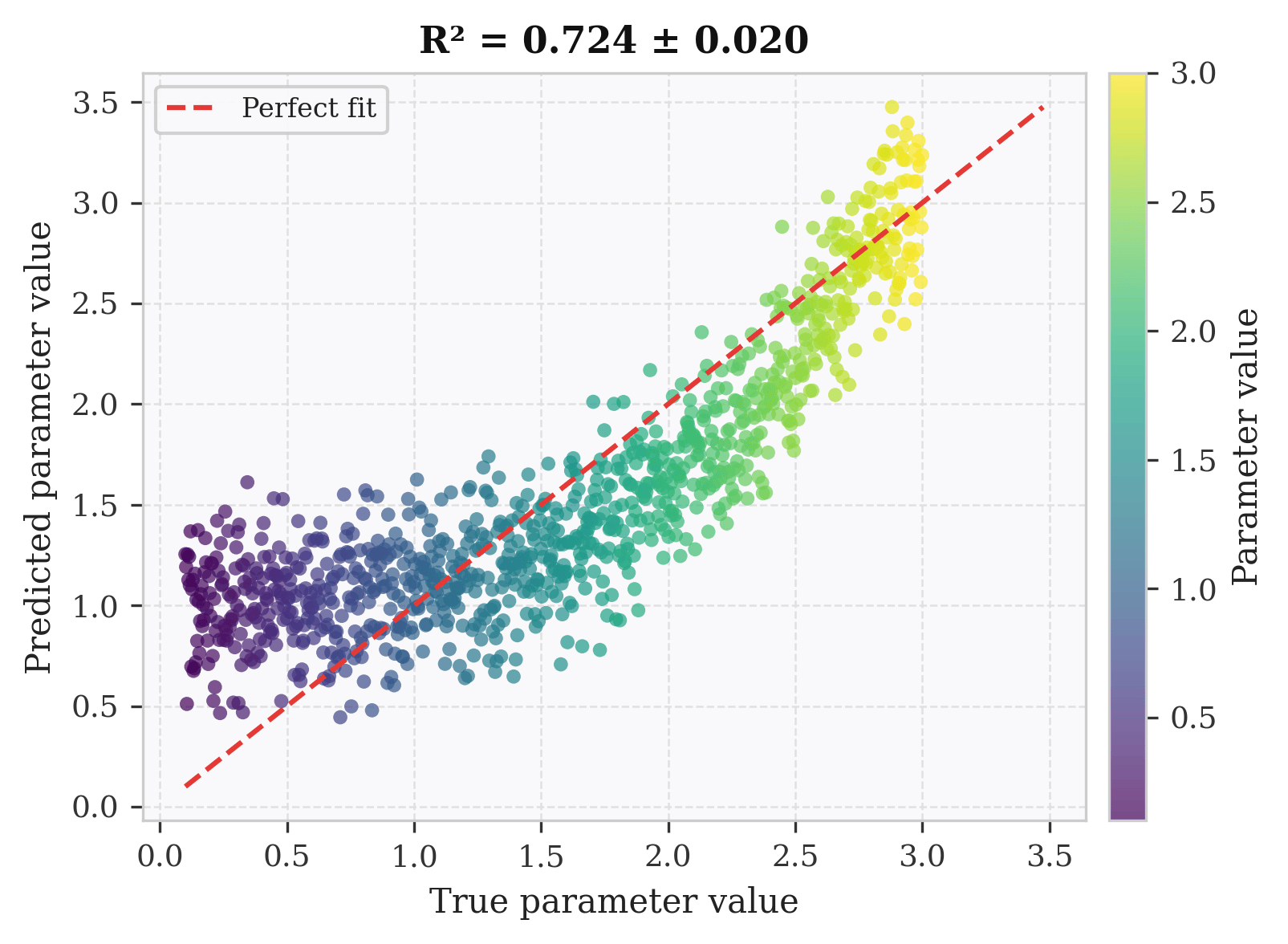}
        \caption{}
    \end{subfigure}%
    \hfill
    \begin{subfigure}[t]{0.19\linewidth}
        \centering
        \includegraphics[width=\linewidth]{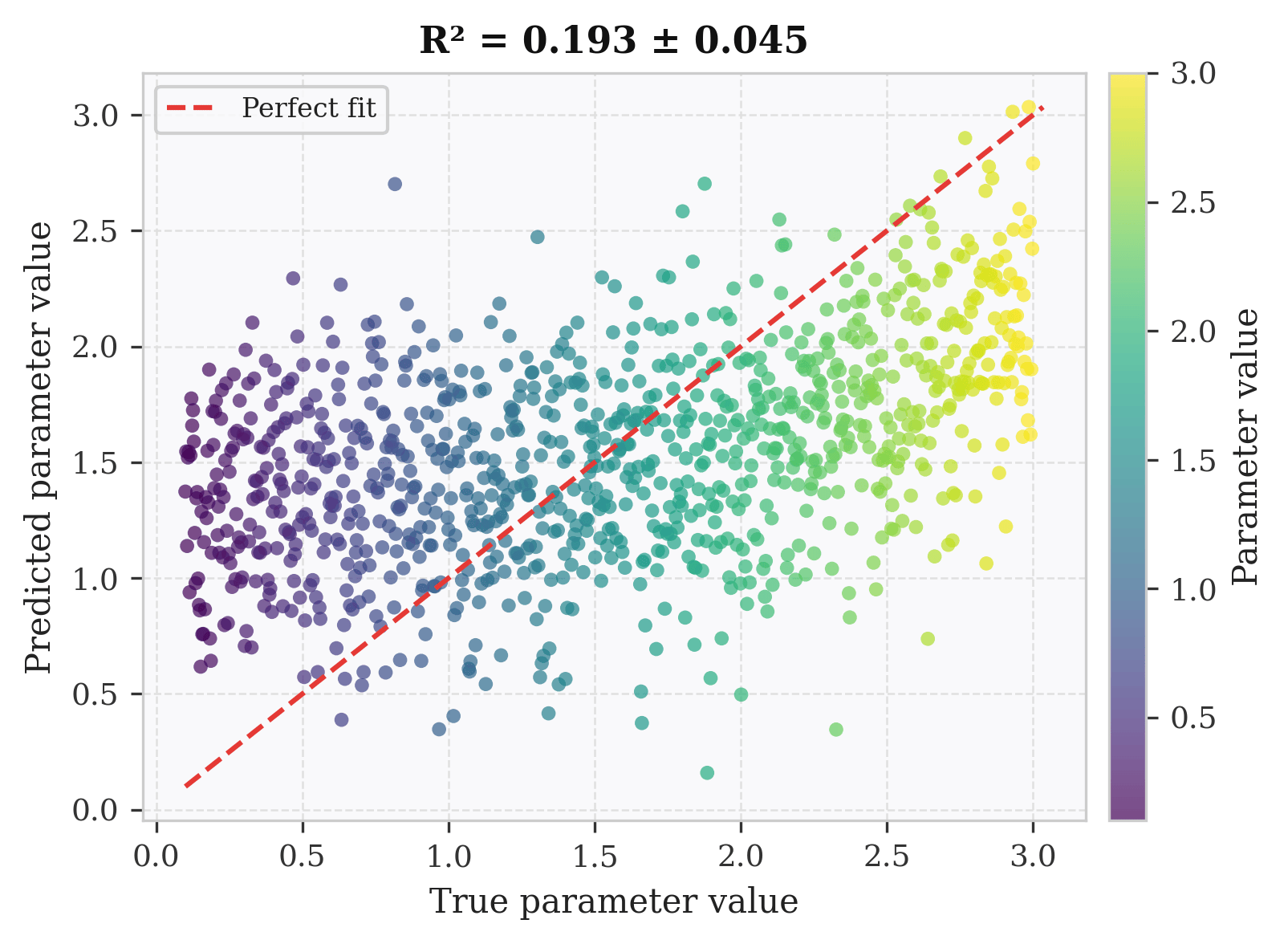}
        \caption{}
    \end{subfigure}%
    \hfill
    \begin{subfigure}[t]{0.19\linewidth}
        \centering
        \includegraphics[width=\linewidth]{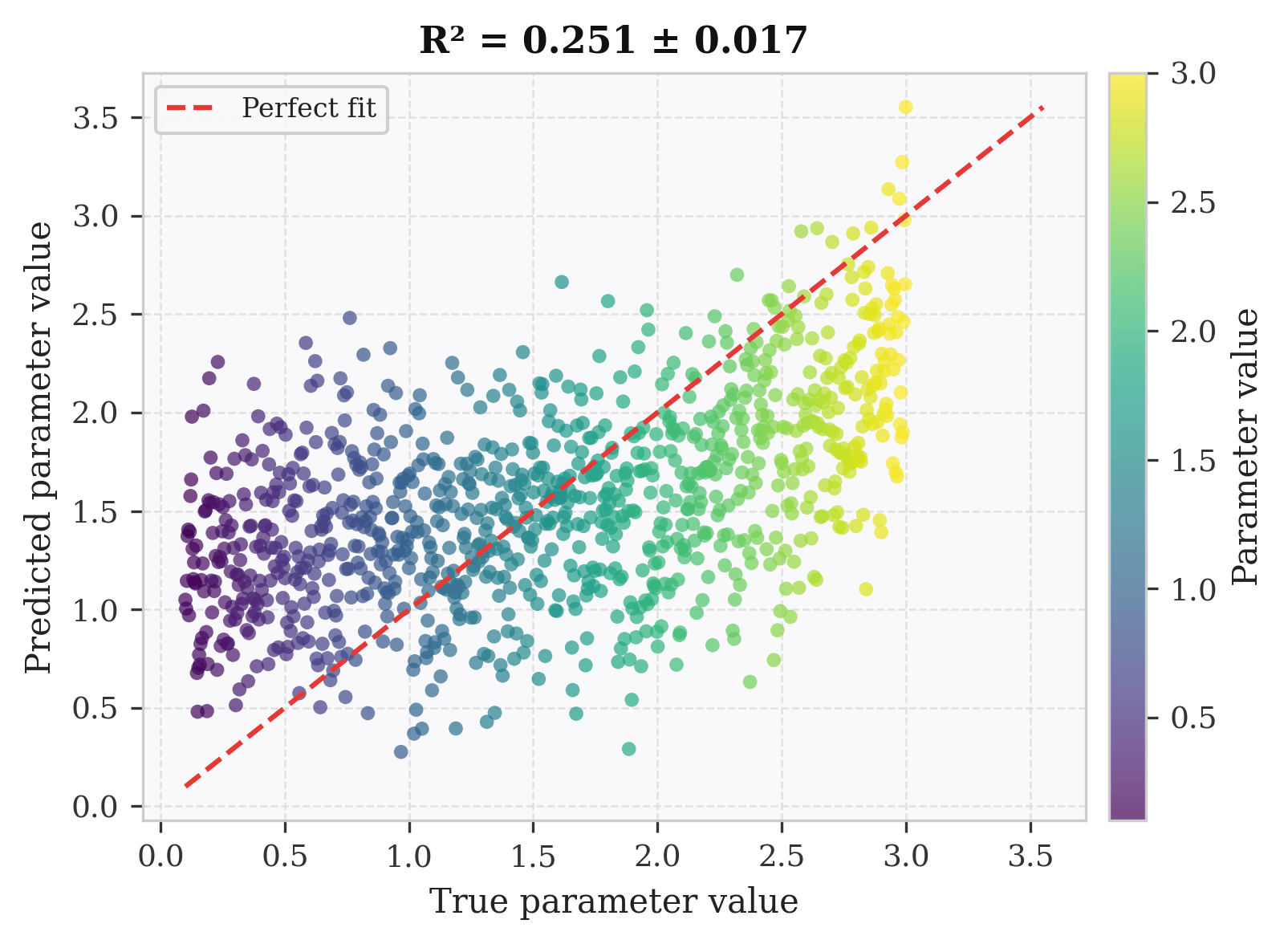}
        \caption{}
    \end{subfigure}
    \hfill
    \begin{subfigure}[t]{0.19\linewidth}
        \centering
        \includegraphics[width=\linewidth]{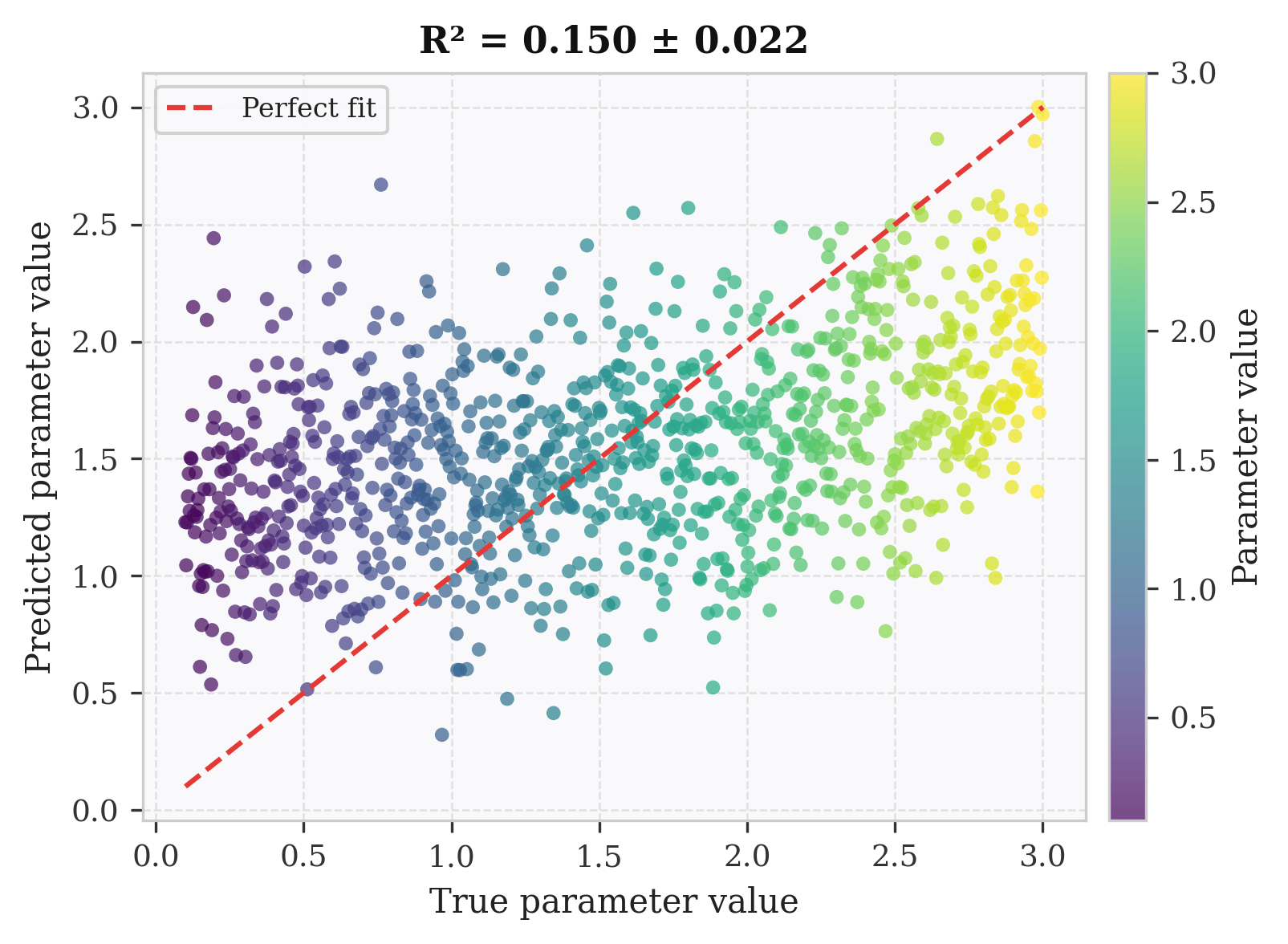}
        \caption{}
    \end{subfigure}

    \vspace{-0.2em}

    \makebox[0pt][r]{\raisebox{1.2cm}[0pt][0pt]{\rotatebox[origin=c]{90}{\textbf{LaBraM}}}\hspace{1em}}%
    \begin{subfigure}[t]{0.19\linewidth}
        \centering
        \includegraphics[width=\linewidth]{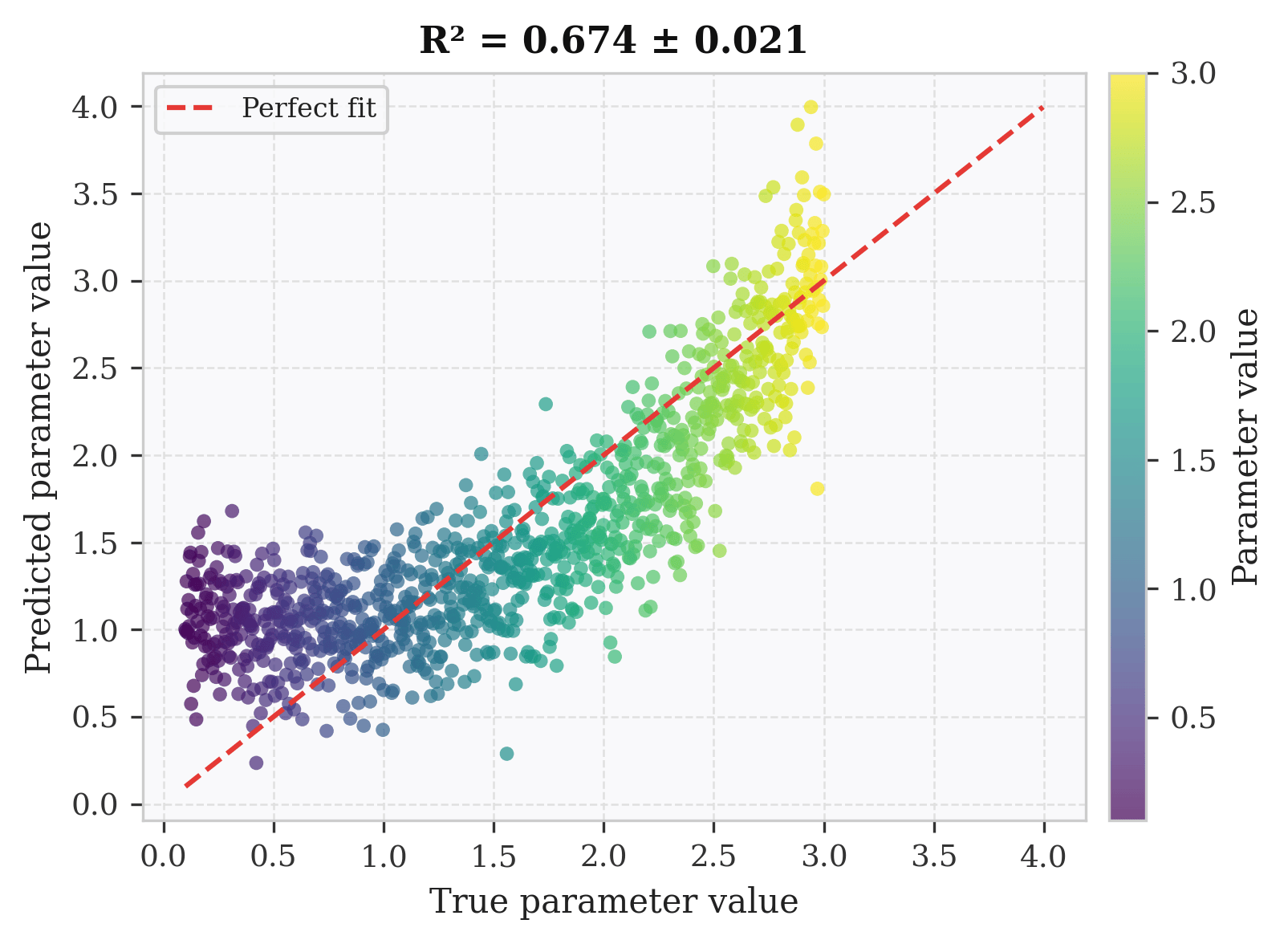}
        \caption{}
    \end{subfigure}%
    \hfill
    \begin{subfigure}[t]{0.19\linewidth}
        \centering
        \includegraphics[width=\linewidth]{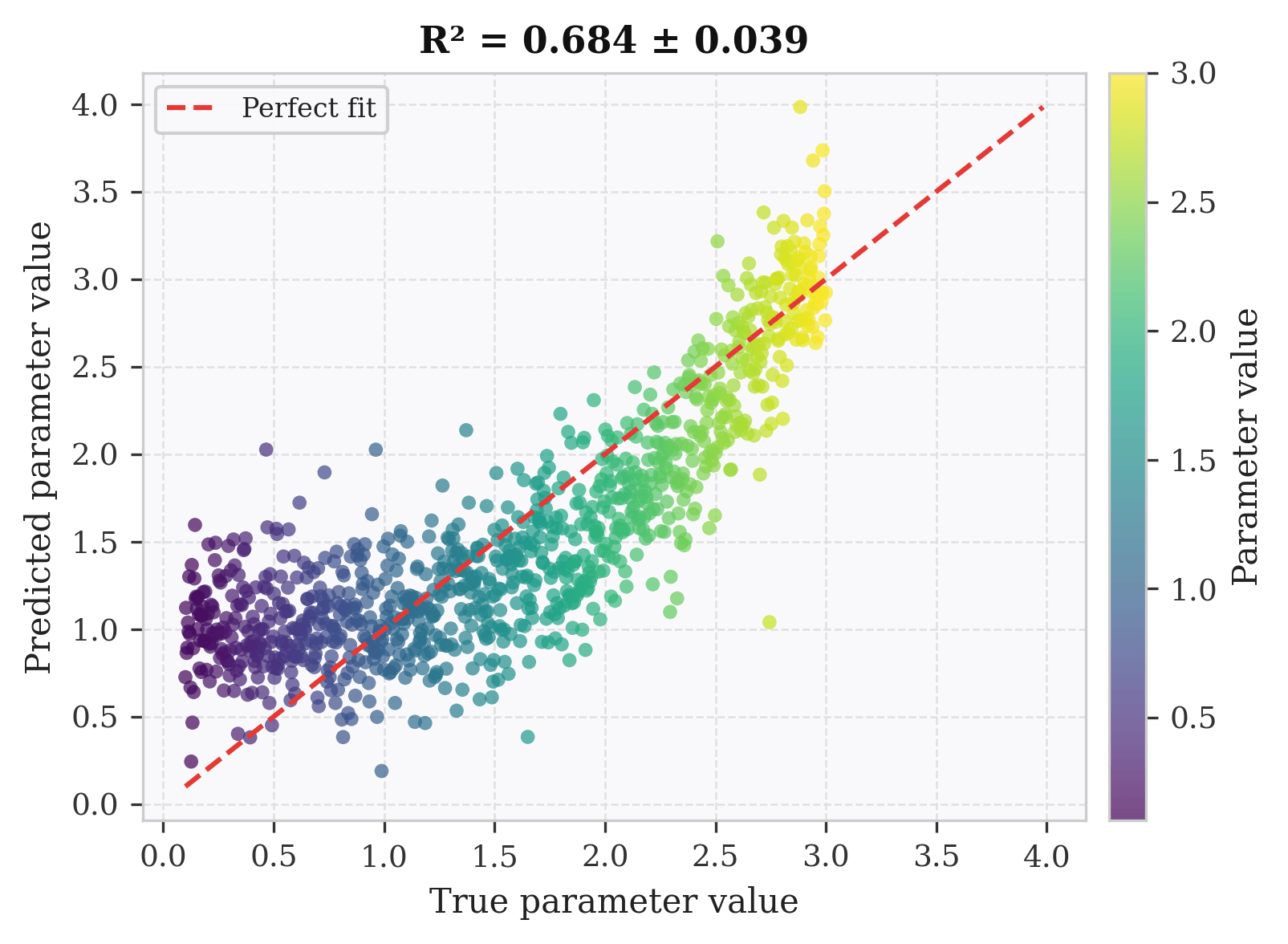}
        \caption{}
    \end{subfigure}%
    \hfill
    \begin{subfigure}[t]{0.19\linewidth}
        \centering
        \includegraphics[width=\linewidth]{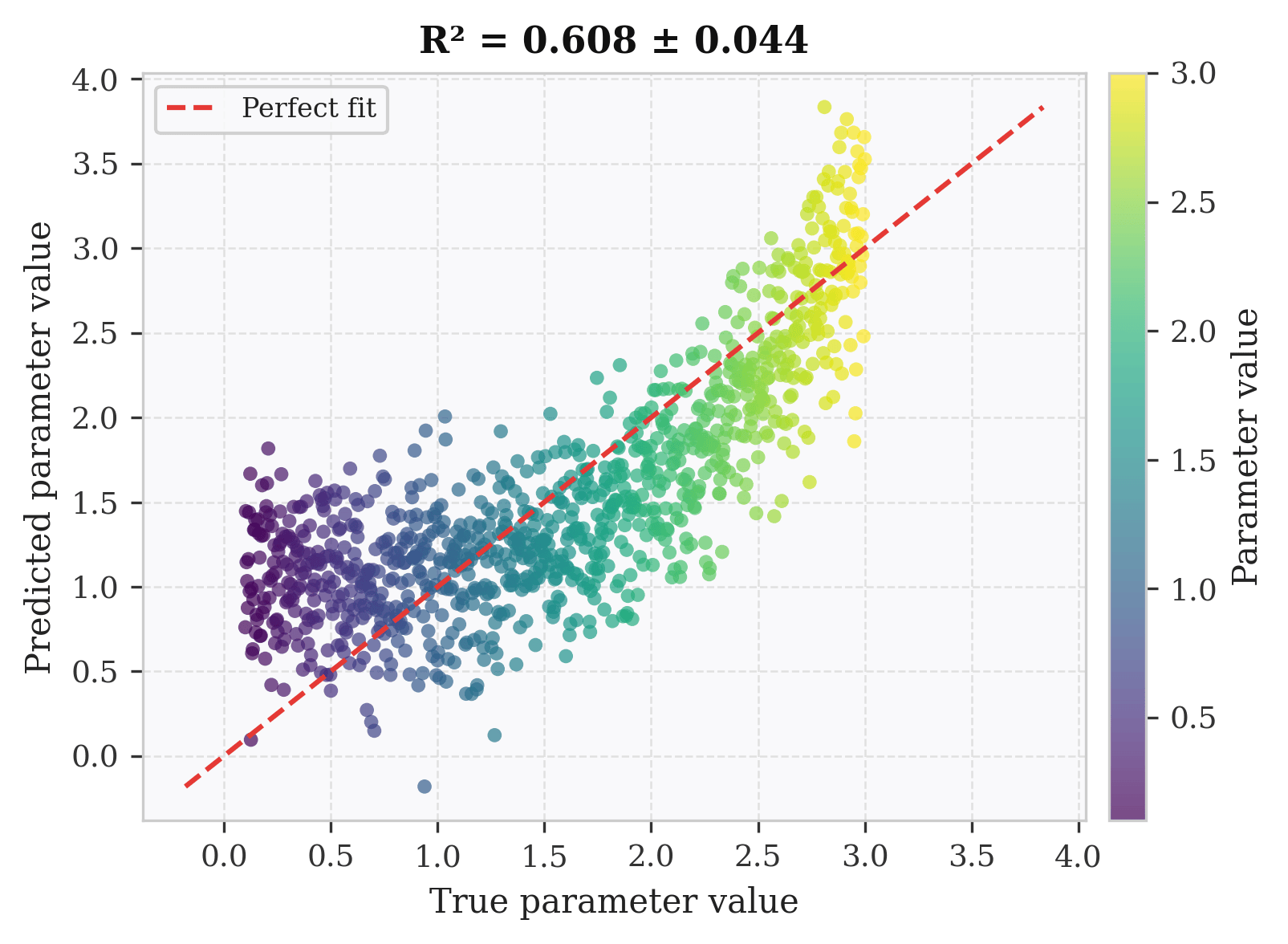}
        \caption{}
    \end{subfigure}%
    \hfill
    \begin{subfigure}[t]{0.19\linewidth}
        \centering
        \includegraphics[width=\linewidth]{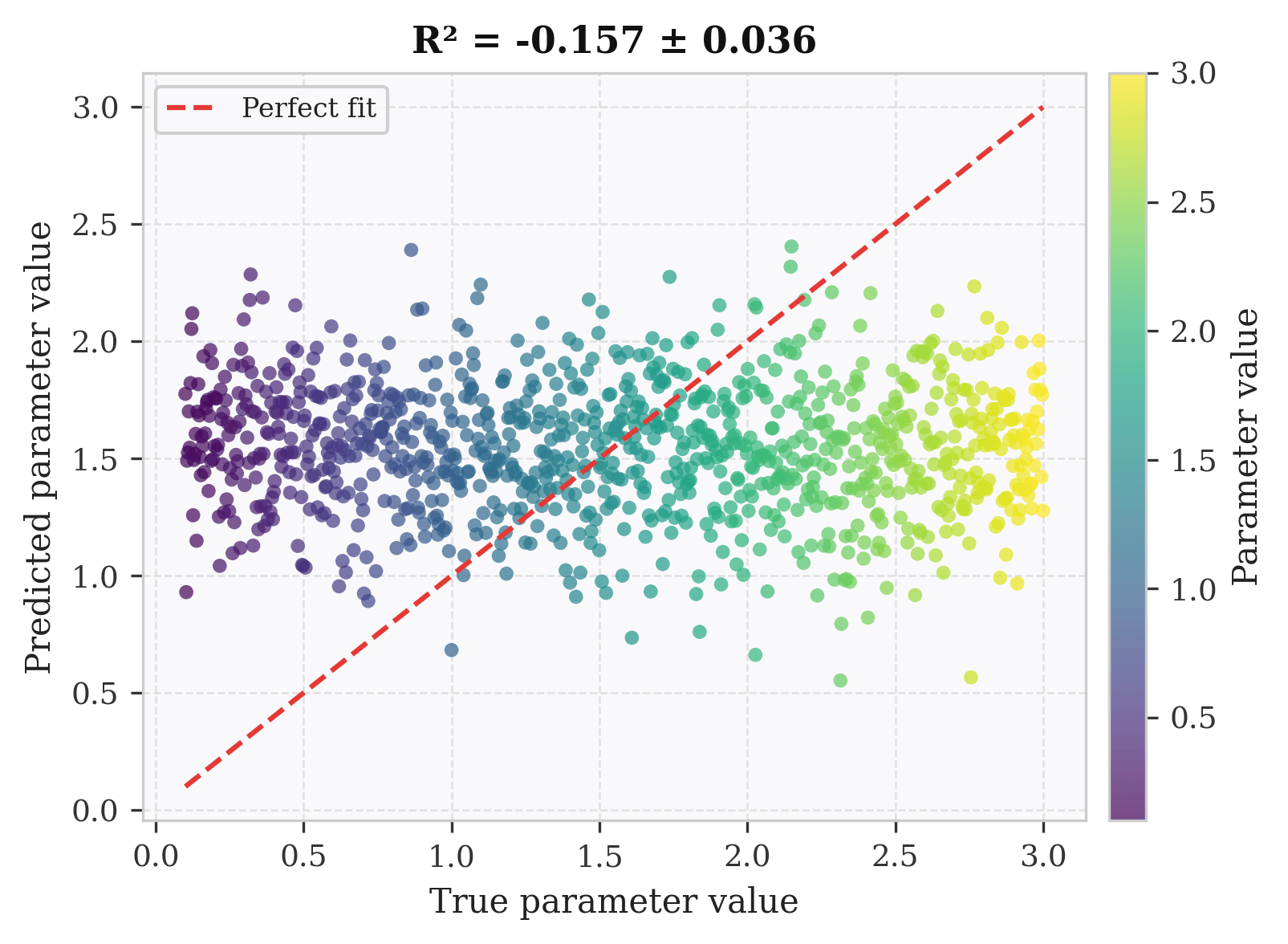}
        \caption{}
    \end{subfigure}
    \hfill
    \begin{subfigure}[t]{0.19\linewidth}
        \centering
        \includegraphics[width=\linewidth]{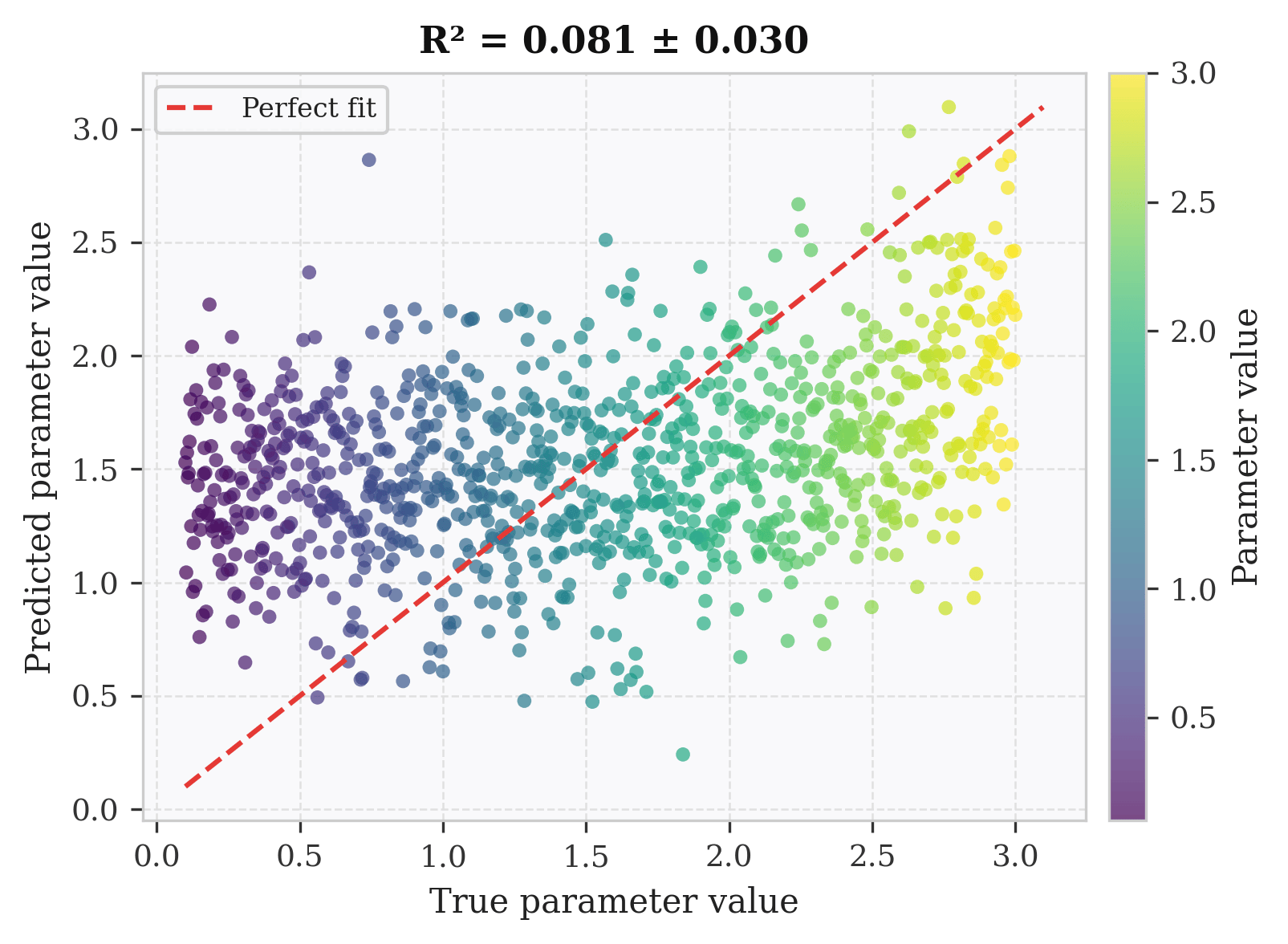}
        \caption{}
    \end{subfigure}

    \caption{
        Linear decodability comparison of Oz channel across oscillatory frequencies ($f_{\text{osc}}$) with varying power of the oscillation ($A_{\text{osc}}$) for CSBrain and LaBraM model.
    }
    \label{fig:oz_oscfreq_power}
\end{figure}

Additional linear decodability plots for channels across different EEG montage lobes, Frontal (Fz) [Fig~\ref{fig:Fz}, Fig~\ref{fig:fz_oscfreq_power}], Parietal (Pz) [Fig~\ref{fig:Pz}, Fig~\ref{fig:pz_oscfreq_power}] and Occipital (Oz) [Fig~\ref{fig:Oz}, Fig~\ref{fig:oz_oscfreq_power}] channels are included. 
The linear decodability values for CBraMod across different channels are identical owing to the convolution layers used to compute the channel embeddings in the model architecture. As we perform the synthetic experiments on single channel inputs, CBraMod would provide identical embeddings as outputs.


\section{Additional t-SNE Plots}
\label{sec:additional_tsne_plots}
\begin{figure}[H]
    \centering
    \includegraphics[width=\linewidth]{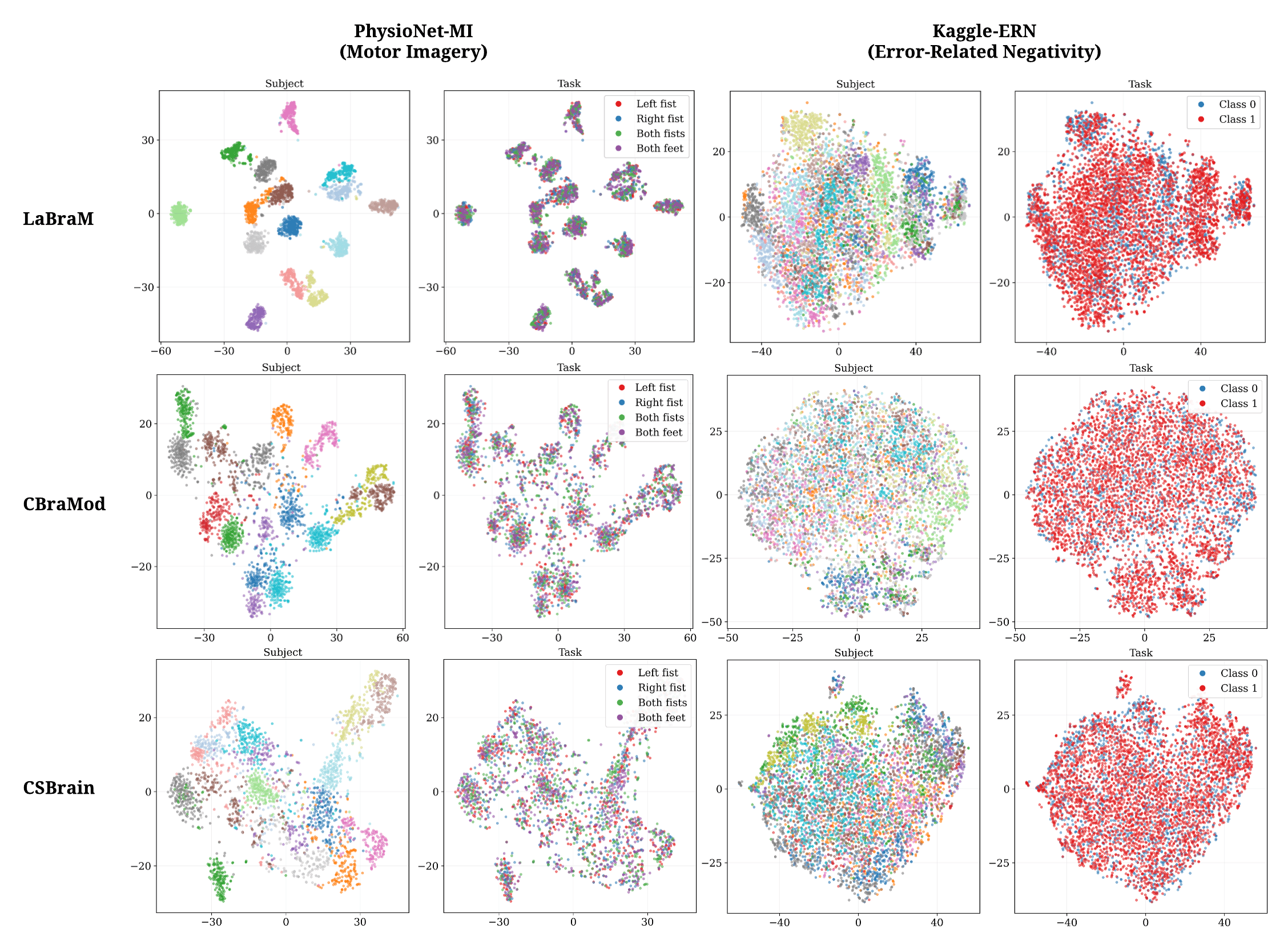}
    \caption{\textbf{t-SNE embeddings of pre-trained LaBraM, CBraMod, and CSBrain models across the PhysioNet-MI and Kaggle-ERN datasets.}}
    \label{fig:tsne-apd}
\end{figure}

\end{document}